\documentclass[manuscript,screen]{acmart}
\usepackage{hyperref}
\usepackage{graphicx}
\usepackage{amsmath}

\DeclareMathOperator*{\argmin}{argmin}
\DeclareMathOperator*{\argmax}{argmax}
\usepackage{algorithmic}
\usepackage{array}
\usepackage{url, multirow, rotating}

\AtBeginDocument{%
  \providecommand\BibTeX{{%
    \normalfont B\kern-0.5em{\scshape i\kern-0.25em b}\kern-0.8em\TeX}}}

\setcopyright{acmcopyright}
\copyrightyear{2020}
\acmYear{2020}
\acmDOI{10.1145/1122445.1122456}

\acmConference[TKDD]{TKDD}{Aug 2020}{TKDD}
\acmBooktitle{TKDD}
\acmPrice{15.00}
\acmISBN{978-1-4503-XXXX-X/18/06}

\begin{document}

\title{Compression of Deep Learning Models for Text: A Survey}

\author{Manish Gupta}
\email{gmanish@microsoft.com}
\orcid{0002-2843-3110}
\author{Puneet Agrawal}
\email{punagr@microsoft.com}
\affiliation{%
  \institution{Microsoft, India}
}
\renewcommand{\shortauthors}{Gupta and Agrawal}

\begin{abstract}
 In recent years, the fields of natural language processing (NLP) and information retrieval (IR) have made tremendous progress thanks to deep learning models like Recurrent Neural Networks (RNNs), Gated Recurrent Units (GRUs) and Long Short-Term Memory (LSTMs) networks, and Transformer~\cite{vaswani2017attention} based models like Bidirectional Encoder Representations from Transformers (BERT)~\cite{devlin2018bert}, Generative Pre-training Transformer (GPT-2)~\cite{radford2019language}, Multi-task Deep Neural Network (MT-DNN)~\cite{liu2019multi}, Extra-Long Network (XLNet)~\cite{yang2019xlnet}, Text-to-text transfer transformer (T5)~\cite{raffel2019exploring}, T-NLG~\cite{rosset2019turing} and GShard~\cite{lepikhin2020gshard}. 
But these models are humongous in size.
On the other hand, real world applications demand small model size, low response times and low computational power wattage. In this survey, we discuss six different types of methods (Pruning, Quantization, Knowledge Distillation, Parameter Sharing, Tensor Decomposition, and Sub-quadratic Transformer based methods) for compression of such models to enable their deployment in real industry NLP  projects. Given the critical need of building applications with efficient and small models, and the large amount of recently published work in this area, we believe that this survey organizes the plethora of work done by the `deep learning for NLP' community in the past few years and presents it as a coherent story.
\end{abstract}

\ccsdesc[500]{Computing methodologies~Neural networks}
\ccsdesc[500]{Computing methodologies~Machine learning}
\ccsdesc[500]{General and reference~Surveys and overviews}
\keywords{Model compression, Deep Learning, Pruning, Quantization, Knowledge Distillation, Parameter Sharing, Tensor Factorization, Sub-Quadratic Transformers}

\maketitle

Deep learning models have revolutionized multiple fields of information systems including natural language processing (NLP), computer vision, and speech analysis. While they have enabled multiple tasks to attain very high accuracy values, model sizes and prediction latencies have increased significantly as well. Specific to text, Recurrent neural networks (RNNs), Gated Recurrent Units (GRUs) and long short term memory (LSTM) networks have been used for quite some time for various natural language processing (NLP) tasks. These models are large especially because of the input and output embedding parameters. 

In the past three years, the field of NLP has made significant progress as is evident from the GLUE~\cite{wang2019glue} and SuperGLUE~\cite{wang2019superglue} leaderboards\footnote{\url{https://gluebenchmark.com/leaderboard}}$^,$\footnote{\url{https://super.gluebenchmark.com/leaderboard}}. Transformer~\cite{vaswani2017attention} based models like Bidirectional Encoder Representations from Transformers (BERT)~\cite{devlin2018bert}, Generative Pre-training Transformer (GPT-2)~\cite{radford2019language}, Multi-task Deep Neural Network (MT-DNN)~\cite{liu2019multi}, Extra-Long Network (XLNet)~\cite{yang2019xlnet}, MegatronLM~\cite{shoeybi2019megatron}, Text-to-text transfer transformer (T5)~\cite{raffel2019exploring}, T-NLG~\cite{rosset2019turing} and GShard~\cite{lepikhin2020gshard} have been major contributors to this success. But these models are humongous in size: BERT (340M parameters), GPT-2 (1.5B parameters), MegatronLM (8.3B parameters), T5 (11B parameters), T-NLG (17B parameters) and GShard (600B parameters). Bianco et al.~\cite{bianco2018benchmark} and Sanh et al.~\cite{sanh2019distilbert} provide a great overview of the sizes of recent deep learning models in computer vision and NLP respectively.

Training these large models needs a lot of GPU infrastructure, and leads to a large power consumption. Training a BERT-base model on GPUs is roughly equivalent to a trans-American flight in terms of power and carbon footprint~\cite{strubell2019energy}. Deployment of such gigantic models is difficult even on high-end servers. Indeed a large number of real world applications run on machines with resource constrained environments, for example, edge devices, sensors and mobile phones. Edge devices could include offline medical equipment, and modules on drones, robots, glasses, etc. Often times, besides desiring a small model size, low response times are critical. For example, applications like driverless cars or apps to aid the blind require processing speeds of around 30 frames per second. Similarly, search engines need to be able to process billions of queries per day.  Overall, the following factors motivate us to study compression of deep learning models.
\begin{itemize}
    \item Memory (RAM) usage
    \item Prediction latency
    \item Power dissipation
    \item Inference on resource constrained devices
    \item Ease of training/finetuning
    \item Ease of deployment and update
    \item Ease of distributed training
\end{itemize}

\begin{figure*}
    \centering
    \includegraphics[width=\textwidth]{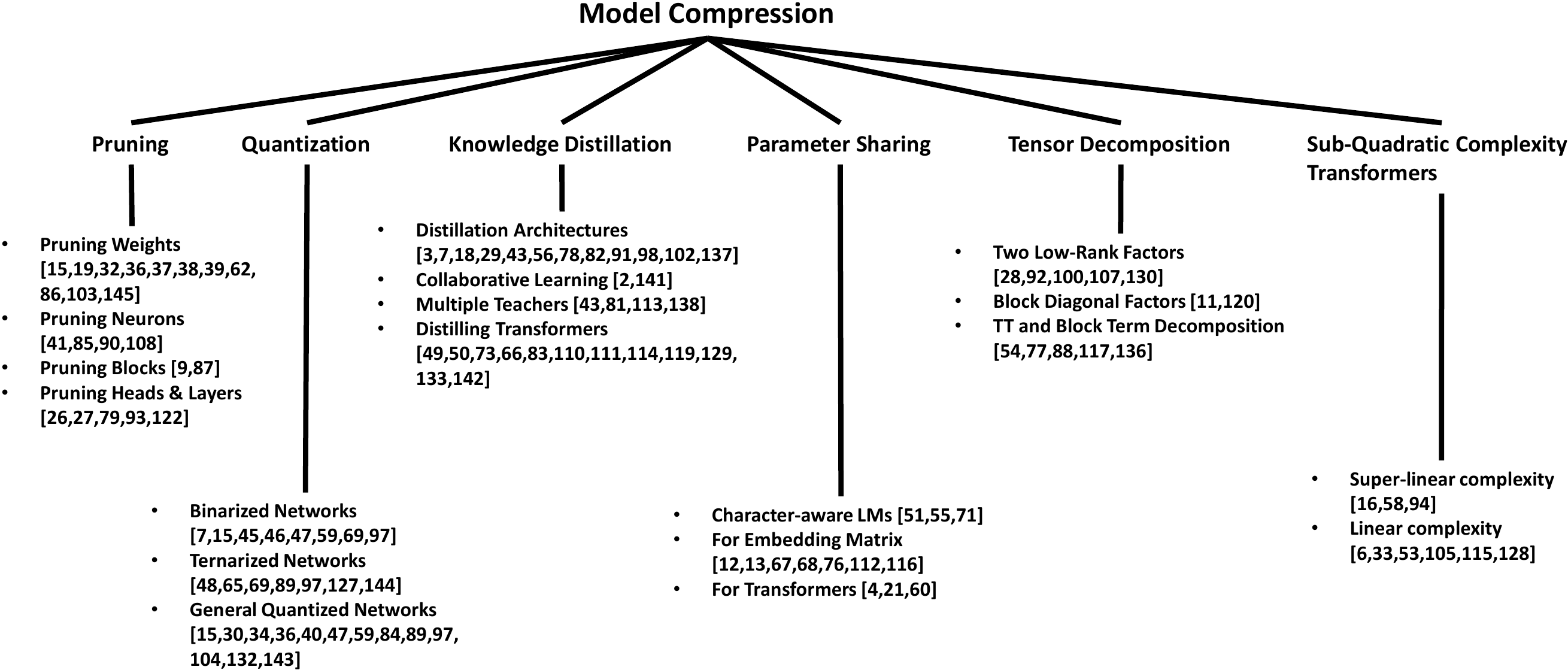}
    \caption{Overview of Model Compression Methods for Text}
    \label{fig:overview}
\end{figure*}

Large networks do not fit in on-chip storage and hence require the more costly external DRAM accesses. Running a 1 billion connection neural network, for example, at 30 frames per second would require (30Hz)(1G)(640pJ) = 19.2W just for DRAM access -- well beyond the power envelope of a typical mobile device. This implies that a mobile phone running such an app could suffer from fast draining of the battery, leading to overheating of the phone. Han et al.~\cite{han2016eie} discuss details of power dissipation for deep learning models. Another option to avoid large RAM, high prediction times and high power dissipation, is to run such massive deep learning models on cloud servers. But for many real world applications, it is desirable to run them on local client devices to avoid network delay, to guard user privacy and to avoid power dissipation in terms of input/output data communication. 

Small models can indeed also lead to low prediction latencies. For example, Diamos et al.~\cite{diamos2016persistent} showed that for small models, one can cache the RNN weights in on-chip memory such as caches, block RAM, or register files across multiple timesteps. This could lead to as much as 146x speedup if the entire RNN layer can be stored in registers rather than the GPU DRAM of an NVIDIA TitanX GPU.

Finally, it is easier to perform software development, deployment and updates with smaller models. Large models are difficult to handle. For example, it is impossible to fine-tune pretrained BERT-large model on a GPU with 12-16 GB RAM. This poses a large barrier of entry for communities without the resources to purchase several large Graphic Processing Units (GPUs). For large models, tuning various configuration parameters needs lots of resources. Smaller models lead to improved speed of learning and allow for more hyper-parameter configurations to be evaluated. Mobile-first companies dislike large apps. App stores are very sensitive to the size of the binary files. For example, iPhone App Store has the restriction ``apps above 150 MB will not download until you connect to Wi-Fi''. Smaller models are easier to use and deploy in real world systems. Large models  need multiple server nodes. On the other hand, multiple instances of smaller models can be run simultaneously on the same machine leading to higher QPS (queries per second). Lastly, smaller models also decrease the communication overhead of distributed training of the models.

Fortunately, there is a large amount of redundancy among the weights of these large neural networks. A small subset of the weights are sufficient to reconstruct the entire network. Denil et al.~\cite{denil2013predicting} showed that by simply using $\sim$5\% of the weights, it is possible to predict the remaining weights without any drop in accuracy. This observation led to a large number of research efforts across multiple communities on compression of large deep learning models. In this survey, we aim to systematically explore this large body of literature in the NLP community by organizing it into various categories. Figure~\ref{fig:overview} shows a broad organization of model compression methods for text. 
In this survey we do not focus on specific methods proposed in other communities like vision and speech only and which make use of image/audio specific architectures and hence cannot be applied to text. Also, we do not discuss methods on hardware optimizations to reduce latency. While there are other surveys in the broad area of model compression~\cite{cheng2017survey,deng2020model} also specifically on knowledge distillation~\cite{wang2020knowledge}, they are either old or focus on computer vision related problems.

\section{Model Compression Methods: Overview}
\label{sec:overview}

In this survey, we discuss compression methods using pruning, quantization, knowledge distillation, parameter sharing, tensor decomposition and sub-quadratic Transformers. 

The most obvious way to reduce model size is to sparsify weight matrices. Pruning methods differ based on what is pruned and the actual logic used to prune. Given a matrix, one can prune 
\begin{itemize}
    \item Some weight entries
    \item Rows or columns (i.e., neurons)
    \item Weight blocks
    \item Attention heads (in case of Transformer based methods)
    \item Layers or a combination of the structures.
\end{itemize}

Other interesting aspects of pruning methods include the following.
\begin{itemize}
    \item How to decide which weights/neurons/blocks/heads to prune?
    \item Should you prune large networks or build small networks?
    \item Should you do one-shot pruning versus iterative/gradual pruning? 
    \item How does regularization help when pruning?
\end{itemize}
 We discuss these aspects of pruning based methods in Section~\ref{sec:pruning}.

Besides removing the weights themselves, another way to reduce the size of weight matrices is to reduce the number of bits needed to represent each weight. Typically weights are stored as 32-bit double values. In an extreme case, weights can be quantized to two values (binary 1 bit). But other popular ways include quantization to three values (ternary) or multiple bits. Quantization can be uniform or non-uniform. Quantization methods can be deterministic or stochastic. Quantization can be performed in a loss-aware or unaware manner. Quantization bin ranges can be trained versus tuned. Finally, the level of quantization needs to be different across layers of RNNs, LSTMs or Transformers to attain a favorable model size versus accuracy tradeoff. We discuss these aspects of quantization based methods in Section~\ref{sec:quantization}.

A third way of doing model compression is using knowledge distillation (also broadly known as student teacher networks). In such methods, the idea is to first train a deep teacher model using labeled data, and then transfer ``dark knowledge'' from teacher to train a shallow student network. Thus, the student model is trained to mimic a pre-trained, larger teacher. After the student is trained, it is deployed while the teacher network can be thrown. Distillation methods vary based on the following design choices.

\begin{itemize}
    \item Different types of teacher model
    \item Different types of loss function like squared error between the logits of the models, KL divergence between the predictive distributions, or some other measure of agreement between the model predictions
    \item Different choices for what dataset the student model trains on (a large unlabeled dataset, a held-out data set, or the original training set)
    \item Different choices for what to mimic from the teacher -- teacher's class probabilities, teacher's feature representation.
    \item Learn from whom -- teacher, teacher assistant, or other fellow students. 
\end{itemize}
We discuss these aspects of knowledge distillation based methods in Section~\ref{sec:knowledgeDistillation}.

Another way that reduces overall weights is to have parameters shared across multiple weight structures. Broadly, the method is called parameter sharing. Methods differ depending on the following aspects.
\begin{itemize}
    \item Which parameters are shared
    \item Technique used to share parameters
    \item The level at which sharing is performed
\end{itemize}
We discuss these aspects of parameter sharing based methods in Section~\ref{sec:parameterSharing}.

Yet another way to avoid large matrices is to approximate them using a combination of smaller matrices. Such tensor decomposition methods for model compression factorize large tensors into multiple smaller components. Methods differ based on the following aspects.
\begin{itemize}
    \item Type of factorization technique
    \item Matrices being factorized
    \item The property of weight matrix being exploited
\end{itemize}
We discuss these aspects of tensor decomposition methods in Section~\ref{sec:matrixDecomposition}.

In Transformer based models, besides the model size, latency is a concern because of quadratic complexity in terms of the input sequence size. Also, RAM needed for Transformer models is quadratic in nature. Hence, very recently, there have been several efforts on designing Transformers with sub-quadratic complexity. Some of these methods are super-linear while many are linear. Linear complexity methods use various techniques for enforcing linearity -- the broad idea is to compute a transformed representation for every token using attention over a fixed small number of other tokens. Methods differ in terms of defining the set of other tokens to be used to compute a transformed representation for the current token. We discuss such methods in Section~\ref{sec:linearTransformers}.

\section{Pruning}
\label{sec:pruning} 

\begin{figure*}
    \centering
    \includegraphics[width=\textwidth]{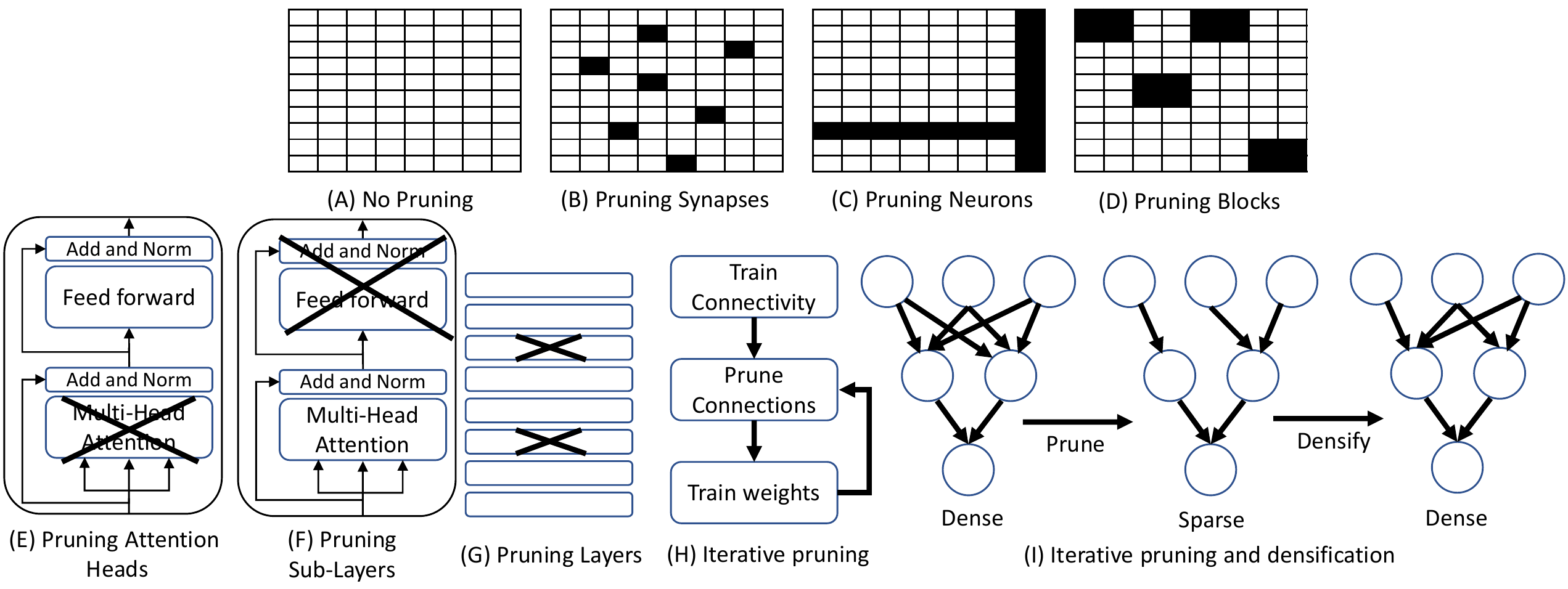}
    \caption{Different Types of Pruning: (A) represents no pruning. Filled cells represent pruned entries. (B), (H) and (I) are unstructured pruning methods discussed in Section~\ref{subsec:PruningWeights}. Remaining are structured pruning methods: (C) is discussed in Section~\ref{subsec:PruningNeurons}, (D) in Section~\ref{subsec:PruningBlock}, (E), (F) and (G) in Section~\ref{subsec:PruningHeadsLayers}.}
    \label{fig:pruning}
\end{figure*}

The first proposed methods for model compression were based on pruning. Pruning can be done in two ways: structured versus unstructured. In unstructured pruning, individual weight connections are removed from a network by setting them to 0. One can prune away weights from a weight matrix depending on various criteria (e.g., prune away low magnitude weights). Such unstructured pruning methods lead to sparse matrices and need special sparse matrix manipulation libraries at inference time. Hence, various structured pruning methods have also been proposed which aim to prune away structures like neurons, weight matrix blocks, attention heads or layers. Fig.~\ref{fig:pruning} provides a broad overview of various pruning styles. Fig~\ref{fig:pruning}(B) depicts unstructured pruning. Fig~\ref{fig:pruning}(C)-(G) shows various structured pruning methods. In this section, we discuss unstructured weight pruning methods in Section~\ref{subsec:PruningWeights} and structured pruning methods in Sections~\ref{subsec:PruningNeurons}-\ref{subsec:PruningHeadsLayers}.

In pruning, the main idea is to grow a large model and then prune away weights to end up with a much smaller but effective model. This is inspired by the following biological observation. Trillions of synapses are generated in the human brain during the first few months of birth. At one year old, synapse count peaks at 1000 trillion. And then natural pruning begins to occur. A ten year old child has nearly 500 trillion synapses. This `pruning' mechanism removes redundant connections in the brain~\cite{walsh2013peter}. 

One natural question is should you prune large networks or build small dense networks? Pruning involves extra processing plus sparse matrices need special handling. Can we avoid it by training smaller models? Zhu et al.~\cite{zhu2017prune} experimented with models of various sizes with/ without pruning of stacked LSTMs models for language modeling, and seq2seq models for Neural Machine Translation (NMT). They found that large-sparse models consistently outperform small-dense models and achieve up to 10x reduction in number of non-zero parameters with minimal loss in accuracy. 

\subsection{Pruning Weights}
\label{subsec:PruningWeights}
The weights to be pruned can be chosen using two heuristics: (1) Using Hessian matrix computation, or (2) using magnitude of the weights. Further, magnitude based pruning methods could do pruning in one shot (typically statically after training is done), or iteratively (also called dynamic or gradual pruning), or iteratively with pruning and densification. Accordingly, there are four main ways of performing unstructured weight pruning: (1) Hessian based methods, (2) magnitude pruning, (3) iterative magnitude pruning, and (4) iterative magnitude pruning and densification. In this subsection, we discuss these methods in detail. 

\subsubsection{Hessian based Methods}
In their seminal work (Optimal Brain Damage or OBD) on proposing weight pruning as a method for model compression, LeCun et al.~\cite{lecun1990optimal} defined saliency of a weight as change in the objective function $E$ caused by deleting that parameter. Using Taylor series and making multiple assumptions, they propose to use the following as a measure of saliency of weight $u_i$.
\begin{eqnarray}
    s(u_i)=\frac{1}{2}\sum_i h_{ii} \delta u_i^2
\end{eqnarray}
\noindent where $h_{ii}=\frac{\partial^2 E}{\partial u_i \partial u_j}$ is the $i^{th}$ element on the diagonal of the Hessian matrix. Weights with low saliency can be pruned and the pruned network can be retrained to adjust the remaining weights. 
The procedure for computation of the diagonal of the Hessian is very similar to the back-propagation algorithm used for computing the first derivatives. Hence, computing the diagonal of the Hessian is of the same order of complexity as computing the gradient. 

OBD ignores cross terms in the Hessian matrix. But on most real datasets, Hessian is strongly non-diagonal. Hence, to avoid pruning of important weights, Hassibi et al.~\cite{hassibi1993second} proposed a method called Optimal Brain Surgeon (OBS) which considers cross terms as well. Using a similar derivative of $E$ wrt weight $w_i$, saliency of the weight is computed as follows.
\begin{eqnarray}
L_i=\frac{1}{2}\frac{w_i^2}{[H^{-1}]_{ii}}    
\end{eqnarray}
Computing $H^{-1}$ is difficult. Hence, they provide a faster recursion relation for calculating $H^{-1}$ from training data and structural information of the network. Also, unlike other methods (like OBD or magnitude pruning), OBS does not demand (typically slow) retraining after the pruning of a weight. In every step, we delete $w_i$  with min $L_i$ and update all weights as follows.
\begin{eqnarray}
    \delta w=-\frac{w_i}{[H^{-1}]_{ii}}H^{-1}e_i
\end{eqnarray}
\noindent where $e_i$ is the unit vector in weight space corresponding to (scalar) weight $w_i$. Unfortunately, these methods (OBD and OBS) are computationally prohibitive as second derivative (i.e. Hessian) computations are expensive. 

\subsubsection{Magnitude Pruning Methods}
A more computationally feasible method for pruning connections and relearning weights based solely on the magnitude of the original weights is to simply prune away weights with small magnitudes. The idea was first proposed by Han et al.~\cite{han2015deep}. Pruning away low magnitude weights makes the matrix sparse. Sparse matrices can be stored in Compressed Sparse Row/Column (CSR/CSC) formats. Further space can be saved by storing the index difference instead of the absolute position, and encoding this difference using small fixed number of bits. See et al.~\cite{see2016compression} experimented with these pruning methods on encoder-decoder LSTM NMT (neural machine translation) models. They perform magnitude pruning on all weight matrices of a 4-layer LSTM. They find that higher layers, attention and softmax weights are the most important, while lower layers and the embedding weights hold a lot of redundancy. At the lower layers the parameters for the input are most crucial, but at higher layers the parameters for the gates also become important. These methods typically have a target pruning percent as a hyper-parameter and pruning is either performed statically (after training the full model) or dynamically (while training itself). Retraining the sparse pruned network helps in improving accuracy.

In a typical encoder-decoder LSTM model, there are these weight classes: source embedding weights, target embedding weights, source layer weights, target layer weights, attention weights and softmax weights. An important consideration related to magnitude pruning is how do we distribute the pruning over these different weight classes of a model, given a target $x$\% pruning? Three ways suggested by See et al.~\cite{see2016compression} include class-blind, class-uniform and class-distribution. In the class-blind way, we take all parameters, sort them by magnitude and prune the $x$\% with smallest magnitude, regardless of the weight class. So some classes are pruned proportionally more than others. In the class-uniform way, Within each class, we sort the weights by magnitude and prune the x\% with smallest magnitude. So all classes have exactly x\% of their parameters pruned. In the class-distribution scheme, for each class $c$, weights with magnitude less than $\lambda\sigma_c$ are pruned. Here, $\sigma_c$ is the standard deviation of that class and $\lambda$ is a universal parameter chosen such that in total, $x$\% of all parameters are pruned. Class-blind pruning is the simplest and adheres to the principle that pruning weights (or equivalently, setting them to zero) is least damaging when those weights are small, regardless of their locations in the architecture. Class-uniform pruning and class-distribution pruning both seek to prune proportionally within each weight class, either absolutely, or relative to the standard deviation of that class. They observe that class-blind pruning outperforms both other schemes. 

\subsubsection{Iterative Magnitude Pruning Methods}
Typically, it has been seen that rather than pruning in one-shot, it is a good idea to prune gradually over epochs. This way of pruning is called iterative or gradual pruning (see Fig.~\ref{fig:pruning}(H)). For starting proportion $x$\% and ending proportion $y$\%,  iterative magnitude pruning procedure prunes $x$\% of each of the weights, does re-training, and then prunes $(y-x)/T$\% of the weights every $K$ iterations. $T$ is the number of times, pruning is done. Sometimes, pruning is started after few warmup iterations have already been performed. Magnitude pruning has been seen to be very effective with regularization ($L_1/L_2$) while training. Dropouts also help in effective pruning. 
In some pruning methods, a weight once pruned cannot be a part of the network in future iterations. On the other hand, other methods do not modify the gradients of a pruned weight in the back-propagation step. In that case, it is possible for the updates of a pruned weight to be larger than the threshold of that layer, and then the weight will be involved in the forward pass again. Also, in every pruning iteration, we could either use a fixed threshold~\cite{han2015learning} or monotonically increase it~\cite{narang2017exploring}. 

In case of gradual pruning~\cite{narang2017exploring}, where pruning threshold $\epsilon$ is monotonically increased, $\epsilon$ is determined as follows in every iteration $i$. Let $f$ be the number of iterations after which $\epsilon$ is updated. Also, after a few warmup iterations, weights are sorted using absolute values and we pick the weight corresponding to the $90^{th}$ percentile as $q$. Pruning threshold $\epsilon$ is increased in two stages. In the first stage (which starts at start iteration $s$ and continues until ramp iteration $r$, we use $\theta$ as the initial slope to prune weights. In the second stage (which starts at ramp iteration $r$ and continues until end iteration $e$), we use $\phi$ as the ramp slope to change the rate of pruning. Typically, $\phi$ is set to 1.5$\theta$ where $\theta$ is calculated as follows.
\begin{eqnarray}
    \theta=\frac{2qf}{2(r-s)+3(e-r)}
\end{eqnarray}
Thus, from iteration $s$ to $r$, we set the pruning threshold as follows.
\begin{eqnarray}
\epsilon=\frac{\theta(i-s+1)}{f}    
\end{eqnarray}
From iterations $r+1$ to $e$, we set the pruning threshold as follows.
\begin{eqnarray}
\epsilon=\frac{\theta(r-s+1)+\phi(i-r+1)}{f}
\end{eqnarray}
Typically when pruning, biases are not pruned since they are much fewer in number. Overall, RNN/LSTM model size can be reduced by 90\% and speed-up is around 2x to 7x using gradual pruning with no deterioration in accuracy. Also, layers closer to input are pruned more aggressively compared to the final layers.

Another way of performing iterative pruning is to set a pruning target per iteration~\cite{zhu2017prune}. In this scheme, we start with an initial sparsity value $s_0$. To achieve a final sparsity value of $s_f$ after $n$ pruning steps with pruning frequency $f$, pruning target in iteration $i$ can be computed as follows.
\begin{eqnarray}
s_i=s_f+(s_0-s_f)\left(1-\frac{i}{nf}\right)^3
\end{eqnarray}
Thus, the sparsity of the network is gradually increased while allowing the network training steps to recover from any pruning-induced loss in accuracy. We prune the network rapidly in the initial phase when the redundant connections are abundant and gradually reduce the number of weights being pruned each time as there are fewer and fewer weights remaining in the network.

Cheong et al.~\cite{cheong2019transformers} found that iterative pruning leads to poor results when pruning Transformer models like BERT. Guo et al.~\cite{guo2019reweighted} found that there are two problems with pruning especially when done with regularization. 
\begin{itemize}
    \item The larger weights $w_j$ are penalized more heavily than smaller weights $w_i$ in $L_1$ regularization, which violates the original intention of weight pruning, ``removing the unimportant connections''. 
    \item Direct optimization of a regularization penalty term causes divergence from the original loss function and has negative effect on the effectiveness of gradient-based update.
\end{itemize}
They propose to perform reweighted $L_1$ minimization where $\alpha_i>0$ are inversely proportional to magnitude of corresponding weights $|w_i|$. Thus, they solve the following optimization problem
\begin{eqnarray}
    \min_w f(w)+\gamma \sum_i \alpha_i |w_i|
\end{eqnarray}
\noindent where $f(w)$ is the original loss function for the network. This optimization is solved using a reweighted proximal pruning (RPP) method (which depends on proximal operators). RPP decouples the goals of high sparsity from minimizing loss, and hence leads to improved accuracy even with high levels of pruning for BERT. 

\subsubsection{Iterative Magnitude Pruning and Densification}
Further, the effectiveness of pruning can be improved by performing pruning and densification~\cite{han2016dsd,dai2018grow} alternately across multiple iterations (see Fig.~\ref{fig:pruning}(I)). There are two ways of doing this. In the first method~\cite{han2016dsd}, in each iteration, either pruning is performed or densification. The sparse training regularizes the model, and the dense training restores the pruned weights, increasing the model capacity without overfitting. Sparsification helps the optimizer escape saddle points, and leads to regularized training which converges to a significantly better minima. In the second method~\cite{dai2018grow}, in every iteration some dormant weights can reappear in the network while other active ones can get pruned out. A dormant $w\in W$ is activated iff $|w.grad|$ is larger than the $(100\alpha)^{th}$ percentile of all elements in $|W.grad|$. A $w\in W$ is removed iff $|w|$ is smaller than the $(100\beta)^{th}$ percentile of all elements in $|W|$. $\alpha$ and $\beta$ refer to growth ratio, and pruning ratio, respectively.

\subsection{Pruning Neurons}
\label{subsec:PruningNeurons}
It is difficult to implement unstructured pruning practically since, at inference time, special support is needed for matrix multiplication in the sparse space. Pruning away neurons leads to removal of a row or a column from a weight matrix, thereby avoiding sparse matrix handling  (see Fig.~\ref{fig:pruning}(C)). However, compared to pruning weights, pruning neurons is less flexible since we need to find entire rows/columns for deletion. In this section, we discuss ways of determining neurons that can be pruned.

\subsubsection{Removing Low Importance Neurons}
He et al.~\cite{he2014reshaping} proposed three node importance functions to determine importance score for neurons. 
\begin{itemize}
    \item Entropy: For a neuron $i$, let $a_i$ ($d_i$) be the \#instances with node output $> (or\leq)$ 0.5 for binary classification with a sigmoid activation. Then entropy for node $i$ can be written as follows.
    \begin{eqnarray}
    \text{Entropy}(i)=\frac{d_i}{a_i+d_i}\log_2 \frac{d_i}{a_i+d_i}+\frac{a_i}{a_i+d_i}\log_2 \frac{a_i}{a_i+d_i}  
    \end{eqnarray}
    The intuition is that if one node's outputs are almost identical on all training data, these outputs do not generate variations to later layers and consequently the node may not be useful. 
    \item Output-weights Norm (onorm): average $L_1$-norm of the weights of its outgoing links.
    \item Input-weights norm (inorm): average $L_1$-norm of the weights of its incoming links.
\end{itemize}

All the neurons are sorted by their scores and nodes with less importance values are removed. In most cases, onorm has been found to be the best among these importance functions. 

Special regularizers have also been proposed to force neurons to push either all incoming or outgoing connection weights towards zero~\cite{murray2015auto, pan2016dropneuron}. Specifically, for handling incoming connections, the following two regularizers are popular.
\begin{itemize}
    \item $L_2$ norm on weight matrix $W$ defined as follows.
    \begin{eqnarray}
        \sum_i ||W_{i:}||_2=\sum_i \left(\sum_j W^2_{ij}\right)^{1/2}
    \end{eqnarray}
    This puts equal pressure on each row, but within each row, the larger values contribute more, and therefore there is more pressure on larger values towards zero. 
    \item $L_\infty$ norm on weight matrix $W$ defined as follows.
    \begin{eqnarray}
        \sum_i ||W_{i:}||_\infty=\sum_i \max_j |W_{ij}|
    \end{eqnarray}
    This puts equal pressure on each row, but within each row, only the maximum value (or values) matter, and therefore the pressure towards zero is entirely on the maximum value(s).
\end{itemize}
Similar regularizers can easily be defined for outgoing connections as well.

\subsubsection{Removing Redundant Neurons}
Consider a simple network with one hidden layer with $n$ neurons. Thus, the output can be computed as follows.
\begin{eqnarray}
z=a_1 h(W_1^TX)+a_2 h(W_2^TX)+...+a_n h(W_n^TX)
\end{eqnarray}
\noindent where $a_i$ and $W_i$ indicate weights. In case $W_1==W_2$, $h(w_1^TX)=h(w_2^TX)$. Thus, we can compute output as follows.
\begin{eqnarray}
z=(a_1+a_2) h(W_1^TX)+0. h(W_2^TX)+...+a_n h(W_n^TX)
\end{eqnarray} 
In general, whenever two weight sets ($W_1$, $W_2$) are equal, one of them can effectively be removed. This should be done with a surgery step, i.e., we need to alter the co-efficient $a_1$ to $a_1+a_2$. Of course, for many pairs of weight sets (i.e., neurons), $W_1$ and $W_2$ are not exactly the same. Hence, Srinivas et al.~\cite{srinivas2015data} proposed this 3 step method for redundant neuron identification and removal. 
\begin{itemize}
    \item Compute saliency $s_{ij}$ for all possible neuron pairs (i, j) as follows.
    \begin{eqnarray}
        s_{ij}=\langle a_j^2 \rangle||\epsilon_{ij}||_2^2
    \end{eqnarray} 
    \noindent where $\langle a_j^2 \rangle$ denotes the average of the quantity over all output neurons. Let $S$ be the matrix with all $s_{ij}$ values. \item Pick the indices $(i',j')$ corresponding to the minimum $s_{ij}$. Delete the $j'$ neuron, and update $a_i'$ as follows.
    \begin{eqnarray}
    a_i'\leftarrow a_i'+a_j'    
    \end{eqnarray}
    \item Update $S$ by removing the $j'^{th}$ column and row, and updating the $i'^{th}$ column (to account for the updated $a_i'$).
\end{itemize}

\subsection{Pruning Blocks}
\label{subsec:PruningBlock}

In weight pruning, irregularity of sparse matrices limits the maximum performance and energy efficiency achievable on hardware accelerators. Pruning neurons avoids sparse matrix issues but is limited in term of overall pruning possible. Hence, block based pruning methods were introduced  (see Fig.~\ref{fig:pruning}(D)).

Block-sparse formats store blocks contiguously in memory reducing irregular memory accesses. If the maximum magnitude weight of a block is below the current threshold, we set all the weights in that block to zeros. Similar to iterative weight pruning, block pruning~\cite{narang2017block} can also be done iteratively using a monotonically growing threshold. Any blocks that had been zeroed out are held at zero even after pruning has ended resulting in a sparse model at the end of training. Just like weight pruning (as discussed in Section~\ref{subsec:PruningWeights}), the start slope $\theta$ and ramp slope $\phi$ determine the rate at which the threshold increases. For block pruning, we need to modify the start slope to account for the number of elements in a block ($N_b$). Thus, the start slope for block pruning is typically set as follows.
\begin{eqnarray}
\theta_b=\theta\times\sqrt[\leftroot{-2}\uproot{2}4]{N_b}    
\end{eqnarray}
Further, to enable effective removal of blocks, Narang et al.~\cite{narang2017block} propose group Lasso regularization method. Group lasso is a type of weight regularization that works on groups of weights and can zero out all the weights in a group. For each block, we add a loss term proportional to the $L_2$ norm of the block. Thus, we optimize for the following.
\begin{eqnarray}
\min_w f(w)+\lambda_g \sum_{g=1}^{G} ||w^{(g)}||_2    
\end{eqnarray}
When we combine group lasso with block pruning,  group lasso guides the selection of blocks to prune. Group lasso regularization is applied to coincide with the pruning schedule, i.e., we turn off regularization when the pruning schedule ends. Typically, inducing block sparsity with 4x4 blocks in vanilla RNNs and GRUs works well, compared to larger block sizes. Larger blocks require lower sparsity to maintain similar accuracy. 

Unfortunately, it becomes challenging to maintain the same model accuracy when block sparsity is applied. Also, block sizes (i.e., pruning granularity) are application-sensitive, making it another hyper-parameter to tune. To avoid these problems, Cao et al.~\cite{cao2019efficient} proposed a new method called Bank-Balanced Sparsity (BBS). BBS splits each weight matrix row into multiple equal-sized banks, and adopts fine-grained pruning to each bank independently to obtain identical sparsity among banks. Each bank has the same number of non-zero values. For example, retaining top two weights in each bank of size 4 implies a sparsity of 50\%. We apply the BBS pruning method iteratively to a pre-trained network, and fine-tune the network after each pruning iteration to restore the model accuracy. BBS achieves almost the same model accuracy as unstructured sparsity and significantly outperforms block sparsity when pruning weights at the same sparsity level. BBS is also amenable to FPGA (Field Programmable Gate Arrays) acceleration because it inherently provides a balanced matrix partitioning for parallel computing. 

\subsection{Pruning Heads and Layers}
\label{subsec:PruningHeadsLayers}

Besides neurons and blocks, for Transformer based models, structured pruning can also be applied to attention heads and entire layers. In this section, we discuss such methods.

\subsubsection{Pruning Attention Heads}
BERT-base model consists of 12 layers each with 12 attention heads. Similarly, a typical NMT encoder-decoder Transformer with 6 layers each for encoder as well as decoder contains 16 attention heads per layer. Michel et al.~\cite{michel2019sixteen} found that majority of attention heads can be removed without deviating too much from the original score. Surprisingly, in some cases removing an attention head  (see Fig.~\ref{fig:pruning}(E)) results in an increase in accuracy. When these heads are removed individually, only 8 (out of 96) heads in 6-layer WMT NMT Transformer (16 heads/layer) cause a statistically significant change in performance when they are removed from the model, half of which actually result in a higher BLEU score. For most layers, one head is indeed sufficient at test time, even though the network was trained with 12 (BERT) or 16 (WMT Transformer) attention heads. One can also do iterative pruning of multiple heads (rather than just one at a time) across layers. For iterative pruning, head importance score is defined using the expected sensitivity of the model to the mask variables $\xi_h$ as follows.
\begin{equation}
    I_h=E_{x\sim X}\left|\frac{\partial L(x)}{\partial \xi_h}\right|=E_{x\sim X}\left|\text{Att}_h(x)^T\frac{\partial L(x)}{\partial \text{Att}_h(x)}\right|
\end{equation}

\noindent where $X$ is the data distribution, $L(x)$ is the loss on sample $x$, and $Att_h(x)$ is the output of the attention head $h$ for instance $x$, . Intuitively, if $I_h$ has a high value then changing $\xi_h$ is liable to have a large effect on the model. Hence, in every iteration heads with low $I_h$ values are pruned out. Michel et al.~\cite{michel2019sixteen} observed that pruning up to 20\% and 40\% of heads from NMT and BERT models respectively, did not lead to any noticeable negative impact on accuracy.

Voita et al.~\cite{voita2019analyzing} used two other head importance scores to prune attention heads from the NMT model. The two scoring methods were: (1) Layer-wise relevance propagation (LRP)~\cite{ding2017visualizing}. LRP is a method for computing the relative contribution of neurons at one point in a network to neurons at another. (2) ``confidence'' of a head which is computed as the average of its maximum attention weight excluding the end of sentence symbol, where the average is taken over tokens in a set of sentences used for evaluation. For pruning the heads, they propose a method based on stochastic gates and a differentiable relaxation of the $L_0$ penalty. $L_0$ norm equals the number of non-zero components and pushes the model to switch off less important heads. They find that only a small subset of heads are important for translation. On the English-Russian WMT dataset, pruning 38 out of 48 encoder heads results in a drop of only 0.15 BLEU.

\subsubsection{Pruning Layers}
Note that dropping attention heads does not reduce runtime as they are usually computed in parallel. While one can prune weights, neurons or attention heads, how can we design a scheme to prune away layers  (see Fig.~\ref{fig:pruning}(G))? The LayerDrop idea proposed in~\cite{fan2019reducing} is inspired by DropConnect. DropConnect randomly drops weights while training on a batch. LayerDrop does structured dropout: it drops groups of weights, heads, feed forward network (FFN) matrices, or layers. The layers to be pruned can be decided using one of these ways: 
\begin{itemize}
    \item Every Other: Prune every other layer (with rate $p$), e.g., every $3^{rd}$ layer in a 12-layer BERT model. 
    \item Search on Validation: Search for a set of layers to be pruned by checking their impact on a validation set. This entails trying various combinations. 
    \item Data Driven Pruning: Learn the drop rate $p_d$ of each layer in a data driven manner.
\end{itemize}

Given a target drop rate $p$, we learn an individual drop rate $p_d$ for the layer at depth $d$ such that the average rate over layers is equal to $p$. At inference time, we forward only the fixed top-$k$ highest scoring layers based on the softmax output. Across the three methods, ``Every Other'' strategy works surprisingly well across many tasks and configurations. ``Search on Validation'' and ``Data Driven Pruning'' only offer marginal gains.

\subsubsection{Pruning General Structures}
Lastly, Prasanna et al.~\cite{prasanna2020bert} experiment with pruning both the FFN layers  (see Fig.~\ref{fig:pruning}(F)) as well as attention heads  (see Fig.~\ref{fig:pruning}(E)) in a BERT network. Just like~\cite{michel2019sixteen}, they assign a mask variable to each of these structures. To decide which structures to prune, we look at the expected sensitivity of the model to the mask variables. High sensitivity implies large impact on the model output and hence corresponding structures should be retained. They find that it is possible to find a subnetwork of elements that achieves performance comparable with that of the full model, and similarly-sized subnetworks sampled from the rest of the model perform worse. 

\begin{table}
    \centering
    \scriptsize
    \begin{tabular}{|l|l|l|l|l|l|l|l|l|}
    \hline
Task&Dataset&Model&Method&Size (Pruned; Orig)&Eval. (Pruned; Orig)&Metric\\
\hline
\hline
Language modeling&Europarl v7 English &2-layer MLP&Prune Neurons~\cite{murray2015auto}&5.06M; 5.1M&57; 100&Perplexity (L)\\
\hline
Language modeling&PTB&2-layer LSTM&Iter. Mag.~\cite{zhu2017prune}&6.6M; 66M&80.24; 78.45&Perplexity (L)\\
\hline
Language modeling&PTB&LSTM&Iter. Mag.~\cite{cao2019efficient}&20M; 66M&78.5; 78.8&Perplexity (L)\\
\hline
Language modeling&PTB&LSTM&Block Sparsity~\cite{cao2019efficient}&20M; 66M&83; 78.8&Perplexity (L)\\
\hline
Language modeling&PTB&LSTM&BBS~\cite{cao2019efficient}&20M; 66M&78.5; 78.8&Perplexity (L)\\
\hline
Language modeling&Wikitext-103&Transformer&LayerDrop~\cite{fan2019reducing}&22M; 44M&19.5; 18.2&Perplexity (L)\\
\hline
Language modeling&AFP from English Gigaword&2-layer MLP&Prune Neurons~\cite{murray2015auto}&5.07M; 5.1M&107; 100&Perplexity (L)\\
\hline
Linguistic acceptability&CoLA&BERT-large&RPP/Iter. Mag.~\cite{guo2019reweighted}&138M/170M; 340M&82.8/76.3; 81.5&Matthews (H)\\
\hline
Machine reading comp.&MRPC&BERT-large&RPP/Iter. Mag.~\cite{guo2019reweighted}&138M/170M; 340M&88.1/83.5; 89.3&Acc (H)\\
\hline
Machine reading comp.&MRPC&BERT-base&LayerDrop~\cite{fan2019reducing}&66M; 110M&85.3; 88.9&Acc (H)\\
\hline
NMT (en$\rightarrow$de) &WMT14&Multi-layer LSTM&Mag.~\cite{see2016compression}&43M; 216M&20.91; 20.5&BLEU (H)\\
\hline
NMT (en$\rightarrow$de) &WMT16&4-layer LSTM&Iter. Mag.~\cite{zhu2017prune}&23M; 211M&26.19; 26.77&BLEU (H)\\
\hline
NMT (en$\rightarrow$de) &WMT16 &Transformer&LayerDrop~\cite{fan2019reducing}&22M; 44M&29; 29&BLEU (H)\\
\hline
NMT (en$\rightarrow$de) &WMT17&Transformer&Iter. Mag.~\cite{cheong2019transformers}&22M; 44M&26.4; 28.09&BLEU (H)\\
\hline
Paraphrasing&QQP&BERT-large&RPP/Iter. Mag.~\cite{guo2019reweighted}&138M/170M; 340M&91.2/85.1; 91.2&Acc (H)\\
\hline
Question answering&SQuAD 1.1&BERT-large&RPP/Iter. Mag.~\cite{guo2019reweighted}&138M/170M; 340M&90.23/85.3; 90.9&Acc (H)\\
\hline
Question answering&SQuAD 2.0&BERT-large&RPP/Iter. Mag.~\cite{guo2019reweighted}&138M/170M; 340M&81.3/75.3; 81.9&Acc (H)\\
\hline
Sentiment analysis&SST-2&BERT-large&RPP/Iter. Mag.~\cite{guo2019reweighted}&138M/170M; 340M&92.4/91.3; 93.2&Acc (H)\\
\hline
Sentiment analysis&SST-2&BERT-base&LayerDrop~\cite{fan2019reducing}&66M; 110M&92.5; 93.5&Acc (H)\\
\hline
Speech recognition&2100 hours English Speech&2 CONV+7 BiRNNs+CTC&Iter. Mag.~\cite{narang2017exploring}&11.1M; 67M&10.59; 10.67&CER (L) on dev\\
\hline
Speech recognition&2100 hours English Speech&2 CONV+7 BiGRUs+CTC&Iter. Mag.~\cite{narang2017exploring}&17.8M; 115M&9.76; 9.55&CER (L) on dev\\
\hline
Speech recognition&2100 hours English speech&2 CONV+7 BiRNNs+CTC&Block Sparsity~\cite{narang2017block}&25.8M; 67M&15.66; 15.36&CER (L) on test\\
\hline
Speech recognition&2100 hours English speech&2 CONV+7 BiRNNs+CTC&Block Sparsity+Group Lasso~\cite{narang2017block}&12.9M; 67M&15.89; 15.36&CER (L) on test\\
\hline
Speech recognition&2100 hours English speech&2 CONV+3 BiGRUs+CTC&Block Sparsity~\cite{narang2017block}&25.6M; 115M&16.23; 15.42&CER (L) on test\\
\hline
Speech recognition&2100 hours English speech&2 CONV+3 BiGRUs+CTC&Block Sparsity+Group Lasso~\cite{narang2017block}&10.8M; 115M&16.78; 15.42&CER (L) on test\\
\hline
Speech recognition&AN4&2 CONV+3 HLSTMs+CTC&Grow and Prune~\cite{dai2018grow}&2.6M; 44.72M&10.37; 8.92&WER (L)\\
\hline
Speech recognition&Switchboard (swb/fsh)&7-layer MLP&Prune Neurons~\cite{he2014reshaping}&12.2M; 32.19M&25.5/28.8; 25.7/28.8&WER (L)\\
\hline
Speech recognition&TIMIT&5-layer MLP&Prune Neurons~\cite{he2014reshaping}&3.5M; 5.76M&20.7; 20.79&PER (L)\\
\hline
Speech recognition&TIMIT&LSTM&Iter. Mag.~\cite{cao2019efficient}&0.32M; 3.2M&23.5; 23.5&PER (L)\\
\hline
Speech recognition&TIMIT&LSTM&Block Sparsity~\cite{cao2019efficient}&0.32M; 3.2M&26.5; 23.5&PER (L)\\
\hline
Speech recognition&TIMIT&LSTM&BBS~\cite{cao2019efficient}&0.32M; 3.2M&23.5; 23.5&PER (L)\\
\hline
Speech recognition&WSJ 92&1 CONV+3 FC+1 BiRNN&DSD~\cite{han2016dsd}&4.07M; 8.14M&27.9; 27.45&WER (L)\\
\hline
Speech recognition&WSJ 92&2 CONV+7 BiRNNs+CTC&DSD~\cite{han2016dsd}&33.5M; 67M&10.65; 9.02&WER (L)\\
\hline
Speech recognition&WSJ 93&1 CONV+3 FC+1 BiRNN&DSD~\cite{han2016dsd}&4.07M; 8.14M&32.99; 31.6&WER (L)\\
\hline
Speech recognition&WSJ 93&2 CONV+7 BiRNNs+CTC&DSD~\cite{han2016dsd}&33.5M; 67M&14.84; 13.44&WER (L)\\
\hline
Summarization&CNN-Dailymail&Transformer&LayerDrop~\cite{fan2019reducing}&22M; 44M&39; 40&ROUGE (H)\\
\hline
Textual entailment&MNLI&BERT-large&RPP/Iter. Mag.~\cite{guo2019reweighted}&138M/170M; 340M&86.1/77; 86.1&Acc (H)\\
\hline
Textual entailment&MNLI-m&BERT-large&RPP/Iter. Mag.~\cite{guo2019reweighted}&138M/170M; 340M&85.7/82.5; 85.9&Acc (H)\\
\hline
Textual entailment&MNLI-m&BERT-base&LayerDrop~\cite{fan2019reducing}&66M; 110M&82.9; 84.6&Acc (H)\\
\hline
Textual entailment&QNLI&BERT-large&RPP/Iter. Mag.~\cite{guo2019reweighted}&138M/170M; 340M&92.3/90.2; 91.3&Acc (H)\\
\hline
Textual entailment&QNLI&BERT-base&LayerDrop~\cite{fan2019reducing}&66M; 110M&89.4; 90.5&Acc (H)\\
\hline
Textual entailment&RTE&BERT-large&RPP/Iter. Mag.~\cite{guo2019reweighted}&138M/170M; 340M&70.1/68.6; 70.1&Acc (H)\\
\hline
    \end{tabular}
    \caption{Comparison of various pruning methods (sorted by Task and then Dataset). CONV=Convolution. CTC=Connectionist temporal classification. FC=Fully connected. HLSTM=hidden-layer LSTM~\cite{dai2018grow}. In the metric column, H means high is better while L means low is better. PER/CER/WER=Phone/Character/Word error rate. For~\cite{murray2015auto}, embedding weights have not been considered when computing model size in the table. Block sparsity methods use block size of 4x4. BBS uses 64 banks.}
    \label{tab:pruningSummary}
\end{table}

\subsection{Summary}
Table~\ref{tab:pruningSummary} compares various pruning methods across different tasks and datasets. Size and accuracy of both the original and the pruned model are shown. The papers report multiple (model size, model accuracy) pairs but we carefully chose the pair such that accuracy is typically within 10\% of the original or the best pruned model accuracy reported. For language modeling, most popular datasets include PTB, Europarl v7 English, Wikitext-103 and AFP from English Gigaword. On PTB using LSTMs, we observe that Bank-Balanced Sparsity method~\cite{cao2019efficient} leads to lowest perplexity. For the Linguistic Acceptability CoLA task, RPP~\cite{guo2019reweighted} resulted into a smaller and a more accurate model compared to iterative magnitude pruning. As expected in some senses, the pruned model provides better accuracy than the unpruned one since pruning causes regularization. For the machine reading comprehension, question answering and paraphrasing tasks also, RPP seems to work better than iterative magnitude pruning. For the machine translation (NMT) task, on English-German WMT datasets, pruned Transformer models provide better accuracy than pruned LSTM with comparable number of parameters. For the sentiment analysis task, on SST-2, although RPP leads to a better pruned model compared to iterative pruning, LayerDrop~\cite{fan2019reducing} improves further on it with a model less than half of the RPP-pruned model. For the speech recognition task, experiments have been reported on 2100 hours English speech data, TIMIT, WSJ, Switchboard and AN4 datasets. On 2100 hours English speech data, Block Sparsity+Group Lasso is better than Block sparsity without regularization. Also, it is better than plain iterative magnitude pruning. On TIMIT, block sparsity~\cite{narang2017block} leads to a more accurate 90\% pruned LSTM model compared to  the unpruned one. For the summarization task, LayerDrop~\cite{fan2019reducing} can compress the model to half without any noticeable accuracy change. Finally, for the textual entailment task, experiments have been done on GLUE~\cite{wang2019glue} datasets: MNLI, MNLI-m, QNLI and RTE. Models pruned from BERT-large perform better than models pruned from BERT-base; RPP performs better than iterative magnitude pruning.

While older methods~\cite{lecun1990optimal,hassibi1993second} claimed that Hessian based methods were more effective than magnitude based pruning, almost all recent methods have been based on magnitude based pruning. See et al.~\cite{see2016compression} proposed three pruning schemes. They make the following observations: (1) Class-blind pruning outperforms both other schemes. Further, the overall performance loss is caused disproportionately by a few classes: softmax weights, source and target embedding weights. (2) It seems that higher layers are more important than lower layers, and that attention and softmax weights are crucial in LSTMs. (3) After retraining the pruned NMT models, baseline performance (20.48 BLEU) is both recovered and improved upon, up to 80\% pruning (20.91 BLEU), with only a small performance loss at 90\% pruning (20.13 BLEU). (4) In LSTMs, the parameters corresponding to the less common words are more dispensable. Weights connecting to the input are most crucial, followed by the input gate, then the output gate, then the forget gate. This is particularly true of the lower layers, which focus primarily on the input. However for higher layers, especially on the target side, weights connecting to the gates are as important as those connecting to the input. 

Narang et al.~\cite{narang2017exploring} observe that for approximately same number of parameters, gradual/iterative pruning is 7\% to 9\% better than hard pruning. They also conclude that the initial layers are pruned more aggressively compared to the final layers. Zhu et al.~\cite{zhu2017prune} advise that in order to get the best-performing sparse model of a certain size, we should train a dense model that is 5x-10x larger and then prune to the desired number of parameters rather than taking the largest and best-performing dense model and pruning this model by 20x or more to the desired number of parameters. Guo et al.~\cite{guo2019reweighted} find that RPP is much better than typical iterative pruning. In their experiments with BERT they find that for both original BERT and BERT pruned with RPP, the low-dimensional manifolds of the language representations are similar, showing the similar projection. This implies that the BERT applied with RPP keeps most of the language representation information similar to that from the original BERT.

For block pruning, Narang et al.~\cite{narang2017block} make the following observations: (1) We can create block-sparse RNNs with sparsity ranging from 80\% to 90\% with small loss in accuracy. This allows us to reduce the model size by roughly 10×. Block sparsity works with a variety of block sizes up to 32×32. (2) For block size 4×4, models with sparsity greater 90\% yield a relative accuracy loss of 30\% or higher. Similarly, for blocks of 16×16, models with sparsity greater than 86\% have 30\% or more accuracy loss. A similar trend is observed for block size 32×32. This indicates that there is a tradeoff between sparsity, block size and accuracy of the model. (3) For both block pruning and weight pruning, we see that the initial layers are pruned more aggressively compared to the final layers. Increasing sparsity in the layers closer to the output results in poor accuracy. Additionally, the variance in sparsity across the layers increases with the block size. Further, Cao et al.~\cite{cao2019efficient} make the following observations comparing block sparsity with BBS: (1) BBS achieves almost the same model accuracy regardless of the change of bank size. For block sparsity, however, increasing the block size adversely affects model accuracy. 

For pruning of attention heads, Michel et al.~\cite{michel2019sixteen} observe that one can prune up to 20\% and 40\% of heads from 6-layer NMT Transformer and BERT resp., without incurring any noticeable negative impact. When trying to remove a head at a time, only 8 (out of 96) heads in 6-layer NMT Transformer (16 heads/layer) cause a statistically significant change in performance when they are removed from the model, half of which actually result in a higher BLEU score. Further Voita et al.~\cite{voita2019analyzing} find that on the English-Russian WMT dataset, pruning 38 out of 48 encoder heads results in a drop of only 0.15 BLEU. 
 
Overall, to summarize, pruning has been the most popular method for model compression. Pruning methods can be unstructured (prune weights) or structured (prune neurons, blocks, attention heads, layers). While weight pruning theoretically leads to pruning to a large extent, practical implementation of sparse data structures is difficult. Pruning and regularization need to be done together carefully. Also, it is critical to define the importance functions for various structures carefully. Among weight pruning methods, while iterative magnitude pruning with regularization works well for RNNs and LSTMs, RPP performs better for Transformer based models. Pruning blocks using BBS is better than pruning neurons. For Transformer models, pruning just the heads do not provide much model compression, but dropping a combination of attention heads and layers is better.

\section{Quantization}
\label{sec:quantization}

\begin{figure}
    \centering
    \includegraphics[width=\columnwidth]{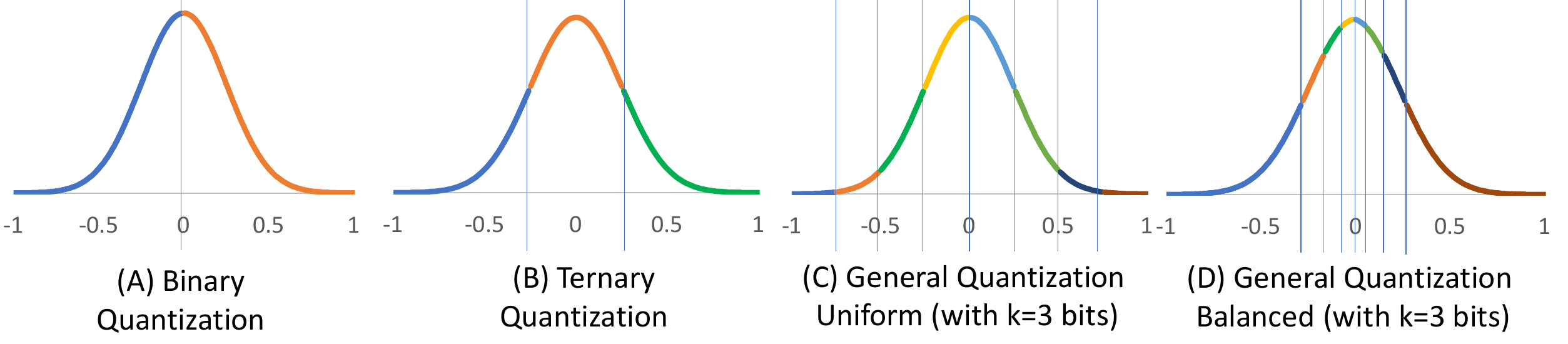}
    \caption{Different Types of Quantization: Binary (A),  Ternary (B) and General Quantized (C and D). Note that X axis denotes the weight value while the Y axis denotes frequency.}
    \label{fig:quantization}
\end{figure}

While pruning saves on the model size by removing weights, quantization aims to reduce the number of bits needed to store weights. Most computer architectures use 32 bits to represent weights. However, estimated precision of the brain (hippocampal spine) synapses is around 4.6 bits~\cite{bartol2015hippocampal}. Empirical evidence suggests that most quantities in the nervous system (for instance, firing of the neurons) have variability of a few percent due to biological noise, or a precision of 1 in 100 at best~\cite{linden2018think}. Thus, each decision could depend on $\log_2 (100)$=6.64 bits. Thus, we should be able to store weights in our artificial neural networks on average in a space of 4--7 bits. Given this motivation, various methods have been proposed which perform 1-bit (or binary quantization), ternary quantization, and general quantization exploring the spectrum between 3 and 32 bits. We discuss such methods in this section.  Fig.~\ref{fig:quantization} provides a broad overview of various quantization styles.

\subsection{Binarized Networks}
\label{subsec:binQuant}
Binarized networks use binary quantization (see Fig.~\ref{fig:quantization}(A)), which quantizes weights using 1 bit. Quantizing weights to 1 bit provides a compression of 32x but leads to a significant drop in accuracy across many tasks. However, in a hybrid quantization scheme, such binary quantization can be very helpful for some layers in a network. Binarization can be done using deterministic methods or could be stochastic in nature. Also, while na\"ive binarization has a very simple way of fixing the binary boundary threshold, one could perform a complex loss aware binarization as well. We discuss these variants of binarization in this section.

\subsubsection{Deterministic Binarization}
Simplest way of binary quantization is to set the weight as 1 for non-negative weights, and to -1 for negative weights. This leads to 32x compression. Also, the matrix multiplication for binary matrices is $\sim$7x faster~\cite{hubara2017quantized} leading to faster model inference. In the forward pass, binary networks drastically reduce memory size and accesses, and replace most arithmetic operations with bit-wise operations, which leads to great increases of power efficiency. Also, in the simplest version, binarization can be performed in a static manner, i.e., after the training is done. However, this method leads to large loss in accuracy. 

A variant of this simple method is to set the weight to a constant $c_1$ for non-negative weights, and to another constant $c_2$ for negative weights. Binary Scheme (BS)-Fixed method~\cite{lam2018word2bits} stores the original weights and during the forward pass replaces the values with a masked value of $c_1$ or $c_2$, where $c_1$ and $c_2$ are fixed and chosen with hyperparameter tuning. Full precision weights are used during training. At the end of training, the weights are replaced with the index of its masked value. Choosing the values of $c_1$ and $c_2$ can be difficult and time-consuming in BS-Fixed. Thus, in the BS-flexible method~\cite{cheong2019transformers}, we initialize $c_1$ and $c_2$ using KMeans with two centroids over the weights, and then update $c_1$ and $c_2$ using back-propagation. Also, in the BS-Flexible method, weights are quantized as follows.
\begin{equation}
    w_b= \begin{cases}
c_1 &\text{if } w \geq (c_1+c_2)/2\\
c_2 &\text{if } w < (c_1+c_2)/2
\end{cases}
\end{equation}

Note that $w$ is the original weight value while $w_b$ is the binarized weight value. These changes eliminate the need for hyper-parameter tuning.

\subsubsection{Stochastic Binarization}
Stochastic~\cite{courbariaux2015binaryconnect} binarization is performed as follows. 
\begin{equation}
    w_b= \begin{cases}
+1 &\text{with probability } p=\sigma(w)\\
-1 &\text{with probability } 1-p
\end{cases}
\end{equation}

\noindent where
\begin{eqnarray}
    \sigma(w)=\text{clip}\left(\frac{w+1}{2},0,1\right)=\max\left(0, \min\left(1, \frac{w+1}{2}\right)\right)
\end{eqnarray}

We only binarize the weights during the forward and backward propagations but not during the parameter update. Keeping good precision weights during the updates is necessary for Stochastic Gradient Descent (SGD). This is possible using something called as ``Straight Through Estimator (STE) trick''~\cite{bengio2013estimating}. As per STE, as the quantized value is an approximation of the original value, we can substitute the gradient with respect to the quantized value for the gradient of original value. The trick allows the inclusion of quantization into the computation graph of back-propagation and allows QNNs to represent parameters, activations and gradients with low bitwidth numbers. For test-time inference, there are three options using such a quantization method: 
\begin{itemize}
    \item Use the resulting binary weights $w_b$ (this makes most sense with the deterministic binarization). 
    \item In the stochastic case, many different networks can be sampled by sampling a $w_b$ for each weight. The ensemble output of these networks can then be obtained by averaging the outputs from individual networks. 
    \item Use original weights. But this does not reduce model size. 
\end{itemize}

Besides this, there have been further  efforts that make train/test faster but do not reduce model size. For example, Lin et al.~\cite{lin2015neural} convert multiplications in the backward pass into bit-shifts by restricting the activations to be power-of-two integers. Hubara et al.~\cite{hubara2016binarized} binarize weights and activations, at the inference phase and the entire training phase of a deep network.

\subsubsection{Loss Aware Binarization (LAB)}
\label{subsubsec:lossAwareBin}
The na\"ive binary quantization methods divide the real number line into two parts and each part was mapped to a quantized weight value. Can we decide per weight value which of the two weights it should be quantized to? Thus the idea behind Binary Weight Networks (BWN)~\cite{rastegari2016xnor} is to approximate the weight vector $W\in R^n$ using a binary vector $B\in\{+1,-1\}^n$ and a scaling factor $\alpha\in R^+$ such that $W\approx\alpha B$. This can be expressed as an optimization problem as follows.
\begin{eqnarray}
    \alpha^*, B^*=\argmin_{\alpha, B}||W-\alpha B||^2
\end{eqnarray}
We can expand and write this as follows.
\begin{eqnarray}
    ||W-\alpha B||^2=\alpha^2B^TB-2\alpha W^TB+W^TW
\end{eqnarray}
Since $B\in{+1,-1}^n$, $B^TB=n$. Also $W^TW$ is a constant. Thus $B^*=\argmax_B W^B$ such that $B\in{+1,-1}^n$. This optimization can be solved by simply assigning $B_i=+1$ when $W_i\geq 0$, and $B_i=-1$ otherwise. To compute $\alpha^*$, we set the derivative of $||W-\alpha B||^2$ wrt $\alpha$ to 0 and get the solution as follows.
\begin{eqnarray}
    \alpha^*=\frac{\sum|W_i|}{n}
\end{eqnarray}

Thus, besides the binarized weight matrix, a scaling parameter is also learned in BWN.

To take this idea further, can we learn $\alpha$ and $B$ to minimize the overall network's loss function? Thus, now, the Weight binarization can be formulated as the following optimization problem.
\begin{eqnarray}
    &&\min_{\hat{w}} \text{loss}(\hat{w})\\
    &&\text{such that } \hat{w}_l=\alpha_l b_l; \alpha_l>0; b_l\in\{+1,-1\}^{n_l}; l=1,..., L
\end{eqnarray}

\noindent where $L$ is the number of layers, $n_l$ is the number of weights in layer $l$. This loss aware binarization~\cite{hou2016loss} problem can be solved using proximal Newton algorithm~\cite{lee2014proximal} to find the best $\alpha_l$ and $B_l$.

\subsection{Ternarized Networks}
\label{subsec:ternaryQuant}
Unfortunately, binary quantization of the recurrent weights in RNNs/LSTMs never worked~\cite{ott2016recurrent}. When the true value of a weight is near zero, its quantized value is either set to -1 or 1. This results into an artificial increase in the magnitude of the weights and the vanishing/exploding gradients problem becomes more severe. Hence, another popular form of quantization is ternary quantization  (see Fig.~\ref{fig:quantization}(B)). Ternary quantization can help achieve a min of 16x compression (up to 32x compression if hardware allows to avoid storing zeros). In this section, we discuss different variants of ternary quantization from the simplest ternary connect networks to hybrid ternary networks like HitNets. 
\subsubsection{Ternary Weight Networks}
The simplest method for ternary quantization is ternary connect~\cite{lin2015neural} whose deterministic form is as follows.
\begin{equation}
    w_t= \begin{cases}
+1 &\text{if } w > 0.5\\
0 &\text{if } -0.5 < w \leq 0.5\\
-1 &\text{if } w \leq -0.5\\
\end{cases}
\end{equation}

Note that $w$ is the original weight value while $w_t$ is the ternarized weight value. Like binary connect, ternary connect also eliminates all multiplications in the forward pass. In the stochastic form, assuming original weights have been normalized to be in the range [-1,1], ternary quantization is done as follows.

\begin{equation}
    w_t= \begin{cases}
+1 &\text{with prob } w \text{ if } w\in (0,1]\\
0 &\text{with prob } 1-w \text{ if } w\in (0,1]\\
0 &\text{with prob } 1+w \text{ if } w\in[-1,0]\\
-1 &\text{with prob } -w \text{ if } w \in [-1,0]\\
\end{cases}
\end{equation}

A slightly related way called as Bernoulli Ternary Quantization where $w_t$ is set to +1 (or -1) with prob $p$ if $w>0$ (or) $<0$, and set to 0 with prob 1-p where $p\sim$Bernoulli($|x|$). Yet another way to set the boundaries for the three ranges is to use Gaussian based ternary weights~\cite{alom2018effective} as follows.
\begin{equation}
    w_t= \begin{cases}
+1 &\text{if } w > -(\mu+\sigma/2)\\
0 &\text{if } -(\mu+\sigma/2)<w\leq (\mu+\sigma/2)\\
-1 &\text{if } w \leq -(\mu+\sigma/2)\\
\end{cases}
\end{equation}
\noindent where $\mu$ and $\sigma$ are the mean and standard deviation of the weight matrix being quantized.

\subsubsection{Trained Ternary Quantization}
Rather than using the rules for ternary quantization as mentioned above, one can learn the boundary ranges or the quantized values for individual weights. One way of learning the right ternary representation per weight value is to minimize the Euclidean distance between full precision weights $W$ and the ternary weights $T$ along with a scaling factor~\cite{li2016ternary}. This can be expressed as the following optimization problem.
\begin{eqnarray}
\alpha^*, T^*=\argmin_{\alpha,T}||W-\alpha T||_2^2\\
\text{such that } \alpha\geq 0; T_i\in \{-1,0,1\}; i=1,2,..., n
\end{eqnarray}
Note that this is equivalent to the BWN method~\cite{rastegari2016xnor}. This does not lead to a closed form solution. Hence, we approximate the solution with threshold-based ternary function. 

\begin{equation}
    w_t= \begin{cases}
+1 &\text{if } w > \Delta\\
0 &\text{if } -\Delta < w \leq \Delta\\
-1 &\text{if } w \leq -\Delta\\
\end{cases}
\label{eq:eq2}
\end{equation}
The approximation works when we set $\Delta$ as follows.
\begin{eqnarray}
\Delta^*=\argmax_{\Delta>0}\frac{1}{|I_\Delta|}\left(\sum_{i\in I_\Delta}|W_i|\right)^2
\end{eqnarray}
\noindent where $I_\Delta$ is the number of weights with magnitude$>\Delta$. Again, this has no straightforward solution, unless we assume that original weights $W_i$'s are generated from uniform or normal distribution. When $W_i$'s are uniformly distributed in $[-a, a]$ and $\Delta$ lies in $(0, a]$, the approximated $\Delta^*$ is $a/3$, which equals to $\frac{2}{3}E(W)$. When $W_i$'s are generated from normal distributions $N(0,\sigma^2)$, the approximated $\Delta^*$ is 0.6$\sigma$ which equals to 0.75$E(|W|)$. Thus, we can use the following rule of thumb for fast and easy computation.
\begin{eqnarray}
\Delta^*\approx 0.7 E(W)=\frac{0.7}{n}\sum_{i=1}^n |W_i|
\end{eqnarray}

Another way to learn the quantization step size $\Delta$ in Eq.~\ref{eq:eq2} is to learn in a loss-aware manner~\cite{hwang2014fixed}, i.e., tuning it to minimize the overall network loss. Given a multi-layered network, we need to perform such quantization layer by layer in a greedy manner. We first train the network with full precision weights. We quantize all input data and signals of hidden layers. Next, we start with the weight quantizer between the input layer and the first hidden layer, try several step sizes around the initial step size and measure the output error of the network with the training set. The initial step size is determined using Lloyd-Max algorithm~\cite{lloyd1982least}. Choose the step size that minimizes the output error and quantize the weights. Further, we perform these steps for the next few layers until the output layer. Finally, the quantized neural network is retrained.

Yet another way of training ternary quantization~\cite{zhu2016trained} is to quantize weights to one of {$-W_l^n$, 0, $W_l^p$} for each layer $l$, where $W_l^n$ and $W_l^p$ are trainable parameters, learned using back-propagation. First, we normalize the full-precision weights to the range [-1, +1] by dividing each weight by the maximum weight. During SGD, we back propagate the gradient to both $W_l^n$ and $W_l^p$ and to the latent full-precision weights. This makes it possible to adjust the ternary assignment (i.e. which of the three values a weight is assigned). To decide the quantization step size $\Delta_l$ for a layer $l$, two heuristics can be used: (1) set $\Delta_l=t\times \max(|w_l|)$ where $t$ is a constant and $w_l$ are the full precision weights in layer $l$. (2) maintain a constant sparsity $r$ for all layers throughout training. By adjusting the hyper-parameter $r$ we can obtain ternary weight networks with various sparsities.

\subsubsection{Hybrid Ternary Quantization}
Given various ternary quantization methods proposed so far, one can combine them and use different methods for different layers. Wang et al.~\cite{wang2018hitnet} found that threshold ternary quantization (TTQ) (Eq.~\ref{eq:eq2}) is preferable for weights in an RNN while Bernoulli Ternary Quantization (BTQ) is preferable for activations. This is based on the observation that in an RNN, the distribution of weights follows normal distribution (with different ranges across different weight matrices), while for activations, the range is [0,1] and most of the values are located near to the two poles instead of the middle of the range. In the training phase (where we need to store the full precision weights), ternary quantization of weights only saves 1.4x memory consumption but quantizing both weights and activations can achieve up to 16x memory savings.

The HitNet architecture~\cite{wang2018hitnet} with this hybrid ternary quantization can be defined using these equations, where $i_t, f_t, o_t$ are the input, forget and output gates; $x_t$ is input at time $t$; $c_t$ is the cell output; and $h_t$ is the hidden layer output; $W_x$, $W_h$, $b_x$, $b_h$ are weights. 

\begin{eqnarray}
    \nonumber i_t,f_t,g_t,o_t&=&\sigma(\text{TTQ}(W_x)x_t+\text{TTQ}(b_x)\\
    \nonumber &+&\text{TTQ}(W_h)h_{t-1}+\text{TTQ}(b_h))\\
    \nonumber c_t&=&f_t\times c_{t-1}+i_t\times g_t\\
    h_t&=&\text{BTQ}(o_t\times \sigma(c_t))
\end{eqnarray}

\subsection{General Quantized Networks}
\label{subsec:generalQuant}
So far we discussed methods designed specifically for binary and ternary quantization. Now, we discuss general $k$-bit quantization methods. We will discuss (1) uniform quantization methods which perform equal width binning, (2) non-uniform methods which are closer to equal frequency binning, (3) loss-aware quantization methods, and (4) methods specifically designed for Transformer models.

\subsubsection{Uniform Quantization}
Uniform $k$-bit Quantization simply splits the range of original weights into $2^k-1$ equal size intervals~\cite{rastegari2016xnor, hubara2017quantized, he2016effective}. Refer Fig.~\ref{fig:quantization}(C).
If original weights are in range [-1,1], they can be quantized as follows.
\begin{equation}
    q_k(x)=2\left(\frac{\text{round}[(2^k-1)\left(\frac{x+1}{2}\right)]}{2^k-1}-\frac{1}{2}\right)
    \label{eq:uq}
\end{equation}
Similarly, if entries are in range [0,1], we could use the following formula.
\begin{eqnarray}
    q_k(x)=\frac{1}{2^k-1}\lfloor(2^k-1)x+\frac{1}{2}\rfloor
\end{eqnarray}
When the weights in matrix $X$ are not in the range [0,1], we can first scale weights as follows.
\begin{eqnarray}
   \tilde{X}=\frac{X-\beta}{\alpha}
\end{eqnarray}
\noindent where $\alpha=\max(X)-\min(X)$ and $\beta=\min(X)$. After quantization, we can apply a reverse transform to approximate the original values. Overall, the quantized result can be written as follows.
\begin{eqnarray}
   Q(X)=\alpha q_k(\tilde{X})+\beta
\end{eqnarray}

Given any quantization function $q_k(x)$, one can use it for quantizing weight matrices of various recurrent models like RNNs, GRUs and LSTMs~\cite{ott2016recurrent}. Typical inference equations for a GRU can be written as follows.
\begin{eqnarray}
    z_t=\sigma(W_z.[h_{t-1},x_t]); r_t=\sigma(W_r.[h_{t-1},x_t])\\
    \tilde{h_t}=\text{tanh}(W.[r_t\times h_{t-1},x_t]); h_t=(1-z_t)h_{t-1}+z_t \tilde{h_t}
\end{eqnarray}

Besides the matrix multiplications needed to compute $z_t$, $r_t$ and $\tilde{h_t}$, the gate structure of $\tilde{h_t}$ and $h_t$ brings in the need for element-wise multiplication. 
As $\tilde{h_t}$ and $h_t$ are also the inputs to computations at the next timestamp, and noting that a quantized value multiplied by a quantized value will have a larger bit-width, we need to insert additional quantization steps after element-wise multiplications.
Another problem with quantization of GRU structure lies in the different value range of gates. The range of tanh is [-1, 1], which is different from the value range [0, 1] of $z_t$ and $r_t$. Keeping in mind these observations, the equations for a quantized GRU can be written as follows, after the weights $W_z$, $W_r$ and $W$ and input $x_t$ have already been quantized to [-1,1].
\begin{eqnarray}
    z_t&=&\sigma(W_z.[h_{t-1},x_t])\\
    r_t&=&\sigma(W_r.[h_{t-1},x_t])\\
    \tilde{h_t}&=&tanh\left(W.\left[2q_k\left(\frac{1}{2}(r_t h_{t-1})+\frac{1}{2}\right)-1,x_t\right]\right)\\
    h_t&=&2q_k\left(\frac{1}{2}((1-z_t) h_{t-1}+z_t  \tilde{h_t})+\frac{1}{2}\right)-1
\end{eqnarray}

Following a similar method, we can also quantize LSTM networks.

\subsubsection{Balanced Quantization}
Uniform quantization is easy to implement but far from optimum when quantizing non-uniform data, which is believed to be the trained weights and activations of deep neural network. One way of performing non-uniform quantization is exponential quantization~\cite{ott2016recurrent}. It quantizes the weight values to an integer power of 2. If we let 
\begin{eqnarray}
    p=\frac{|W|}{2^{\lfloor \log_2|W|\rfloor}}-1
\end{eqnarray}
deterministic exponential quantization can be written as follows.
\begin{equation}
    \log_2 W_q= \begin{cases}
\lceil \log_2|W| \rceil &\text{if } p>0.5\\
\lfloor \log_2|W| \rfloor  &\text{otherwise }
\end{cases}
\end{equation}
Similarly, stochastic exponential quantization can be written as follows.
\begin{equation}
    \log_2 W_q= \begin{cases}
\lceil \log_2|W| \rceil &\text{with prob } p\\
\lfloor \log_2|W| \rfloor  &\text{with prob } 1-p
\end{cases}
\end{equation}

Exponential quantization enables storing weights in low precision and eliminating multiplications. However, it still does not perform quantization in a way which is sensitive to the distribution of the weights. Distributions of parameters in neural networks are often imbalanced, such that the uniform quantization determined from extremal values may under utilize available bitwidth. When we quantize values, it may be desirable to make the quantized values have balanced distributions, to take full advantage of the available parameter space. Balanced quantization method~\cite{zhou2017balanced} starts by partitioning numbers into $2^{k}$ bins containing roughly the same number of entries (percentiles).  Refer Fig.~\ref{fig:quantization}(D). Each partition is then mapped to an evenly-divided interval in the closed interval [0, 1]. Finally, the quantization step maps intervals into discrete values using Eq.~\ref{eq:uq} and transforms the value range to be approximately the same as input.

A na\"ive implementation using percentiles as thresholds would require sorting of weight values during each forward operation in back-propagation, which may slow down the training process. The $2^{k}$ evenly spaced percentiles required in histogram equalization can be computed from the recursive application of partitioning of numbers by medians. Further, the mean $\mu$ can be used to approximate the median $m$. Thus, we can perform approximate histogram equalization without doing sorting. 

\subsubsection{KMeans based Quantization Schemes}
Yet another way of performing non-uniform quantization is to decide bin boundaries using clustering in a static manner. In this static-KMeans method~\cite{muller2015rounding}, We first train the neural network with full-precision parameters. Then apply KMeans to the weights. After clustering, the value of each pixel is set to the value of the center of the cluster it belongs to. We also need to store mapping from integers to cluster centers. Given $k$ clusters, we only need $\log (k)$ bits to code the clusters.

A better approach is to perform KMeans clustering during training. In this method~\cite{han2015deep}, multiple connections (belonging to the same cluster) share the same weight, and we fine-tune those shared weights. For the forward pass, the cluster index stored for each connection is mapped to a centroid which is then used as the weight. For back-propagation, during update, all the gradients are grouped by the cluster index and summed together, multiplied by the learning rate and subtracted from the shared centroids from last iteration. We use KMeans clustering to identify the shared weights for each layer of a trained network, so that all the weights that fall into the same cluster will share the same weight. Weights are not shared across layers. To calculate the compression rate, given $k$ clusters, we only need $\log_2 k$ bits to encode the index. In general, for a network with $n$ connections and each connection is represented with $b$ bits, constraining the connections to have only $k$ shared weights will result in a compression rate of
\begin{eqnarray}
 r=\frac{nb}{n\log_2 k+kb}   
\end{eqnarray}

There are two other ways of using KMeans for non-uniform quantization: Product Quantization (PQ) and Residual Quantization (RQ)~\cite{gong2014compressing}. In product quantization (PQ), we partition the vector space into many disjoint subspaces, and perform quantization (KMeans) in each subspace. Weight matrix $W$ is partitioned columnwise: $W=[W^1, W^2, ..., W^s]$ where $W^i\in R^{m\times n/s}$ assuming $n$ is divisible by $s$. Then we perform KMeans on each submatrix $W^i$ to obtain clusters $c^i_1, ..., c^i_k$. Thus, we get $s$ codebooks. The reconstructed matrix is $\hat{W}=[\hat{W}^1, \hat{W}^2, ..., \hat{W}^s]$ where $\hat{W}^i_j$ is the closest centroid $c^i_j$. PQ can be applied to either the x-axis or the y-axis of the matrix. We need to store the cluster indexes and codebooks for each subvector. The compression rate for this method can be written as follows.
\begin{eqnarray}
 r=\frac{32mn}{32kn + log_2(k)ms} 
\end{eqnarray}
Residual quantization (RQ) is similar. In RQ, we first quantize the vectors into k-centers. Next we find out the residuals for each data point ($w-c$) and perform KMeans on the residuals. Do it recursively $t$ times.
Then the resultant weight vectors are calculated as $\hat{W}_z=c^1_j+c^2_j+...+c^t_j$ given we have recursively performed $t$ iterations. We need to store all the codebooks for each iteration, which potentially needs large amount of memory. The compression rate for this method can be written as follows.
\begin{eqnarray}
 r=\frac{m}{tk + log_2(k)tn}   
\end{eqnarray}

\subsubsection{Loss Aware Quantization (LAQ)}
Generalizing the loss aware binarization approach (Sec.~\ref{subsubsec:lossAwareBin})~\cite{rastegari2016xnor}, we can perform $k$-bit quantization~\cite{guo2017network} by attempting to solve the following problem. 
\begin{eqnarray}
 \min_{\{\alpha_i,b_i\}_{i=1}^k} \left|\left|w-\sum_{i=1}^k \alpha_i b_i\right|\right|^2   
\end{eqnarray}
where $w\in R^n$ is the original weight vector, $\alpha_i\in R$ and $b_i\in\{-1,+1\}^n$ are variables to be learned. This NP-hard problem can be solved using an iterative greedy approximation which sequentially minimizes the residue. In each iteration, first the residue is computed as 
\begin{eqnarray}
r_{i-1}=w-\sum_{j=1}^{i-1}\alpha_j b_j,
\end{eqnarray}
and then $\alpha_i$ and $b_i$ are computed as $\alpha_i=\frac{1}{n}||r_{i-1}||_1$ and $b_i=\text{sign}(r_{i-1})$. Further, refined greedy approximation~\cite{guo2017network} extends this to further decrease the quantization error. In the $j^{th}$ iteration after $\alpha_j$ and $b_j$ have been updated, the method adds one extra step to refine all computed $\{\alpha_i\}_{i=1}^j$ with the least squares solution as follows.
\begin{eqnarray}
[\alpha_1, ..., \alpha_j]=((B_j^TB_j)^{-1}B_j^Tw)^T
\end{eqnarray}
where $B_j=[b_1,...,b_j]$. Typically refined greedy is more accurate than the greedy approach. In refined greedy approximation, after modification on the computed $\alpha$'s, $b$'s are no longer optimal while the method keeps all of them fixed. To improve the refined greedy approximation, alternating minimizing $\alpha$'s and $b$'s becomes a natural choice. Xu et al.~\cite{xu2018alternating} find that only two alternating cycles is good enough to find high precision quantization. Further, similar to~\cite{rastegari2016xnor}, for an LSTM, we can combine overall network loss minimization with the multi-bit quantization loss minimization using this bi-level optimization.
\begin{eqnarray}
\min_{w,\{\alpha_i, b_i\}_{i=1}^k} \text{LSTM}\left(\sum_{i=1}^k\alpha_i b_i\right)\\
\text{such that }\{\alpha_i, b_i\}_{i=1}^k=\argmin_{\{\alpha_i',b_i'\}_{i=1}^k}\left|\left|w-\sum_{i=1}^k \alpha_i' b_i'\right|\right|^2
\end{eqnarray}

\subsubsection{Quantization for Word Embeddings and Transformers}
Each word vector is typically represented as a 300--500 dimensional vector, with each parameter being 32 bits. As there are millions of words, word vectors may take up to 3--6 GB of memory/storage. Can we quantize word vectors? We can clearly quantize them after training. But, we could also quantize when learning word embeddings. For example, Lam et al.~\cite{lam2018word2bits} perform 1-bit and 2-bit quantization while performing word2vec~\cite{mikolov2013word2vec} training using the Continuous Bag of Words (CBOW) method. They observe that quantization while training leads to better results compared to quantization after training. 

Cheong et al.~\cite{cheong2019transformers} applied BS-Fixed and BS-Flexible binary quantization to Transformer models. They observed that the Transformer architecture is highly resistant to quantization, and is able to match the original model up to a 4-bit representation. Simple iterative pruning is much worse compared to quantization. Lastly, Shen et al.~\cite{shen2019q} propose mixed-precision quantization for BERT based on the observation that  different encoder layers should use different number of bits for quantization. Layers that exhibit flatter curvature of the loss gradient surface can be quantized to lower bit precision. Thus, they use different number of bits at different levels of granularity: layers, attention heads and groups of neurons. They observe that quantizing embedding layers with 8 bits and other weight matrices with 2--4 bits leads to results comparable with full-precision BERT.

\subsection{Summary}

\begin{table}
    \centering
    \scriptsize
    \begin{tabular}{|l|l|l|l|l|l|l|l|}
    \hline
Task&Dataset&Model&Method&Eval&Bits (weights)&Bits (activation)&Metric\\
\hline
\hline
Language modeling&IMDB&GRU&Uniform Q.~\cite{he2016effective}&0.882; 0.905&4&4&Acc (H)\\
\hline
Language modeling&Linux Kernel&LSTM&BinaryConnect~\cite{courbariaux2015binaryconnect}&3.532; 1.329&1&FP&CE (L)\\
\hline
Language modeling&Linux Kernel&LSTM&Loss Aware B.~\cite{hou2016loss}&1.305/1.409; 1.329&1&FP/1&CE (L)\\
\hline
Language modeling&Linux Kernel&LSTM&BNN~\cite{hubara2017quantized}&3.624; 1.329&1&1&CE (L)\\
\hline
Language modeling&Linux Kernel&LSTM&BWN~\cite{rastegari2016xnor}&1.307; 1.329&1&FP&CE (L)\\
\hline
Language modeling&PTB&GRU&Uniform Q.~\cite{he2016effective}&102; 100&4&4&PPW (L)\\
\hline
Language modeling&PTB&GRU&Balanced Q.~\cite{zhou2017balanced}&116; 100&4&4&PPW (L)\\
\hline
Language modeling&PTB&LSTM&Greedy~\cite{guo2017network}&118.9; 89.8&2&FP&PPW (L)\\
\hline
Language modeling&PTB&LSTM&Refined Loss Aware~\cite{guo2017network}&95.6; 89.8&2&3&PPW (L)\\
\hline
Language modeling&PTB&LSTM&Uniform Q.~\cite{he2016effective}&152/114; 109&2/4&2/4&PPW (L)\\
\hline
Language modeling&PTB&LSTM&BNN~\cite{hubara2017quantized}&100; 97&4&4&PPW (L)\\
\hline
Language modeling&PTB&LSTM&HitNet~\cite{wang2018hitnet}&110.3; 97.2&2&2&PPW (L)\\
\hline
Language modeling&PTB&LSTM&Alternating LAQ~\cite{xu2018alternating}&103.1/91.4; 89.8&2/4&FP&PPW (L)\\
\hline
Language modeling&PTB&LSTM&Alternating LAQ~\cite{xu2018alternating}&91.9; 89.8&2&3&PPW (L)\\
\hline
Language modeling&PTB&LSTM&Balanced Q.~\cite{zhou2017balanced}&126/123; 106&2&2/3&PPW (L)\\
\hline
Language modeling&Text-8&LSTM&Refined Loss Aware~\cite{guo2017network}&122.3; 101.1&2&3&PPW (L)\\
\hline
Language modeling&Text-8&LSTM&HitNet~\cite{wang2018hitnet}&169.1; 151.4&2&2&PPW (L)\\
\hline
Language modeling&Text-8&LSTM&Alternating LAQ~\cite{xu2018alternating}&105.1; 101.1&2&3&PPW (L)\\
\hline
Language modeling&Text-8&RNN&Exponential Q.~\cite{ott2016recurrent}&1.639; 1.588&2&FP&BPC (L)\\
\hline
Language modeling&War and Peace&LSTM&BinaryConnect~\cite{courbariaux2015binaryconnect}&2.942; 1.268&1&FP&CE (L)\\
\hline
Language modeling&War and Peace&LSTM&Loss Aware B.~\cite{hou2016loss}&1.291/1.376; 1.268&1&FP/1&CE (L)\\
\hline
Language modeling&War and Peace&LSTM&BNN~\cite{hubara2017quantized}&3.05; 1.268&1&1&CE (L)\\
\hline
Language modeling&War and Peace&LSTM&BWN~\cite{rastegari2016xnor}&1.313; 1.268&1&FP&CE (L)\\
\hline
Language modeling&Wikidata-2&LSTM&HitNet~\cite{wang2018hitnet}&126.72; 114.37&2&2&PPW (L)\\
\hline
Language modeling&WikiText-2&LSTM&Refined Loss Aware~\cite{guo2017network}&105.8; 114.37&2&3&PPW (L)\\
\hline
Language modeling&WikiText-2&LSTM&Alternating LAQ~\cite{xu2018alternating}&102.7; 100.1&2&3&PPW (L)\\
\hline
Named Entity Recognition&CoNLL-03&BERT-base&QBERT~\cite{shen2019q}&91.06; 95&2w8e&8&F1 (H)\\
\hline
Named Entity Recognition&CoNLL-03&BERT-base&Mixed-precision Q.~\cite{shen2019q}&94.37; 95&2-3w8e&8&F1 (H)\\
\hline
NMT (en$\rightarrow$de)&WMT17&Transformer&BS-Fixed/BS-Flexible~\cite{cheong2019transformers}&11.61/12.11; 28.09&1&FP&BLEU (H)\\
\hline
NMT (en$\rightarrow$de)&WMT17&Transformer&K-Means 1/4-bit Q.~\cite{cheong2019transformers}&12.07/27.65; 28.09&1&FP&BLEU (H)\\
\hline
NMT (en$\rightarrow$de)&WMT17&Transformer&K-Means 1-bit att-Q.~\cite{cheong2019transformers}&24.96; 28.09&1&FP&BLEU (H)\\
\hline
NMT (en$\rightarrow$de)&WMT17&Transformer&BS-Flexible 1-bit att-Q.~\cite{cheong2019transformers}&25.54; 28.09&1&FP&BLEU (H)\\
\hline
Question answering&SQuAD&BERT-base&QBERT~\cite{shen2019q}&79.6; 88.69&2w8e&8&F1 (H)\\
\hline
Question answering&SQuAD&BERT-base&Mixed-precision Q.~\cite{shen2019q}&86.95; 88.69&2-3w8e&8&F1 (H)\\
\hline
Question answering&SQuAD&Facebook’s DrQA&BS-Fixed~\cite{lam2018word2bits}&77.04; 75.28&2&FP&F1 (H)\\
\hline
Sentiment analysis&IMDB&LSTM&Gaussian Q.~\cite{alom2018effective}&79.64; 82.87&1&FP&Acc (H)\\
\hline
Sentiment analysis&IMDB&LSTM&Gaussian T./B.~\cite{alom2018effective}&76.86/76.25; 82.87&2&FP&Acc (H)\\
\hline
Sentiment analysis&SST-2&BERT-base&QBERT~\cite{shen2019q}&84.63; 93&2w8e&8&Acc (H)\\
\hline
Sentiment analysis&SST-2&BERT-base&Mixed-precision Q.~\cite{shen2019q}&92.08; 93&2-3w8e&8&Acc (H)\\
\hline
Speech recognition&TIDIGITS&GRU&Pow2 T.~\cite{ott2016recurrent}&99.18; 99.1&2&FP&Acc (H)\\
\hline
Speech recognition&TIMIT&4-layer MLP&Loss Aware T.~\cite{hwang2014fixed}&29.97/28.35; 26.24&1/2&1&FER (L)\\
\hline
Speech recognition&WSJ&4-layer BiLSTM&Pow2 T.~\cite{ott2016recurrent}&10.49; 11.16&2&FP&WER (L)\\
\hline
Textual entailment&MNLI&BERT-base&QBERT~\cite{shen2019q}&77.02; 84.4&2w8e&8&Acc (H)\\
\hline
Textual entailment&MNLI&BERT-base&Mixed-precision Q.~\cite{shen2019q}&82.29; 84.4&2-3w8e&8&Acc (H)\\
\hline
Textual entailment&MNLI-m&BERT-base&QBERT~\cite{shen2019q}&76.56; 84&2w8e&8&Acc (H)\\
\hline
Textual entailment&MNLI-m&BERT-base&Mixed-precision Q.~\cite{shen2019q}&81.75; 84&2-3w8e&8&Acc (H)\\
\hline
Word similarity&M. Turk &Word embeddings&BS-Fixed~\cite{lam2018word2bits}&0.602; 0.617&2&FP&CHR (H)\\
\hline
Word similarity&MEN &Word embeddings&BS-Fixed~\cite{lam2018word2bits}&0.764; 0.745&2&FP&CHR (H)\\
\hline
Word similarity&Rare Words &Word embeddings&BS-Fixed~\cite{lam2018word2bits}&0.362; 0.4&2&FP&CHR (H)\\
\hline
Word similarity&SimLex&Word embeddings&BS-Fixed~\cite{lam2018word2bits}&0.387; 0.358&2&FP&CHR (H)\\
\hline
Word similarity&WordSim Relatedness &Word embeddings&BS-Fixed~\cite{lam2018word2bits}&0.594; 0.529&2&FP&CHR (H)\\
\hline
Word similarity&WordSim Similarity&Word embeddings&BS-Fixed~\cite{lam2018word2bits}&0.752; 0.741&2&FP&CHR (H)\\
\hline
    \end{tabular}
\caption{Comparison of various quantization methods (sorted by Task and then Dataset). Q.=Quantization, B.=Binarization, T.=Ternarization, PPW=Perplexity per word, BPC=Bits per character, CE=cross-entropy, FER=frame error rate, CHR=correlation with human rankings. FP=full precision (32 bits). For~\cite{lam2018word2bits}, we report results with word embedding dimensions set to 1000. In the metric column, H means high is better while L means low is better. For  quantization of the BERT-base model~\cite{shen2019q}, we report number of bits used for encoders as well as for embeddings. `2-3w8e' means 2 or 3 bits were used for encoder weights while 8 bits were used for embeddings. For NMT results by Cheong et al.~\cite{cheong2019transformers}, ``att-Q'' means only attention layers were quantized.}
    \label{tab:quantizationSummary}
\end{table}

Ott et al.~\cite{ott2016recurrent} observed that the weight binarization methods do not work with RNNs. Hubara et al.~\cite{hubara2017quantized} were the first to attempt to quantize both weights and activations by trying to evaluate the accuracy of quantized recurrent models trained on the Penn Treebank dataset. Similar to~\cite{ott2016recurrent}, Hubara et al.~\cite{hubara2017quantized} found that binarization of weight matrices lead to large accuracy degradation. Later techniques like the one by Xu et al.~\cite{xu2018alternating} with 2 bits for weights and 3 bits for activations showed better results.

Table~\ref{tab:quantizationSummary} compares various quantization methods across different tasks and datasets. Accuracy of both the original and the quantized model are shown. Also, we report number of bits used for weights (which indicate the model size) as well as activations. For the same task, dataset and model combination, different papers report different accuracy of the full precision model because of slight changes in training hyper-parameters; hence we report accuracy of full precision model for each row. 

For language modeling, PTB, Text-8, WikiText-2, Linux Kernel, IMDB and ``War and Peace'' are the popular datasets. Across all the datasets, loss aware binarization outperforms other weight binarization schemes. On the Linux Kernel dataset, it is even better than the full-precision network. BinaryConnect does not work well here because of the problem of exploding gradients. On PTB, Xu et al.'s Alternating LAQ~\cite{xu2018alternating} with 2 bits for weights and 3 bits for activations leads to an LSTM which is just 2.1 points worse in terms of perplexity per word. By 3-bit quantization, Alternating LAQ can achieve $\sim$10.5x memory saving and $\sim$3× real inference acceleration. Uniform and Balanced quantization are rule-based and not aim at minimizing the error. Balanced quantization proposed by Zhou et al.~\cite{zhou2017balanced} performs better than HitNet~\cite{wang2018hitnet} and uniform quantization~\cite{he2016effective}. Balanced quantization leads to better results compared to unbalanced counterparts, especially when quantizing to 2-bit weights. However, for 4-bit weights, there is no clear gap between scaling by mean and scaling by max (i.e. balanced and unbalanced quantization).

Across multiple tasks like named entity recognition with CoNLL-03, question answering with SQuAD,  sentiment analysis with SST-2, textual entailment using MNLI, Shen et al.~\cite{shen2019q} show that mixed precision quantization (where different number of bits are used for different groups of neurons -- 128 groups in each layer) of BERT is better than QBERT. The reason behind this is the discrepancy (in QBERT) that not all the layers have the same sensitivity to quantization. For more sensitive layers, higher bit precision needs to be set, while for layers that are less sensitive, 2-bits setting is already sufficient. With only additional 5MB memory storage, 2/3-bits mixed-precision Q-BERT is able to retain the performance drop within 2.3\% for MNLI, SQuAD and 1.1\% for SST-2, CoNLL-03, with up to 13x compression ratio in weights. For machine translation, Cheong et al.~\cite{cheong2019transformers} observe that Transformer architecture is highly resistant to quantization (unlike to pruning), and are, in essence, able to match the original model up to a 4-bit representation. Binary Scheme Flexible outperforms 1-bit k-means in both pure 1-bit compression and quantizing only the attention layers, suggesting not tying the weights to particular centroids improves performance, and outperform Binary Scheme Fixed, indicating learning the values to be a superior method. After binarizing only the attention layers, we are still able to recover 90\% of the model's performance. Lam et al.~\cite{lam2018word2bits} experimented with quantization of word embeddings and showed that 2-bit quantized word vectors outperform full precision vectors on word similarity tasks, but do worse on word analogy tasks. Intuitively, they reason that full precision Word2Vec is prone to overfitting with increased epochs of training; quantized training does not seem to suffer as much from this. For sentiment analysis on IMDB, Alom et al.~\cite{alom2018effective} show that quantization to 4 bits is better than to 3 or 2 bits, which is expected. They also show that the normal distribution shows better performance against uniform distribution with quantized weights. For speech recognition, Ott et al.~\cite{ott2016recurrent} show that pow-2 ternarization is the best. 

Cheong et al~\cite{cheong2019transformers} were the first to quantize Transformers. They observed that in the last attention layer of the decoder over the encoder hidden states, the attention distribution of the original and 4-bit model are highly similar, indicating 4 bit weights, i.e weights that take on one of 16 values, is enough to get the full effects of attention. Attention distributions in the encoder layers of the Transformer for the original and 4-bit models are almost indistinguishable from one another. This again highlights the idea that self-attention is highly resistant to quantization and could be heavily compressed. Later Shen et al.~\cite{shen2019q} showed that comparable performance to full precision BERT can be achieved with at most 2.3\% performance degradation across many tasks, even with ultra-low precision quantization down to 2 bits, corresponding up to 13x compression of the model parameters, and up to 4x compression of the embedding table as well as activations. 

Overall, quantization performs model compression by reducing the number of bits per weight value. Binary quantization does not work well by itself for text based neural models. But ternary and higher-bit quantization lead to significant model size reduction without loss in accuracy across tasks. One consideration for quantization is that 3-bit quantized execution is typically not supported in hardware. It is however possible to load 3-bit quantized values and cast them to higher bit precision such as 4 or 8 bits in the execution units. This would still have the benefit of reduced memory volume to/from DRAM. Non-uniform quantization methods like balanced quantization or KMeans based quantization methods are better than uniform quantization methods. Loss aware quantization done while training is better than static loss-unaware quantization. Mixed-precision quantization combined with pruning is highly effective for Transformer based models. 

\section{Knowledge Distillation (KD)}
\label{sec:knowledgeDistillation}

KD methods are the most popular model compression methods for Transformer networks. Also called student-teacher networks, the main idea is to first train a deep teacher network, and then learn a shallow student network that mimics the teacher. After training, the student model is deployed. What information (``dark knowledge'') from the teacher can be used to train the student? What loss functions can be used to ensure right flow of information from teacher to student? Can we have an ensemble of teachers, or teacher assistants or rather fellow students who can train the student? 
We discuss these aspects in this section.

\subsection{Various Distillation Architectures}
\begin{figure}
    \centering
    \includegraphics[width=\textwidth]{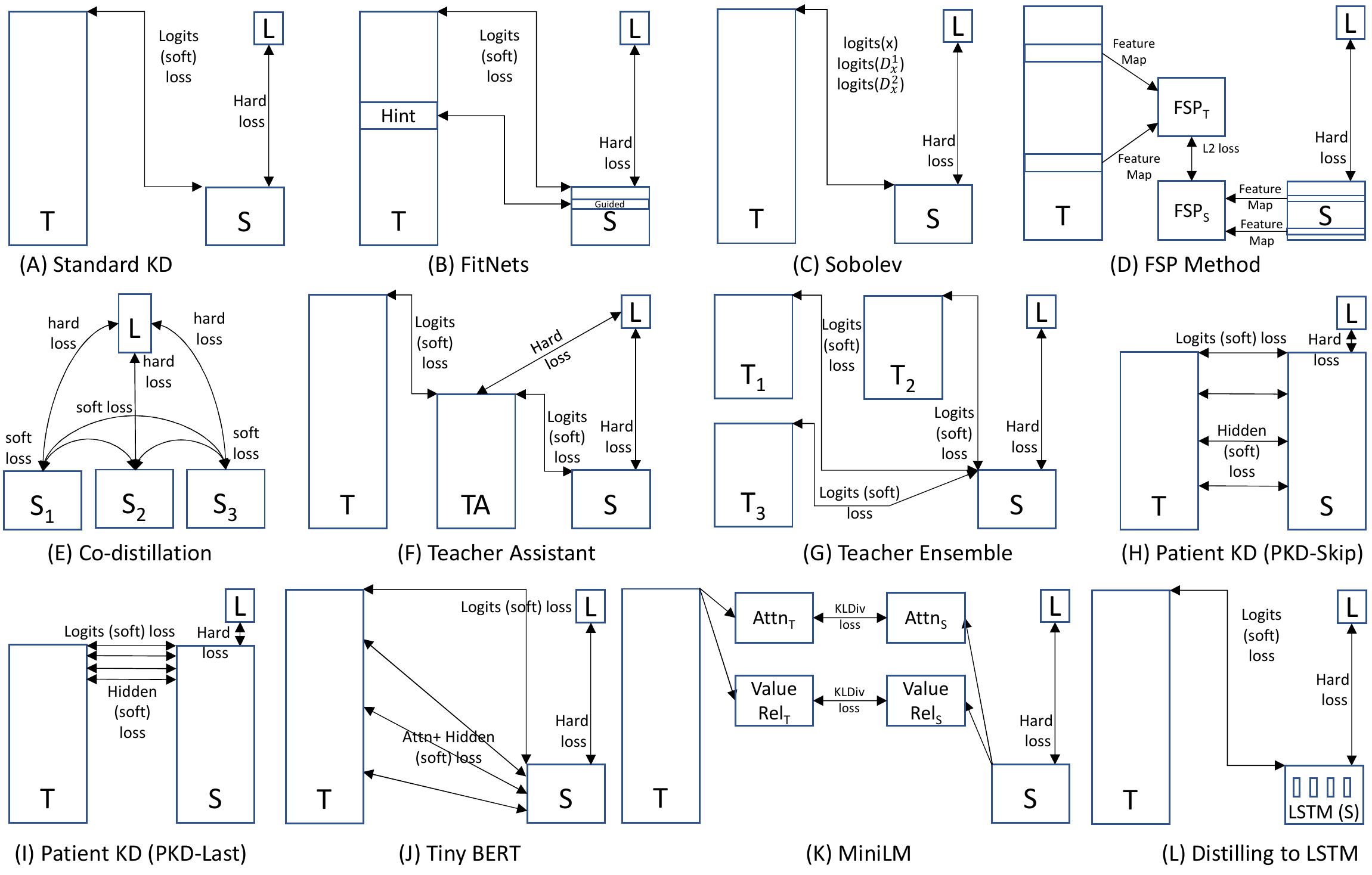}
        \caption{Different Types of Knowledge Distillation methods.}
    \label{fig:kd}
\end{figure}

Ba and Caruna~\cite{ba2014deep} proposed Student Teacher networks (or mimic models) where the student uses the logits before softmax from the teacher network for training (see Fig.~\ref{fig:kd}(A)). The student model is not trained on the original labels; it is trained to learn the function that was learned by the teacher model. Thus, the student model is optimized to minimize the L2 loss between the teacher logits and the student logits across all training instances. Such distilled student models are more accurate than the same shallow student trained directly on the original labeled training data mainly because: (1) Teacher removes noisy labels, if any. (2) The uncertainty from the teacher is more informative to the student than the original 0/1 labels. (3) The original targets may depend in part on features not available as inputs for learning, but the student sees targets that depend only on the input features. The dependence on unavailable features has been eliminated by filtering targets through the teacher.

Yet another way of utilizing logits is to have the student learn from noisy teacher logits~\cite{sau2016deep}. After obtaining logits from the teacher, Gaussian noise with mean 0 and standard deviation $\sigma$ is added to teacher’s logits. This perturbation can be applied to samples selected with probability $\alpha$. The perturbed outputs produce the effect of a regularizer. 

While Ba and Caruna~\cite{ba2014deep} suggested using only the logits, Hinton et al.~\cite{hinton2015distilling} suggested training the student by minimizing the cross entropy loss between the teacher softmax output and the student softmax output, besides minimizing the cross entropy between student prediction and actual label. The first part is called the soft loss and the second one is called the hard loss. Typically hard loss is given much lower weight compared to the soft loss term. To make the softmax output non-peaked and thereby transfer more useful information from teacher to student, softmax with temperature $>$1 should be used. The same temperature should be used for training both the teacher and the student, but after the student has been trained the temperature can be set to 1 at test time. Besides logits and softmax output, Sobolev training for neural networks is a method for incorporating target derivatives in addition to the target values while training student network  (see Fig.~\ref{fig:kd}(C)). Czarnecki et al.~\cite{czarnecki2017sobolev} experiment with first two derivatives of the targets.

KD has also been used along with quantization for better model compression~\cite{bengio2013estimating,mishra2017apprentice,polino2018model}. We start with a trained full-precision large teacher network and an apprentice (student) network that has been initialised with full-precision weights. The apprentice network's precision is lowered and is fine-tuned using KD. 

Why just use the output from the last layer of the teacher for training the student? In FitNets~\cite{romero2014fitnets}, the student performs hint-based training, i.e., the student is trained using not only the outputs but also the intermediate representations learned by the teacher as hints to improve the training process and final performance of the student  (see Fig.~\ref{fig:kd}(B)). we choose a hidden layer of the FitNet, the guided layer, to learn from the teacher's hint layer. Because the student intermediate hidden layer will generally be smaller than the teacher's intermediate hidden layer, additional parameters are introduced to map the student hidden layer to the prediction of the teacher hidden layer. 

While the methods discussed so far use logits, softmax output or their derivatives to transfer knowledge, Yim et al.~\cite{yim2017gift} proposed a ``flow of solution procedure (FSP)'' method where the distilled knowledge is transferred in terms of flow between layers, which is calculated by computing the inner product between features from two layers  (see Fig.~\ref{fig:kd}(D)). What does this ``flow'' capture intuitively? If we view the input of a deep network as the question and the output as the answer, we can think of the generated features at the middle of the network as the intermediate result in the solution process. There are many ways to solve the problem of generating the output from the input. Hence, mimicking the generated features of the teacher can be a hard constraint for the student. Learning the solution process from teacher is important. More concretely, the student is trained to minimize the L2 difference between the teacher and student FSP matrices computed across various pairs of layers and across multiple training instances. A similar method called Representational distance learning (RDL) has also been proposed in~\cite{mcclure2016representational}.

Lastly, multiple KD variants have been proposed for sequence-level predictions~\cite{kim2016sequence,freitag2017ensemble}, e.g., for neural machine translation (NMT). In word-level KD, cross-entropy is minimized between the student/teacher distributions for each word in the actual target sequence, as well as between the student distribution and the degenerate data distribution, which has all of its probability mass on one word. In sequence-level KD (Seq-KD) the student network is trained on the output from beam search of the teacher network that had the highest score. In sequence-level interpolation (Seq-Inter) the student is trained on the output from beam search of the teacher network that had the highest similarity (say using BLEU score) with the target sequence.

\subsection{Collaborative Learning}
Can multiple students learn from each other? Is a powerful teacher really required? In the deep mutual learning (DML) method~\cite{zhang2018deep}, different from the one-way transfer between a static pre-defined teacher and a student in model distillation, with DML, an ensemble of students learn collaboratively and teach each other throughout the training process. Surprisingly, no prior powerful teacher network is necessary -- mutual learning of a collection of simple student networks works, and moreover outperforms distillation from a more powerful yet static teacher. Specifically, each student is trained with two losses: a conventional supervised learning loss, and a mimicry loss that aligns each student's class posterior with the class probabilities of other students. 

Anil et al.~\cite{anil2018large} propose a similar method but suggest letting the students learn independently just using the conventional supervised learning (hard) loss at least for a few burn in iterations  (see Fig.~\ref{fig:kd}(E)). After this, the mutual learning can be done as in DML. They also propose a variant of their Co-Distillation method to perform this training in a distributed scenario where communication efficiency is also important. To update the parameters of one network using co-distillation one only needs the predictions of the other networks, which can be computed locally from copies of the other networks' weights. Empirically, using stale predictions instead of up-to-date predictions for the other neural networks has little to no adverse effect on the quality of the final trained model produced by co-distillation.

\subsection{Multiple Teachers}
So far we have talked about a student mimicing a single teacher. However, it is interesting to explore if the student can learn better in presence of multiple teachers or from a teacher assistant.

Intuitively and also observed empirically, student network performance degrades when the gap between student and teacher is large. Given a fixed student network, one cannot employ an arbitrarily large teacher, or in other words, a teacher can effectively transfer its knowledge to students up to a certain size, not smaller. To alleviate this shortcoming, Mirzadeh et al.~\cite{mirzadeh2019improved} introduced multi-step KD, which employs an intermediate-sized network (teacher assistant) to bridge the gap between the student and the teacher  (see Fig.~\ref{fig:kd}(F)).
The teacher assistant (TA) models are distilled from the teacher, and the student is then only distilled from the TAs. One could also perform multi-step TA distillation, for example, distillation path could be $10 \rightarrow 6  \rightarrow 4  \rightarrow 2$.

A simple way to do KD with multiple teachers is to train student with cross entropy loss between student  predictions and average prediction from multiple teachers  (see Fig.~\ref{fig:kd}(G)). A more effective method is to augment this with a relative dissimilarity (RD) loss~\cite{you2017learning} defined over intermediate layer outputs generated for a triplet of instances between the student and an ensemble of teachers. For the student, the middle layer is selected. For each teacher, we select the layer such that most teachers are consistent with the resulting order relationships under the voting strategy. We discuss the RD loss given a student and a teacher. Consider a triplet of instances ($x_i$, $x_i^+$, $x_i^-$) such that at an intermediate layer of the teacher network, distance between activations for $x_i^+$ and $x_i$ is smaller than the distance between activations for $x_i^-$ and $x_i$. Let $p_i$ be the intermediate output from student for example $x_i$. Then the RD loss for the triplet ($x_i$, $x_i^+$, $x_i^-$) can be written as follows. 
\begin{eqnarray}
\text{Loss}=\max(0,d(p_i, p_i^+)-d(p_i, p_i^-)+\delta)   
\end{eqnarray}
where $d$ is the distance function, and $\delta$ is a small constant to prevent the trivial solution. To extend this loss function definition to multiple teachers, the order between the instances $x_i^+$ and $x_i^-$ given $x_i$ is decided based on majority voting between the teachers. 

There are also specific settings when distilling from multiple teachers becomes natural, e.g., when the number of classes is large~\cite{hinton2015distilling} or in multi-lingual settings~\cite{tan2019multilingual}. When the number of classes is very large, the teacher model could be an ensemble that contains one generalist model trained on all the data and many ``specialist'' models, each of which is trained on data that is highly enriched in examples from a very confusable subset of the classes (like different types of mushroom). Softmax distribution vector of this type of specialist can be made much smaller by combining all of the classes it does not care about into a single dustbin class. Each specialist model is initialized with the weights of the generalist model. These weights are then slightly modified by training the specialist with half its examples coming from its special subset and half sampled at random from the remainder of the training set. To derive groupings of object categories for the specialists, we focus on categories that the full generalist network often confuses. When training the student, for each instance, we first find the $set $k$ of n$ most probable classes according to the generalist model. Then, we take all the specialist models, $m$, whose special subset of confusable classes has a non-empty intersection with $k$ and call this the active set of specialists $A_k$. Given student's full probability distribution $q$ over all the classes, we minimize the following.
\begin{eqnarray}
\text{Loss}=KL(p^g,q)+\sum_{m\in A_k} KL(p^m, q)   
\end{eqnarray}
where $p^g$ is output distribution from the generalist model, and $p^m$ is the output distribution from the $m^{th}$ specialist model.

An ensemble of teachers is also very useful in a multi-lingual NMT setting~\cite{tan2019multilingual}. Individual models for each language pair are first trained and regarded as teachers, and then the multilingual model is trained to fit the training data and match the outputs of individual models simultaneously through KD. When the accuracy of multilingual model surpasses the individual model for the accuracy threshold $\tau$ on a certain language pair, we remove the distillation loss and just train the model with original negative log-likelihood loss for this pair. Lastly, when learning from a teacher ensemble, it is burdensome to load all the teacher models in the GPU memory for distillation. Alternatively, we first generate the output probability distribution of each teacher model for each instance offline, and then just load the top-$K$ probabilities of the distribution into memory and normalize them so that they sum to 1 for distillation. This reduces the memory cost from the scale of $|V|$ (the vocabulary size) to $K$.

\subsection{Distilling Transformers}

Recently, there has been a lot of work around distilling Transformers to smaller Transformers with less number of layers or to Bidirectional LSTMs. Some of these methods aim at improving the accuracy versus model size tradeoff, while others focus on complex settings like mismatching student-teacher vocabulary~\cite{zhao2019extreme} or mismatch number of attention heads. 

Zhao et al.~\cite{zhao2019extreme} learn a student with small vocabulary compared to the teacher using a dual training method. During distillation, for a given training sequence input to the teacher model, they mix the teacher and student vocabularies by randomly selecting tokens from the sequence to segment using the student vocabulary, with the other tokens segmented using the teacher vocabulary. As part of the masked language model (MLM) task, the model now needs to learn to predict words from the student vocabulary using context words segmented using the teacher vocabulary, and vice versa. The expectation is that the student embeddings can be learned effectively this way from the teacher embeddings as well as teacher model parameters. We perform dual training only for the teacher model inputs. The student model receives words segmented exclusively using the student vocabulary. Also, during MLM, the model uses different softmax layers for the teacher and the student vocabularies depending on which one was used to segment the word in question. Instead of distilling solely on the teacher model's final-layer outputs, layer-wise teacher model parameters can also be leveraged to directly optimize parameters of corresponding layers in the student model.

In Patient KD (PKD)~\cite{sun2019patient}, the student learns from the teacher's output after every $k$ layers (see Fig.~\ref{fig:kd}(H)) or the output from the last few layers of the teacher (see Fig.~\ref{fig:kd}(I)). The student BERT is initialized using some layers of the pre-trained teacher BERT. TinyBERT~\cite{jiao2019tinybert} further extends this idea by using extensive knowledge from embedding layer, and attention and hidden sub-layers of multiple teacher layers, and also the overall teacher output (see Fig.~\ref{fig:kd}(J)). Each student layer is first mapped to a teacher layer before the student training. Liu et al.~\cite{liu2019improving} distill a multi-task student from a multi-task teacher, given the soft targets of the training data across multiple tasks. If task $t$ has a teacher, the task-specific loss is the average of two objective functions, one for the correct targets and the other for the soft targets assigned by the teacher. In MiniLM~\cite{wang2020minilm}, the student is trained by deeply mimicking the self-attention behavior of the last Transformer layer of the teacher (see Fig.~\ref{fig:kd}(K)). Besides self-attention distributions, MiniLM introduces the self-attention value-relation transfer to help the student achieve a deeper mimicry. The value-relation is computed as pairwise correlation between different components of the value matrix across various attention heads of the final layer.  Pretrained Distillation~\cite{turc2019well} pretrains the student model with a self-supervised masked LM objective on a
large corpus first, then performs a standard KD on supervised tasks.  

Most of these models learn one-to-one layer mapping, where each student layer is guided by only one specific teacher layer. Li et al.~\cite{li2020bert} propose a method where each student intermediate layer learns from every teacher intermediate layer with learnable attention weights. Both the embedding-layer distillation and the prediction-layer distillation employ the one-to-one layer mapping as in TinyBERT and BERT-PKD.

Tang et al.~\cite{tang2019distilling} propose distillation of a BERT model to a single layer BiLSTM using KL divergence between student and teacher logits (see Fig.~\ref{fig:kd}(L)). Mukherjee et al.~\cite{mukherjee2020xtremedistil} also distill a multi-lingual BERT (mBERT) model to a BiLSTM. Representation transfer is done from Transformer-based teacher model to BiLSTM-based student model with different embedding dimensions and disparate output spaces. Distillation features include teacher logits and internal teacher representations for one teacher layer. To make all output spaces compatible, a non-linear projection of the parameters in student representation is done to have same shape as teacher representation for each token. The projection parameters are learned by minimizing the KL-divergence (KLD) between the representations of the student and the chosen layer from the teacher. Overall there are multiple loss functions for the student training: supervised hard loss, soft loss wrt output logits, and soft loss wrt internal teacher layer. Rather than optimizing for all loss functions jointly, stage-wise training is performed where each loss function is sequentially used for optimization. 

Lastly, there have been recent efforts~\cite{sun2020mobilebert,iandola2020squeezebert} to distill Transformer models to slightly modified Transformer architectures. MobileBERT~\cite{sun2020mobilebert} is a thin version of BERT-large, while equipped with bottleneck structures and a carefully designed balance between self-attentions and feed-forward networks (FFN). To train MobileBERT, we first train a specially designed teacher model, an inverted-bottleneck incorporated BERT-large model (IB-BERT). The IB-BERT uses the inverted-bottleneck structures to adjust its feature map size to 512. Thus, in the bottleneck structure, the inputs to the multi-head attention (MHA) module are from wider feature maps (of inter-block size), while the inputs to the FFN are from narrower bottlenecks (of intra-block size). To fix this issue, MobileBERT uses stacked feed-forward networks to re-balance the relative size between MHA and FFN. Each MobileBERT layer contains one MHA but 4 stacked FFN after each MHA. Then, we conduct knowledge transfer from this teacher to MobileBERT using feature map transfer and attention transfer across all layers. Also, while distilling they perform progressive knowledge transfer, i.e., they progressively train each layer in $L$ stages, where $L$ is the number of layers. When the $l$-th layer is trained, all the trainable parameters in the layers below are frozen. Another difference in the MobileBERT student architecture is usage of high information flow residual connections between the high-channel-count layers. MobileBERT is 4.3x smaller and 5.5x faster than BERT-base while achieving competitive results on well-known benchmarks. Iandola et al.~\cite{iandola2020squeezebert} propose a new Transformer model architecture called SqueezeBERT which is much like BERT-base, but with the position-wise fully connected layers implemented as convolutions, and grouped convolutions for many of the layers. Distillation for SqueezeBERT is rather straightforward. Distillation is applied only to the final layer, and only during finetuning using soft cross entropy loss with respect to a weighted sum of the teacher's logits and a one-hot encoding of the ground-truth. Teacher model is a BERT-base model finetuned independently on each GLUE task, and these task-specific teacher weights are used for distillation. Xu et al.~\cite{xu2020bert} propose a method for progressive module replacement for compressing BERT. Their approach first divides the original BERT into several modules and builds their compact substitutes. Then, the original modules are randomly replaced with their substitutes to train the compact modules to mimic the behavior of
the original modules. We progressively increase the probability of replacement through
the training. In this way, their approach brings
a deeper level of interaction between the original and compact models.

\subsection{Summary}

\begin{table}
    \centering
    \scriptsize
    \begin{tabular}{|l|l|l|l|l|l|p{0.5in}|}
    \hline
Task&Dataset&Teacher/Student models&Method&Metric&Size (Distilled; Orig)&Eval. (Distilled; Orig)\\
\hline
\hline
Abs. summarization&CNN/DailyMail&UniLM-large/12-layer BERT&MiniLM~\cite{wang2020minilm}&ROUGE-L (H)&33M; 340M&39.73; 40.34\\
\hline
Ad CTR prediction&Criteo&No teacher/DNN&CoDistillation~\cite{anil2018large}&MAE (L)&-&0.019; 0.022\\
\hline
Cross-lingual NLI&XNLI (15 langs.)&XLM-R-base/12-layer BERT&MiniLM~\cite{wang2020minilm}&Acc (H)&21M; 85M&71.1; 74.5\\
\hline
Cross-lingual QA&MLQA (7 langs.)&XLM-R-base/12-layer BERT&MiniLM~\cite{wang2020minilm}&F1 (H)&21M; 85M&63.2; 64.9\\
\hline
Intent detection&SNIPS (7-class)&BERT-base/6-layer BERT&Mixed-vocabulary training~\cite{zhao2019extreme}&Acc (H)&6.2M; 109M&98.7; 98.6\\
\hline
Language modeling&Common Crawl&No teacher/2-layer LSTM&CoDistillation~\cite{anil2018large}&CE (L)&-&3.91; 3.92\\
\hline
Machine reading comp.&RACE&BERT-base/6-layer BERT&Patient KD~\cite{sun2019patient}&Acc (H)&67M; 109M&60.34; 65.30\\
\hline
Machine reading comp.&RACE&BERT-base/6-layer BERT&Vanilla KD~\cite{sun2019patient}&Acc (H)&67M; 109M&58.74; 65.30\\
\hline
NER (41 langs.)&Wikiann-41&Multiple mBERT/BiLSTM&XtremeDistill~\cite{mukherjee2020xtremedistil}&F1 (H)&31.8M; 179M*41&88.64; 90.76\\
\hline
NMT (de$\rightarrow$en)&OpenNMT&4-layer LSTM/2-layer LSTM&Vanilla KD~\cite{polino2018model}&BLEU (H)&64.8M; 84.8M&15.48; 15.88\\
\hline
NMT (de$\rightarrow$en)&OpenNMT&4-layer LSTM/2-layer LSTM&Quantized Distillation (4 bits)~\cite{polino2018model}&BLEU (H)&64.8M; 84.8M&15.26; 15.88\\
\hline
NMT (de$\rightarrow$en)&WMT13&4-layer LSTM/2-layer LSTM&Quantized Distillation (4 bits)~\cite{polino2018model}&BLEU (H)&81.6M; 84.8M&35.32; 34.7\\
\hline
NMT (de$\rightarrow$en)&WMT13&4-layer LSTM/2-layer LSTM&Vanilla KD~\cite{polino2018model}&BLEU (H)&81.6M; 84.8M&30.21; 34.7\\
\hline
NMT (en$\rightarrow$Others)&IWSLT (13 langs.)&Multiple Transformers/Transformer&Multiple teachers~\cite{tan2019multilingual}&BLEU (H)&44M; 44M*12&22.96; 22.72\\
\hline
NMT (en$\rightarrow$Others)&IWSLT (13 langs.)&Multiple Transformers/Transformer&Seq-KD with Multiple teachers~\cite{kim2016sequence}&BLEU (H)&44M; 44M*12&21.98; 22.72\\
\hline
NMT (en$\rightarrow$Others)&WMT (7 langs.)&Multiple Transformers/Transformer&Multiple teachers~\cite{tan2019multilingual}&BLEU (H)&44M; 44M*6&24.47; 24.50\\
\hline
NMT (en$\rightarrow$de)&WMT14&4-layer LSTM/2-layer LSTM&Word-KD~\cite{kim2016sequence}&BLEU (H)&49M; 221M&14.9; 17.7\\
\hline
NMT (en$\rightarrow$de)&WMT14&4-layer LSTM/2-layer LSTM&Seq-KD~\cite{kim2016sequence}&BLEU (H)&49M; 221M&18.1; 17.7\\
\hline
NMT (en$\rightarrow$de)&WMT14&4-layer LSTM/2-layer LSTM&Seq-KD + Seq-Inter + Word-KD~\cite{kim2016sequence}&BLEU (H)&49M; 221M&18.5; 17.7\\
\hline
NMT (en$\rightarrow$de)&WMT14&4-layer LSTM/2-layer LSTM&Pruned Seq-KD + Seq-Inter~\cite{kim2016sequence}&BLEU@5 (H)&8M/17M; 221M&18.5/19.1; 19.5\\
\hline
NMT (Others$\rightarrow$en)&IWSLT (13 langs.)&Multiple Transformers/Transformer&Multiple teachers~\cite{tan2019multilingual}&BLEU (H)&44M; 44M*12&30.34; 29.7\\
\hline
NMT (Others$\rightarrow$en)&Ted Talk (45 langs.)&Multiple Transformers/Transformer&Multiple teachers~\cite{tan2019multilingual}&BLEU (H)&44M; 44M*43&28.95; 25.17\\
\hline
NMT (Others$\rightarrow$en)&WMT (7 langs.)&Multiple Transformers/Transformer&Multiple teachers~\cite{tan2019multilingual}&BLEU (H)&44M; 44M*6&28.61; 27.07\\
\hline
NMT (th$\rightarrow$en)&IWSLT15&4-layer LSTM/2-layer LSTM&Word-KD~\cite{kim2016sequence}&BLEU (H)&8M; 47M&11.8; 14.3\\
\hline
NMT (th$\rightarrow$en)&IWSLT15&4-layer LSTM/2-layer LSTM&Seq-KD~\cite{kim2016sequence}&BLEU (H)&8M; 47M&12.8; 14.3\\
\hline
NMT (th$\rightarrow$en)&IWSLT15&4-layer LSTM/2-layer LSTM&Seq-KD + Seq-Inter + Word-KD~\cite{kim2016sequence}&BLEU (H)&8M; 47M&14.2; 14.3\\
\hline
Question generation&SQuAD 1.1&UniLM-large/12-layer BERT&MiniLM~\cite{wang2020minilm}&BLEU@4 (H)&33M; 340M&23.27; 24.32\\
\hline
Slot filling&SNIPS (39 slots)&BERT-base/6-layer BERT&Mixed-vocabulary training~\cite{zhao2019extreme}&F1 (H)&6.2M; 109M&95.0; 97.0\\
\hline
    \end{tabular}
\caption{Comparison of various knowledge distillation methods (sorted by Task and then Dataset). CE=cross entropy, MAE=mean absolute error. en=English, th=Thai. MRC=Machine Reading Comprehension. NLI=Natural Language Inference. QA=Question Answering. NER=Named Entity Recognition. In the metric column, H means high is better while L means low is better.}
    \label{tab:kdSummary1}
\end{table}

Table~\ref{tab:kdSummary1} compares various knowledge distillation methods across different tasks and datasets. Accuracy of both the original and the distilled model are shown in the Eval column. Also, we report model size for both the student as well as the teacher models. Note that sometimes smaller student models perform better than the teacher models. This could be because (1) for some (task, dataset) pairs, the smaller models are a better regularized fit compared to potentially overfitted teacher models, and (2) student models can be rigorously trained using additional semi-supervision while teacher models depend on limited labeled training data.

While knowledge distillation has been used for distilling NLP models across many applications, NMT is the most popular one. For abstractive summarization, MiniLM~\cite{wang2020minilm} leads to a student models which is less than one tenth of the teacher without much loss in ROUGE-L. Similarly, MiniLM shows good results for cross-lingual NLI and multi-lingual QA as well. For Ad click through rate (CTR) prediction and language modeling, Anil et al.~\cite{anil2018large} show that co-distillation leads to lower MAE and cross entropy respectively compared to the individually trained models. Zhao et al.~\cite{zhao2019extreme}'s mixed-vocab training leads to 6-layer model that retains over 95\% of the BERT-base model’s slot filling F1 score while being 30x smaller ($<$10 MB without quantization) and 57x faster on a mobile device, yet task-agnostic. For Named Entity Recognition (NER), Mukherjee et al.~\cite{mukherjee2020xtremedistil} show that XtremeDistil leads to massive compression of teacher models like mBERT by upto 35x in terms of parameters and 51x in terms of latency for batch inference while retaining 95\% of its F1-score for NER over 41 languages. 

For NMT, experiments have been done on OpenNMT, WMT, IWSLT and TedTalk datasets. Kim et al.~\cite{kim2016sequence} make the following observations: (1) Sequence-level knowledge distillation (Seq-KD) does better than word-level knowledge distillation (Word-KD) on English $\rightarrow$ German and performs similarly on Thai $\rightarrow$ English. (2) Combining them (Seq-KD + Word-KD) results in further gains, indicating that these methods provide orthogonal means of transferring knowledge from the teacher to the student: Word-KD is transferring knowledge at the the local (i.e. word) level while Seq-KD is transferring knowledge at the global (i.e. sequence) level. (3) Applying weight pruning on top of knowledge distillation results in a student model that has 13x fewer parameters than the original teacher model, with a decrease of 0.4 BLEU. Tan et al.~\cite{tan2019multilingual} show that one model is enough to handle multiple languages (up to 44 languages), with comparable or even better accuracy than individual models. Their method achieves larger improvements on some languages, such as Da, Et, Fi, Hi and Hy, than others. This is correlated with the data size of the languages. When a language is of smaller data size, it may get more improvement due to the benefit of multilingual training.

\begin{table}
    \centering
    \scriptsize
    \begin{tabular}{|p{0.3in}|l|l|l|l|l|l|l|l|l|l|l|l|l|l|}
\hline
Method&Method&Teacher/Student&Size&MRPC&MNLI&MNLI-m&SST-2&QQP&QNLI&RTE&CoLA&STS-B&SQuAD&SQuAD\\ 
Category&& models&& F1& Acc& Acc& Acc&F1& Acc& Acc& MCC&Spearman& 1.1 F1& 2.0 F1\\
\hline
\hline
\multirow{3}{0.3in}{Original Models}&BERT-B~\cite{devlin2018bert}&-& 109M&88.9&83.4&84.6&93.5&71.2&90.5&66.4&52.1&85.8&88.4&77.7\\
\cline{2-15}
&BERT-L~\cite{devlin2018bert}&-& 340M&89.3&85.9&86.7&94.9&72.1&92.7&70.1&60.5&86.5&90.9&81.9\\
\cline{2-15}
&MTDNN (ensemble)~\cite{liu2019multi}&-& 340*4M&90.0&87.2&86.7&95.6&72.4&93.9&80.9&61.5&88.3&-&-\\
\hline
\hline
\multirow{19}{*}{\rotatebox{90}{Knowledge Distillation Methods}}&Distilled-BiLSTM~\cite{tang2019distilling}&BERT-L/BiLSTM&0.96M&-&72.6&73.0&90.7&68.2&-&-&-&-&-&-\\
\cline{2-15}
&Mixed-vocab. training~\cite{zhao2019extreme}&BERT-L/BERT-12&10.9M&87.2&80.5&80.7&90.6&-&-&-&-&-&-&-\\
\cline{2-15}
&TinyBERT~\cite{jiao2019tinybert}&BERT-B/BERT-4&14.5M&86.4&81.8&82.5&92.6&71.3&87.7&66.6&44.1&80.4&82.1&71.8\\
\cline{2-15}
&BERT-EMD~\cite{li2020bert}&BERT-B/BERT-4&14.5M&87.6&80.6&82.1&91.0&69.3&87.2&66.2&25.6&82.3&-&-\\
\cline{2-15}
&MobileBERT~\cite{sun2020mobilebert}&BERT-L/BERT-6&25.3M&88.8&82.6&83.3&92.8&70.2&90.6&66.2&50.5&84.4&90.0&79.2\\
\cline{2-15}
&MobileBERT+Quantization~\cite{sun2020mobilebert}&BERT-L/BERT-6&25.3M&87.0&-&83.9&91.9&-&90.8&-&-&-&90.0&-\\
\cline{2-15}
&MiniLM~\cite{wang2020minilm}&BERT-B/BERT-12&33M&89.5&-&85.7&93.0&91.3&91.5&73.3&58.5&-&-&81.7\\
\cline{2-15}
&SqueezeBERT~\cite{iandola2020squeezebert}&BERT-B/SqueezeBERT&51.1M&87.8&81.1&82.0&91.4&80.3&90.1&73.2&46.5&86.7&-&-\\
\cline{2-15}
&DistilBERT~\cite{sanh2019distilbert}&BERT-B/BERT-4&52.2M&82.4&78.0&78.9&91.4&68.5&85.2&54.1&32.8&76.1&81.2&64.1\\
\cline{2-15}
&Patient KD~\cite{sun2019patient}&BERT-B/BERT-4&52.2M&82.6&79.3&79.9&89.4&70.2&85.1&62.3&24.8&79.8&79.5&64.6\\
\cline{2-15}
&BERT-of-Theseus~\cite{xu2020bert}&BERT-B/BERT-6&66M&87.6&82.1&82.4&92.2&71.6&89.6&66.2&47.8&84.1&-&-\\
\cline{2-15}
&MiniLM~\cite{wang2020minilm}&BERT-B/BERT-6&66M&88.4&-&84.0&92.0&91.0&91.0&71.5&49.2&-&-&76.4\\
\cline{2-15}
&BERT-EMD~\cite{li2020bert}&BERT-B/BERT-6&66M&89.8&83.5&84.7&93.3&72.0&90.7&71.7&47.5&86.8&-&-\\
\cline{2-15}
&Patient KD~\cite{sun2019patient}&BERT-B/BERT-6&67M&85.0&81.0&81.5&92.0&70.7&89.0&65.5&43.5&81.6&85.3&69.8\\
\cline{2-15}
&Vanilla KD~\cite{hinton2015distilling}&BERT-B/BERT-6&67M&86.2&79.8&80.2&91.5&70.1&88.3&64.7&-&-&-&-\\
\cline{2-15}
&DistilBERT~\cite{sanh2019distilbert}&BERT-B/BERT-6&67M&86.9&81.3&82.6&92.5&70.1&88.9&58.4&49.0&81.3&86.2&69.5\\
\cline{2-15}
&TinyBERT~\cite{jiao2019tinybert}&BERT-B/BERT-6&67M&87.3&83.2&84.6&93.1&71.6&90.4&70.0&51.1&83.7&87.5&77.7\\
\cline{2-15}
&Pretrained Distillation~\cite{turc2019well}&BERT-B/BERT-6&67M&86.8&82.2&82.8&91.8&70.4&88.9&65.3&-&-&-&-\\
\cline{2-15}
&MTDNN-KD~\cite{liu2019improving}&MTDNN/MTDNN-KD&340M&91.1&86.7&87.5&95.6&72.7&96.0&85.1&65.4&89.6&-&-\\
\hline
\hline
Param.&ALBERT-B~\cite{lan2019albert} (dev)&-&12M&-&-&81.6&90.3&-&-&-&-&-&89.3&80\\
\cline{2-15}
Sharing&ALBERT-L~\cite{lan2019albert} (dev)&-&18M&-&-&83.5&91.7&-&-&-&-&-&90.6&82.3\\
\hline
Tensor Decomp.&FLOP~\cite{wang2019structured}&-&80M&88.61&-&-&92.09&-&89.05&-&-&88.18&-&-\\
\hline
\multirow{3}{*}{Pruning}&RPP Iterative Magnitude Pruning~\cite{guo2019reweighted}&-&138M&88.1&86.1&85.7&92.4&91.2&92.3&70.1&82.8&-&90.23&75.3\\
\cline{2-15}
&Iterative Magnitude Pruning~\cite{guo2019reweighted}&-&170M&83.5&77&82.5&91.3&85.1&90.2&68.6&76.3&-&85.3&-\\
\cline{2-15}
&LayerDrop~\cite{fan2019reducing}&-&66M&85.3&-&82.9&92.5&-&89.4&-&-&-&-&-\\
\hline
Quant.&Mixed-precision quant. (QBERTMP)~\cite{shen2019q}&-&-&-&82.29&81.75&92.08&-&-&-&-&-&86.95&-\\
\hline
SubQuad Trans.&Linformer~\cite{wang2020linformer}&-&-&-&-&-&93.1&90.8&91.2&-&-&-&-&-\\
\hline
    \end{tabular}
\caption{Comparison of various methods  across various GLUE~\cite{wang2019glue} and SQuAD tasks. Please refer~\cite{wang2019glue} for detailed description of tasks. Top part (first three rows) shows results for basic Transformer methods (teacher models). Middle part shows results for knowledge distillation methods. Bottom part shows results for a mix of other methods across categories. BERT-L=BERT-large, BERT-B=BERT-base, BERT-$i$=$i$-layer BERT. MCC refers to Matthews correlation. Results for SQuAD are on dev set. Empty entries indicate that the papers do not report those results, or NA entries.}
    \label{tab:kdSummary2}
\end{table}

Table~\ref{tab:kdSummary2} compares various knowledge distillation methods (besides other model compression methods) across various GLUE~\cite{wang2019glue} and SQuAD tasks. Also, we report model size for both the student as well as the teacher models. Different distillation methods use one of these as the teacher: BERT-large, BERT-base or MTDNN. PKD~\cite{sun2019patient} outperforms the Vanilla KD~\cite{hinton2015distilling} on almost all the datasets except for MRPC. TinyBERT-4 significantly outperforms the 4-layer BERT-PKD and DistilBERT by a margin of at least 4.4\%, with ~28\% parameters and 3.1x inference speedup. Compared with the teacher BERT-base, 4-layer TinyBERT is 7.5x smaller and 9.4x faster in the model efficiency, while maintaining competitive performances. The 6-layer TinyBERT achieves comparable results with the teacher. Overall, TinyBERT consistently outperforms both the 4-layer and 6-layer baselines like PKD, DistilBERT and MiniLM.

Turc et al.~\cite{turc2019well} show how appropriate pretraining can improve the quality of distillation. MiniLM outperforms DistilBERT and TinyBERT across most tasks. The 6-layer MiniLM is 2.0x faster than original BERTBASE, while retaining more than 99\% performance on a variety of tasks, such as SQuAD 2.0 and MNLI. Distilled MT-DNN significantly outperforms the original MT-DNN on 7 out of 9 GLUE tasks. 4-layer BERT-EMD outperforms 4-layer DistilBERT and BERT-PKD by a substantial margin, even with only 30\% parameters and inference time. Furthermore, it exceeds the TinyBERT model by 2.3\% accuracy on RTE, 2.2\% F1 on MRPC, and 1.9\% Spearman correlation on STS-B. 6-layer BERT-EMD performs better than the 12-layer BERT-base model on 7 out of 9 tasks, with only about 50\% parameters and inference time of the original BERT-base model. Tang et al.~\cite{tang2019distilling} distill BERT to BiLSTMs. They observe that the distilled BiLSTM model uses 349 times fewer parameters than BERT-large and is 434 times faster. Also, mixed vocab training by Zhao et al.~\cite{zhao2019extreme} produces a small 12-layer model that performs competitively with 6-layer PKD and 4-layer DistilBERT while being $\sim$5-6x smaller.

MobileBERT is 4.3x smaller and 5.5x faster than BERT-base.  On the SQuAD v1.1/v2.0 question answering task, MobileBERT achieves a dev F1 score of 90.0/79.2 (1.5/2.1 higher than BERT-base). On the natural language inference tasks of GLUE, MobileBERT can achieve a GLUE score of 77.7, which is only 0.6 lower than BERT-base, with a latency of 62 ms on a Pixel 4 phone. While quantization can further compress MobileBERT by 4x, there is nearly no performance degradation from it.  SqueezeBERT is approximately half the size of BERT-base. MobileBERT is half the size of SqueezeBERT. SqueezeBERT is 4.3x faster than BERT-base, while MobileBERT is 3.0x faster than BERT-base. On 4 of GLUE tasks SqueezeBERT outperforms the accuracy of MobileBERT, while on the other 4 of the GLUE tasks MobileBERT outperforms SqueezeBERT. MobileBERT and SqueezeBERT outperform BERT-base significantly across all tasks.

To summarize, KD is a popular method for text based model compression. Various methods have proposed information copying using logits, softmax output, attention sub-layer output, value relation, relative dissimilarity information from both the last layer as well as intermediate layers of the teacher. Many methods have been proposed to handle complex teacher-student configuration mismatches in terms of vocabulary, number of attention heads, and hidden layer sizes. Also, KD has been found to be very effective in complex problem settings like multi-lingual tasks and tasks with large number of classes. Learning from noisy teachers, teacher assistants, an ensemble of teachers has been found to be effective as well. KD is the best model compression method especially in settings where a large amount of unlabeled data exists; distillation with data pseudo-labeled by teacher leads to very effective students.

\section{Parameter sharing}
\label{sec:parameterSharing}
Rather than removing weights or reducing \#bits to store them, parameter sharing methods reduce model size by finding weight blocks that can share the same weight. Character-based language models learn embeddings for characters and use them to compose word embeddings. In some senses, we can think of various words sharing these character embedding parameters.  Further, various parameter sharing methods have been proposed to reduce the large word embedding matrix size. Finally, there are multiple Transformer architectures which benefit from the parameter sharing philosophy. We discuss these methods in this section. 

\subsection{Character-aware Language Models}
\begin{figure}
    \centering
    \includegraphics[width=0.6\columnwidth]{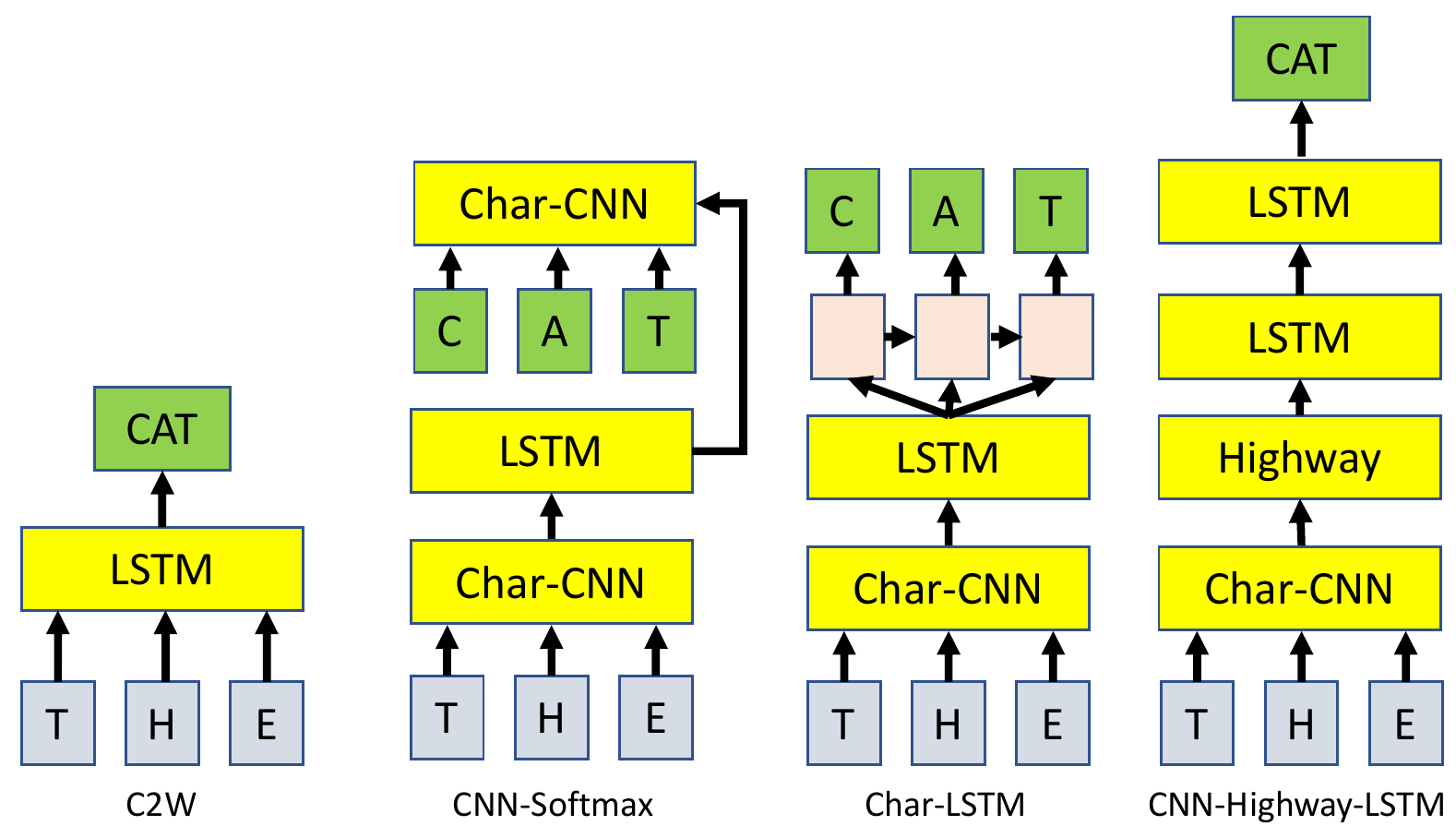}
    \caption{Character-aware Language Models}
    \label{fig:charCNN}
\end{figure}

Fig.~\ref{fig:charCNN} illustrates various character-aware language model architectures. Ling et al.~\cite{ling2015finding} proposed their character to word (C2W) model which constructs vector representations of words by composing characters using BiLSTMs. Relative to traditional word representation models that have independent vectors for each word type, C2W requires only a single vector per character type and a fixed set of parameters for the compositional model. As input, we define an alphabet of characters $C$. For English, this vocabulary would contain an entry for each uppercase and lowercase letter as well as numbers and punctuation. Thus compared to the word embedding matrix, this model is much smaller. Despite the compactness of this model, this ``composed'' word representations yield comparable results across multiple text classification tasks. 

Jozefowicz et al.~\cite{jozefowicz2016exploring} propose two variants for composing word embeddings using character embeddings. In the first CNN-Softmax variant, they use character CNNs (Convolutional Neural Networks) to compose word embeddings from character embeddings both at the input side as well as at the output softmax layer. The character-CNN sub-networks at the input or the output do not share weights. The composed word embeddings are fed to an LSTM to generate the output. In the second Char-LSTM variant, character CNN is used to compose word embeddings on the input side. The composed word embeddings are fed to an LSTM to generate an output which is further fed to a small LSTM that predicts the target word one character at a time. Thus, the word and character-level models are combined, and predictions are made one character at a time, thus allowing to compute probabilities over a much smaller vocabulary. Kim et al.~\cite{kim2016character} propose another variant where at the output side they continue to use word embeddings, but at the input side they compose word embeddings using a highway network on top of a character CNN. The highway network's output is used as the input to a multi-layer LSTM, whose last hidden state output is fed to the output softmax layer.

\subsection{Parameter Sharing in the Embedding Matrix}
Given a weight matrix $W$ and a budget $K$, we want to share weights within $W$ to have a max of $K$ unique values. A na\"ive implementation of random weight sharing can be trivially achieved by maintaining a secondary matrix consisting of each connection's group assignment. But this needs memory space itself. Hence, Chen et al.~\cite{chen2015compressing} propose to use hashing. HashedNets use a low-cost hash function (like xxhash\footnote{\url{https://code.google.com/p/xxhash/
}}) to randomly group connection weights into hash buckets, and all connections within the same hash bucket share a single parameter value.

Unlike HashedNets where weights are randomly grouped, parameter sharing mechanisms in Toeplitz-like structured matrices~\cite{lu2016learning} are highly specific and deterministic. Toeplitz matrices have parameters tied along diagonals. The displacement rank of all Toeplitz matrices is up to 2. Toeplitz-like matrices allow the displacement rank $r$ to be higher. They include products and inverses of Toeplitz matrices, and their linear combinations. The displacement rank $r$ serves as a knob on modeling capacity. High displacement rank matrices are increasingly unstructured. With displacement rank $r$, there are $2nr$ free parameters in the Toeplitz-like structured matrix. Toeplitz transforms can be applied not just to embedding matrix but to all weight matrices in an RNN model. Tay et al.~\cite{tay2019lightweight} use a similar Toeplitz-like structured matrix method with Hamilton Products in Quaternion Algebra to propose Quaternion Transformers which lead to 75\% parameter reduction in the Transformer architecture.

Another method for parameter sharing is to share low-rank factors across layers in a recurrent model. In this method, we first represent a weight matrix $W$ using matrix factorization as $W=W_a W_b$. Thus, hidden layer output for layer $l$ at time $t$ can be written as follows.
\begin{eqnarray}
h_t^l=\sigma\left[W_a^l W_b^l h_t^{l-1}+U_a^l U_b^l h_{t-1}^l + b^l\right]    
\end{eqnarray}
But we can share some low-rank factors by setting $W_b^l=U_b^{l-1}$. The combination of matrix factorization and parameter sharing leads to large model compression.

Another way of compressing the embedding matrix is to divide the vocabulary $V$ into frequent and infrequent word sets $B$ and $C$ respectively. Infrequent words' embeddings are represented with frequent words' by sparse linear combinations~\cite{chen2016compressing}. This is inspired by the observation that, in a dictionary, an unfamiliar word is typically defined by common words. A dense embedding is assigned to each common word; an infrequent word, on the other hand, computes its vector representation by a sparse combination of common words' embeddings. This compression is useful for both word embedding matrix as well as output layer of RNNs/LSTMs. Let $U\in R^{E\times |B|}$ be the learned embedding matrix of common words where $E$ is the embedding dimension. For a word $w\in C$, we shall learn a sparse vector $x\in R^{|B|}$ as the sparse code of the word. Once we know $x$, embedding for a word $w\in C$ can be written as follows.
\begin{eqnarray}
    \text{embedding}(w)=\sum_{j=1}^B x_jU_j
\end{eqnarray}
where $U_j$ is the $j^{th}$ column of $U$. To learn the sparse representation of word $w\in C$, the following problem needs to be solved.
\begin{eqnarray}
\min_x ||Ux-A||_2^2+\alpha ||x||_1+\beta |1^Tx-1|+\gamma 1^T \max(0,-x)
\end{eqnarray}
where $A$ is embedding for the rare word $w$. The last two regularization terms favor a solution that sums to 1 and that is non-negative (for psychological interpretation concerns), respectively. 

LightRNN~\cite{li2016lightrnn} compresses word embedding matrix from $O(|V|)$ to $O(\sqrt{|V|})$. It uses a 2-Component shared embedding for word representations. We allocate every word in the vocabulary into a word-allocation table, each row of which is associated with a learned vector, and each column associated with another learned vector. Table~\ref{tab:wordAllocation} shows an example of a word allocation table. Depending on its position in the table, a word is jointly represented by two components: a row vector and a column vector. Thus, we only need $2\sqrt{|V|}$ vectors to represent a vocabulary of $|V|$ unique words, which are far less than the $|V|$ vectors. 
The input and output use different embedding row/column vectors but they share the same word-allocation table. Word Allocation table creation uses a bootstrap procedure to iteratively refine word allocation based on the learned word embedding. Embeddings (i.e. row and column vectors) are learned using language modeling loss using an RNN on top of the embedding layer.

\begin{table}
    \centering
    \begin{tabular}{|c|c|c|c|c|}
    \hline
Embedding&$x_1^c$&$x_2^c$&$x_3^c$&$x_4^c$\\
\hline
$x_1^r$&january&february&...&...\\
\hline
$x_2^r$&one&two&...&...\\
\hline
$x_3^r$&...&...&...&...\\
\hline
$x_4^r$&...&...&...&...\\
\hline
\end{tabular}
    \caption{An Example of a Word Allocation Table}
    \label{tab:wordAllocation}
\end{table}

Finally, Suzuki et al.~\cite{suzuki2016learning} propose a Skipgram~\cite{mikolov2013word2vec} training method with parameter sharing as follows. Split every embedding vector of size $D$ into $B$ equal sub-vectors of size $C$. Thus $D=B\times C$. We assign a limited number of reference vectors to each block of block-splitting vectors. E.g., the number of reference vectors becomes $K\times B$ if we assign $K$ reference vectors to each block. Each reference vector is of size $C$. Skipgram training optimization remains the same except for these extra parameter sharing constraints (applied to both the input and output embedding vectors). Liu et al.~\cite{li2018slim} propose a very similar method, Slim Embeddings, where the embeddings are learned using a RNN language model rather than the Skipgram method. Slim Embeddings is very related to HashedNets~\cite{chen2015compressing}, LightRNN~\cite{li2016lightrnn} and
Character Aware Language model~\cite{kim2016character}. In HashedNets, all elements in a parameter matrix are mapped into a vector through a hash function. However in Slim Embeddings approach, we randomly share subvectors instead of single elements. Slim Embeddings differs from LightRNN  in
that it is able to control the compression ratio
to any arbitrary value, while LightRNN can only compress
at the rate of square or cube root of vocabulary size, which
could be too harsh in practical applications. In character aware language model, if we treat the sequence of subvector ids (virtual characters) as each word's representation,
the word embedding then can be treated as concatenated unigram character feature vectors. The drawback of such an approach is that the model is
more complicated and to speed up inference, it needs to pre-compute the word embeddings for the words, so it couldn't
stay in its compact form during inference. The Slim embeddings model is much simpler, and easier to tune. And during inference, it uses much less space and can even decrease the complexity of inference.

\subsection{Parameter Sharing in Transformers}

A standard Transformer does not share parameters across layers and also has a fixed number of encoder layers. ALBERT~\cite{lan2019albert} incorporates two parameter reduction techniques: 
\begin{itemize}
    \item Factorized embedding parameterization. That is, it decomposes large vocabulary embedding matrix into two small matrices. Thus, it reduces the embedding parameters from $O(V \times H)$ to $O(V \times E + E \times H)$ where $H >> E$.
    \item Cross-layer parameter sharing: There are multiple ways to share parameters, e.g., only sharing feed-forward network (FFN) parameters across layers, or only sharing attention parameters. The default decision for ALBERT is to share all parameters across layers.
\end{itemize}

An ALBERT configuration similar to BERT-large has 18x fewer parameters and can be trained about 1.7x faster. Dehghani et al.~\cite{dehghani2018universal} propose Universal Transformers where the number of encoder layers are not pre-decided, and all the encoder layers share the parameters. Certain symbols (e.g. some words or phonemes) are usually more ambiguous than others. It is therefore reasonable to allocate more processing resources to these more ambiguous symbols. Thus, ambiguous symbols undergo more self-attention transformations compared to non-ambiguous ones. Thus, they provide a dynamic per-position halting mechanism for dynamically modulating the number of computational steps needed to process each input symbol (called the ``ponder time'') before the representation is passed on as input to the decoder. The idea of sharing weights across layers in Transformers has also been explored in~\cite{bai2019deep}. 

\subsection{Summary}

\begin{table}
    \centering
    \scriptsize
    \begin{tabular}{|l|p{0.8in}|l|l|l|l|l|l|}
    \hline
Task&Dataset&Method&Base Model&Metric&Size (Comp; Orig)&Eval. (Comp; Orig)
\\
\hline
\hline
Language modeling&1B Word Benchmark&Char-CNN (input embeddings)~\cite{jozefowicz2016exploring}&2-layer word-BiLSTM&Perplexity (L)&1.04B; 1.8B&30.0; 30.6\\
\hline
Language modeling&1B Word Benchmark&Char-CNN (in/output embeddings)~\cite{jozefowicz2016exploring}&2-layer word-BiLSTM&Perplexity (L)&0.39B; 1.8B&35.8; 30.6\\
\hline
Language modeling&1B Word Benchmark&LightRNN~\cite{li2016lightrnn}&word-LSTM&Perplexity (L)&41M; 1.6G&66; 85\\
\hline
Language modeling&1B Word Benchmark&Slim Embeddings~\cite{li2018slim}&2-layer word-BiLSTM&Perplexity (L)&0.25B; 1.8B&38.3; 30.6\\
\hline
Language modeling&ACLW-Czech&CNN+Highway network~\cite{kim2016character}&word-LSTM&Perplexity (L)&64M; 83M&578; 701\\
\hline
Language modeling&ACLW-Czech&LightRNN~\cite{li2016lightrnn}&word-LSTM&Perplexity (L)&18M; 83M&558; 701\\
\hline
Language modeling&ACLW-Czech&Slim Embeddings~\cite{li2018slim}&word-LSTM&Perplexity (L)&17M; 83M&528; 701\\
\hline
Language modeling&ACLW-English&CNN+Highway network~\cite{kim2016character}&word-LSTM&Perplexity (L)&20M; 25M&216; 236\\
\hline
Language modeling&ACLW-English&LightRNN~\cite{li2016lightrnn}&word-LSTM&Perplexity (L)&17M; 25M&191; 236\\
\hline
Language modeling&ACLW-English&Slim Embeddings~\cite{li2018slim}&word-LSTM&Perplexity (L)&7M; 25M&187; 236\\
\hline
Language modeling&ACLW-French &CNN+Highway network~\cite{kim2016character}&word-LSTM&Perplexity (L)&44M; 56M&190; 202\\
\hline
Language modeling&ACLW-French &LightRNN~\cite{li2016lightrnn}&word-LSTM&Perplexity (L)&17M; 56M&176; 202\\
\hline
Language modeling&ACLW-French &Slim Embeddings~\cite{li2018slim}&word-LSTM&Perplexity (L)&12M; 56M&162; 202\\
\hline
Language modeling&ACLW-German &CNN+Highway network~\cite{kim2016character}&word-LSTM&Perplexity (L)&104M; 137M&305; 347\\
\hline
Language modeling&ACLW-German &LightRNN~\cite{li2016lightrnn}&word-LSTM&Perplexity (L)&18M; 137M&281; 347\\
\hline
Language modeling&ACLW-German &Slim Embeddings~\cite{li2018slim}&word-LSTM&Perplexity (L)&17M; 137M&261; 347\\
\hline
Language modeling&ACLW-Russian&CNN+Highway network~\cite{kim2016character}&word-LSTM&Perplexity (L)&152M; 200M&313; 353\\
\hline
Language modeling&ACLW-Russian&LightRNN~\cite{li2016lightrnn}&word-LSTM&Perplexity (L)&19M; 200M&288; 353\\
\hline
Language modeling&ACLW-Russian&Slim Embeddings~\cite{li2018slim}&word-LSTM&Perplexity (L)&19M; 200M&274; 353\\
\hline
Language modeling&ACLW-Spanish&CNN+Highway network~\cite{kim2016character}&word-LSTM&Perplexity (L)&48M; 61M&169; 186\\
\hline
Language modeling&ACLW-Spanish&LightRNN~\cite{li2016lightrnn}&word-LSTM&Perplexity (L)&18M; 61M&157; 186\\
\hline
Language modeling&ACLW-Spanish&Slim Embeddings~\cite{li2018slim}&word-LSTM&Perplexity (L)&8M; 61M&149; 186\\
\hline
Language modeling&PTB&CNN+Highway network~\cite{kim2016character}&word-LSTM&Perplexity (L)&5M; 20M&92.3; 85.4\\
\hline
Language modeling&Wikipedia articles (ca)&C2W~\cite{ling2015finding}&word emb.&Perplexity (L)&182K; 4.3M&34.92; 35.34\\
\hline
Language modeling&Wikipedia articles (de)&C2W~\cite{ling2015finding}&word emb.&Perplexity (L)&183K; 6.3M&41.94; 43.02\\
\hline
Language modeling&Wikipedia articles (en)&C2W~\cite{ling2015finding}&word emb.&Perplexity (L)&180K; 4.3M&57.39; 59.38\\
\hline
Language modeling&Wikipedia articles (pt)&C2W~\cite{ling2015finding}&word emb.&Perplexity (L)&178K; 4.2M&40.92; 46.17\\
\hline
Language modeling&Wikipedia articles (tr)&C2W~\cite{ling2015finding}&word emb.&Perplexity (L)&174K; 5.7M&32.88; 44.01\\
\hline
Machine reading comp.&ReAding Comprehension from Examinations&ALBERT~\cite{lan2019albert}&BERT-B&Acc (H)&18M; 108M&68.5; 68.2\\
\hline
NLI&MNLI&Quaternion Attention~\cite{tay2019lightweight}&2-layer att-GloVe&Acc (H)&200K; 700K&72.3; 73.6\\
\hline
NLI&SciTail&Quaternion Attention~\cite{tay2019lightweight}&2-layer att-GloVe&Acc (H)&200K; 700K&79.6; 79.0\\
\hline
NLI&SNLI&Quaternion Attention~\cite{tay2019lightweight}&2-layer att-GloVe&Acc (H)&200K; 700K&85.4; 86.2\\
\hline
NMT  (en$\rightarrow$et)&IWSLT15&Quaternion Transformer~\cite{tay2019lightweight}&Transformer&BLEU (H)&11M; 44M&13.1; 14.1\\
\hline
NMT (en$\rightarrow$ro)&IWSLT15&Quaternion Transformer~\cite{tay2019lightweight}&Transformer&BLEU (H)&11M; 44M&18.5; 22.8\\
\hline
NMT (en$\rightarrow$vi)&IWSLT15&Quaternion Transformer~\cite{tay2019lightweight}&Transformer&BLEU (H)&11M; 44M&28.0; 28.4\\
\hline
POS Tagging&PTB (ca)&C2W~\cite{ling2015finding}&word-BiLSTM&Acc (H)&150K; 2M&98.92; 98.09\\
\hline
POS Tagging&PTB (de)&C2W~\cite{ling2015finding}&word-BiLSTM&Acc (H)&150K; 2M&98.08; 97.51\\
\hline
POS Tagging&PTB (en)&C2W~\cite{ling2015finding}&word-BiLSTM&Acc (H)&150K; 2M&97.36; 96.97\\
\hline
POS Tagging&PTB (pt)&C2W~\cite{ling2015finding}&word-BiLSTM&Acc (H)&150K; 2M&97.47; 95.67\\
\hline
POS Tagging&PTB (tr)&C2W~\cite{ling2015finding}&word-BiLSTM&Acc (H)&150K; 2M&91.59; 83.43\\
\hline
Question answering&BABI&Universal Transformer~\cite{dehghani2018universal}&Transformer&Avg error (L)&7.3M; 44M&0.21; 15.2\\
\hline
Question answering&WikiQA&Quaternion Attention~\cite{tay2019lightweight}&2-layer att-GloVe&MAP (H)&200K; 700K&66.2; 67.2\\
\hline
Sentiment analysis&IMDB&Quaternion Transformer~\cite{tay2019lightweight}&2-layer Transformer&Acc (H)&100K; 400K&83.9; 82.6\\
\hline
Sentiment analysis&SST&Quaternion Transformer~\cite{tay2019lightweight}&2-layer Transformer&Acc (H)&100K; 400K&80.5; 78.9\\
\hline
Speech recognition&2000 hour En. Speech&Toeplitz-like~\cite{lu2016learning}&3-layer RNN&WER (L)&790K; 1.85M&48.4; 43.5\\
\hline
Speech recognition&2000 hour En. Speech&Toeplitz-like~\cite{lu2016learning}&5-layer LSTM&WER (L)&2.00M; 9.12M&33.5; 33.1\\
\hline
Subject verb agreement&SVA dataset&Quaternion Transformer~\cite{tay2019lightweight}&2-layer Transformer&Acc (H)&100K; 400K&94.7; 94.8\\
\hline
Subject verb agreement&SVA dataset&Universal Transformer~\cite{dehghani2018universal}&Transformer&Acc (H)&7.3M; 44M&0.992; 0.962\\
\hline
Word analogy &GSEm, GSYN, MSYN&Shared ref. vec. ($K$=16/256, $B$=256)~\cite{suzuki2016learning}&SGNS&Acc (H)&98/196MB; 1565MB&64.6/64.5; 64.7\\
\hline
Word analogy &GSEm, GSYN, MSYN&k-means quant. ($K$=16/256, $B$=256)~\cite{suzuki2016learning}&SGNS&Acc (H)&98/196MB; 1565MB&64.0/64.5; 64.7\\
\hline
Word similarity&7 datasets&Shared ref. vec. ($K$=16/256, $B$=256)~\cite{suzuki2016learning}&SGNS&Acc (H)&98/196MB; 1565MB&65.5/67.1; 67.2\\
\hline
Word similarity&7 datasets&k-means quant. ($K$=16/256, $B$=256)~\cite{suzuki2016learning}&SGNS&Acc (H)&98/196MB; 1565MB&64.4/67.0; 67.2\\
\hline
    \end{tabular}
\caption{Comparison of various parameter sharing methods (sorted by Task and then Dataset). 7 datasets for word similarity are MEN, MTurk, RARE, SLex, SCWS, WSR, WSS. SGNS=SkipGram with negative sampling. In the metric column, H means high is better while L means low is better.}
    \label{tab:paramSharingSummary}
\end{table}

Table~\ref{tab:paramSharingSummary} compares various parameter sharing methods across different tasks and datasets. Accuracy of both the original and the compressed (comp.) model are shown. Also, we report model size (in terms of number of parameters or memory size) for both the original as well as the compressed models. For the same task, dataset and model combination, different papers report different accuracy of the original model because of slight changes in training hyper-parameters; hence we report accuracy of the original model for each row. Since many parameter sharing methods have been used to compress word embeddings, the most common application is language modeling. 

For language modeling, experiments have been done on the One Billion Word Benchmark, ACLW, PTB and Wikipedia articles datasets across multiple languages using parameter sharing methods like Char-CNN~\cite{jozefowicz2016exploring}, LightRNN~\cite{li2016lightrnn}, Slim embeddings~\cite{li2018slim}, CNN+Highway networks~\cite{kim2016character} and C2W~\cite{ling2015finding}. C2W~\cite{ling2015finding} always outperforms word lookup tables and improvements are especially pronounced in Turkish, which is a highly morphological language, where word meanings differ radically depending on the suffixes used. Jozefowicz et al.~\cite{jozefowicz2016exploring} observe that Char-CNNs especially with character embeddings being used both at input as well as output can lead to 4.6x model size reduction with a slight increase in perplexity. On the One-Billion-Word benchmark, LightRNN~\cite{li2016lightrnn} achieves much lower perplexity compared to word-LSTMs, whilst reducing the model size by a factor of 40-100, and speeding up the training process by a factor of 2. LightRNN~\cite{li2016lightrnn} significantly reduces the model size, while at the same time outperforms CNN+Highway network~\cite{kim2016character} method. While the model sizes of the CNN+Highway network method increase linearly with respect to the vocabulary size, the model size of LightRNN almost keeps constant on the ACLW datasets. Slim embeddings~\cite{li2018slim} is way better than LightRNN (low perplexity with smaller models). On PTB and 44M giga word corpus datasets, Slim embeddings applied at input layer can maintain same perplexity for a word-LSTM using just 1\% (and 0.2\%) of trainable parameters respectively. 

k-means quantization and shared reference vectors are also methods for compression of word embeddings using parameter sharing. Suzuki et al.~\cite{suzuki2016learning} showed significant gains over typical skipgram (SGNS) embeddings in terms of model size reduction while retaining similar accuracies for word analogy and similarity tasks. The model size of their method with shared reference vectors with `K = 16, B = 64' was just 24MB, approximately 65 times smaller than that of original SGNS method. They also showed that SGNS with shared reference vectors was better than SGNS with block-wise k-means post-processing method. Unfortunately, there exists no good comparison between the slim embeddings and the shared reference vectors methods.

For the MRC task, ALBERT~\cite{lan2019albert} pushed the accuracy from 68.5 to 68.2 using 6x fewer parameters compared to BERT-base. On the NLI task, a tiny Quaternion-Att model (50 dimensions) achieves comparable (or occasionally marginally better or worse) performance compared to typical attention over GloVe (200 dimensions), gaining a 68\% parameter savings across three datasets. For sentiment analysis, Quaternion Transformers~\cite{tay2019lightweight} leads by +1.3\%/1.6\% gains on IMDb and SST datasets respectively while maintaining a 75\% reduction in parameter cost. Quaternion Transformers~\cite{tay2019lightweight} have been shown to outperform the vanilla Transformer for the mathematical language understanding task as well, with 25\% parameters. At  the same compression rate, Quaternion Transformers lose only very minor BLEU on IWSLT 2015 NMT datasets.

For POS tagging, we can observe that the model using word lookup tables performs consistently worse than the C2W model~\cite{ling2015finding}. Universal Transformers~\cite{dehghani2018universal} (while being 1/6th the size) outperform standard Transformers on a wide range of algorithmic and language understanding tasks, including the challenging LAMBADA language modeling task. On speech recognition, Lu et al.~\cite{lu2016learning} study mechanisms for learning compact RNNs and LSTMs via low-rank factorizations and parameter sharing schemes. A hybrid strategy of using structured matrices in the bottom layers and shared low-rank factors on the top layers is found to be particularly effective, reducing the parameters of a standard LSTM by 75\%, at a small cost of 0.3\% increase in WER, on a 2000-hr English Voice Search task. For LSTM parameter reduction, architecting upper layers with projection nodes to moderate rank, and bottom layers with Toeplitz-like transforms was found to be a particularly effective strategy. 

Overall, besides model compression, parameter sharing methods also act as a good regularizer. Parameter sharing in Transformers has been very successful. ALBERT was at the top of the GLUE leaderboard when it was proposed. Parameter sharing methods have also been widely used for compressing embedding matrix. Slim embeddings has the best method for compressing embedding matrices.

\section{Tensor decomposition}
\label{sec:matrixDecomposition}
Sparse Matrix decomposition has been traditionally used for applications like feature selection, collaborative filtering, topic mining from text, etc. In this section, we discuss how various popular tensor decomposition methods like Singular Value Decomposition (SVD), Tensor-Train~\cite{oseledets2011tensor}, CP (CANDECOMP/PARAFAC)~\cite{carroll1970analysis} and Tucker~\cite{tucker1966some} can be used for model compression. 
\subsection{Two Low-Rank Factors}
In this part, we will discuss methods where a matrix is factorized into two low-rank factors. Specifically, we replace a weight matrix $W$ with $W_1\times W_2$ such that the total number of parameters are significantly lesser.

A multi-layer RNN can be represented as follows.

\begin{eqnarray}
    h_t^l=\sigma(W_x^{l-1}h_t^{l-1}+W_h^l h_{t-1}^l +b^l)\\
    h_t^{l+1}=\sigma(W_x^l h_t^l + W_h^{l+1}h_{t-1}^{l+1}+b^{l+1})
\end{eqnarray}

Thus, there are two important weight matrices: the recurrent $W_h^l$ and inter-layer matrices $W_l^x$. Prabhavalkar et al.~\cite{prabhavalkar2016compression} propose a method to jointly compress the recurrent and inter-layer matrices corresponding to a specific layer $l$ by determining a suitable recurrent projection matrix, denoted by $P^l\in R^{r_l\times N_l}$ of rank $r^l<N^l$ such that $W_h^l=Z_h^l P^l$ and $W_l^x=Z_x^l P^l$. First, $P^l$ is determined by  computing a truncated SVD of the recurrent weight matrix, which we then truncate, retaining only the top $r^l$ singular values. Thus, $W_h^l$ can be written as follows.
\begin{eqnarray}
    W_h^l=(U_h^l \Sigma_h^l)(V_h^l)^T=Z_h^lP^l
\end{eqnarray}
Thus, $P^l$ is set to $(V_h^l)^T$. Further, we determine $Z_x^l$ as the solution to the following least-squares problem.
\begin{eqnarray}
   Z_x^l=\argmin_Y ||YP^l-W_x^l||_2^2
\end{eqnarray}
This solution can also be easily extended to LSTMs. Sak et al.~\cite{sak2014long} also proposed a similar solution based on a combination of parameter sharing and matrix decomposition but without SVD initialization. However, typically SVD initialization has been found to perform better.

Besides SVD, another way of matrix decomposition is sparse coding. Faruqui et al.~\cite{faruqui2015sparse} propose using sparse coding to decompose word embedding matrices. Thus, given vocabulary of size $V$, word embedding matrix  $X\in R^{L\times V}$, sparse coding aims at representing each input vector $x_i$ as a sparse linear combination of basis vectors $a_i$ by solving the following problem. 
\begin{eqnarray}
  \argmin_{D,A} \sum_{i=1}^V ||x_i-Da_i||_2^2+\lambda||a_i||_1+\tau ||D||_2^2
\end{eqnarray}
where $D\in R^{L\times K}$ and $A\in R^{K\times V}$. Further, for interpretability, one can enforce all elements of $A$ and $D$ to be non-negative. For further compression, one can also enforce $A$ to be binary or ensure that each column of $A$ is a $K$ sized one hot vector~\cite{shu2017compressing}. 

Lastly, Wang et al.~\cite{wang2019structured} combine pruning with matrix factorization for model compression and propose the FLOP (Factorized Low-rank Pruning) method. Let $W$ be a weight matrix. Structured pruning (removing a neuron, i.e., removing a column from weight matrix) can be achieved by replacing the computation $Wx$ by $WGx$ where diagonal sparsity-inducing matrix $G$ is learned using $L_0$ regularization over $WG$ along with the supervised loss. This effectively removes a subset of columns of $W$ for column indices $k$ with $z_k=0$. One limitation is that this structured pruning method tends to produce lower performance than its unstructured counterpart. Hence, in the FL0P (Factorized L0 Pruning) model, we first factorize $W=PQ$. Let $r$ be \#columns of $P$ (or equivalently \#rows of $Q$), $p_k$ and $q_k$ be the $k$-th column of $P$ and $k$-th row of $Q$ respectively. We achieve structured pruning by introducing a pruning variable $z_k$ for each component. Thus, now, we can write $W$ as follows.
\begin{eqnarray}
  W=PGQ=\sum_{k=1}^r z_k\times (p_kq_k)
\end{eqnarray}
where $G$ is again the diagonal matrix of pruning variables. After training, only columns and rows corresponding to non-zero diagonal values need to be stored, resulting in much smaller (but still dense) matrices $P$ and $Q$. The nonzero values of $G$ can be absorbed into either $P$ or $Q$. This structured pruning with factorization is much more effective compared to the vanilla structured pruning.

\subsection{Factorizing into Block Diagonal Matrices}
The last layer of a language model is very large of the size $HV$ where $H$ is the size of the hidden layer and $V$ is vocabulary size. Each word by an output embedding of the same size $H$. Chen et al.~\cite{chen2015strategies} propose a differentiated softmax method which varies the dimension of the output embeddings across words depending on how much model capacity is deemed suitable for a given word. In particular, it is meaningful to assign more parameters to frequent words than to rare words. By definition, frequent words occur more often in the training data than rare words and therefore allow to fit more parameters. They define partitions of the output vocabulary based on word frequency and the words in each partition share the same embedding size. Partitioning results in a sparse final weight matrix which arranges the embeddings of the output words in blocks, each one corresponding to a separate partition. The size of the final hidden layer $H$ is the sum of the embedding sizes of the partitions. While this method does not involve creation of multiple factors, it factorizes the original matrix into multiple blocks while setting the remaining part of the matrix to 0.

Variani et al.~\cite{variani2019west} propose a method called Word Encoded Sequence Transducers (WEST) which factorizes a matrix $E=C\times D$ where $D$ is constrained to be a block diagonal matrix. The block diagonal nature of the second factor leads to large compression rates.

\subsection{Tensor Train and Block Term Decomposition}
Tensor train decomposition (TTD)~\cite{oseledets2011tensor} is a standard tensor decomposition technique which decomposes a high dimensional tensor into multiple 2D and 3D tensors which can be multiplied together to reconstruct the original tensor. These factors are called TT-cores and the other dimensions are referred to as TT-ranks. TTD can be leveraged to compress various weight matrices in RNNs and LSTMs~\cite{tjandra2017compressing,khrulkov2019tensorized}. The first step is to represent a matrix as a multi-dimensional tensor by simple reshaping transformation and then use TTD on it. The values of TT–ranks directly define the compression ratio, so choosing them to be too small or too large will result into either significant performance drop or little reduction of the number of parameters. Typically TT-ranks around 16 for small matrices and 64-192 for larger matrices result in a good trade-off between compression ratio and the accuracy metric of interest. Also, when we use TTD for weight matrices, we also need change the inputs appropriately to be compatible in terms of dimensions. 

Compared with TT-RNN, Block-Term RNN (BTRNN)~\cite{ye2018learning} is not only more concise (when using the same rank), but also able to attain a better approximation to the original RNNs with much fewer parameters. BTD decomposes a high order tensor into a sum of multiple Tucker decomposition models. The redundant dense connections between input and hidden state is first tensorized to a $d$-dimensional tensor and then decomposed using low-rank BTD into a sum of $N$ different Tucker decompositions where $N$ is the CP-rank. Each Tucker decomposition in turn consists of a core $d$-dimensional tensor and $d$ 3-dimensional factor tensors. While Ye et al.~\cite{ye2018learning} used BTD to compress RNNs, Ma et al.~\cite{ma2019tensorized} used BTD to compress the self-attention matrix in Transformers. They first build a single-block attention based on the Tucker decomposition where the query, key and value are mapped into three factor matrices and the core tensor is trainable and randomly initialized. It is then straightforward to represent the multi-head attention using BTD.  

\subsection{Summary}

\begin{table}
    \centering
    \scriptsize
    \begin{tabular}{|l|l|l|l|l|l|l|l|}
    \hline
Task&Dataset&Method&Base Model&Metric&Size (Comp; Orig)&Eval. (Comp; Orig)
\\
\hline
\hline
CTR prediction&Criteo CTR Challenge&TT-embedding~\cite{khrulkov2019tensorized}&MLP&LogLoss (L)&4.7M; 41.2M&0.4433; 0.4440\\
\hline
Language modeling&1B Word Benchmark&BTD~\cite{ma2019tensorized}&Transformer-XL Large&Perplexity (L)&0.16B; 0.8B&19.5; 21.8\\
\hline
Language modeling&Enwiki-8&FLOP~\cite{wang2019structured}&Transformer&BPC (L)&8M; 41M&1.13; 1.08\\
\hline
Language modeling&PTB&WEST~\cite{variani2019west}&LSTM&Perplexity (L)&3.5M; 4.51M&116.84; 115.91\\
\hline
Language modeling&PTB&BTD~\cite{ma2019tensorized}&Transformer-XL-Base&Perplexity (L)&12M; 24M&49.8; 54.52\\
\hline
Language modeling&PTB&TT-embedding~\cite{khrulkov2019tensorized}&Transformer-XL-Base&Perplexity (L)&18M; 24M&55.4; 54.52\\
\hline
Language modeling&WikiText-103&FLOP~\cite{wang2019structured}&Transformer&Perplexity (L)&50M; 151M&25.3; 24.1\\
\hline
Language modeling&WikiText-103&TT-embedding~\cite{khrulkov2019tensorized}&Transformer-XL&Perplexity (L)&91M; 192M&25.67; 24.37\\
\hline
Language modeling&WikiText-103&BTD~\cite{ma2019tensorized}&Transformer-XL-Base&Perplexity (L)&85.3M; 151M&20.9; 24.0\\
\hline
Language modeling&WikiText-103&TT-embedding~\cite{khrulkov2019tensorized}&Transformer-XL-Base&Perplexity (L)&130M; 151M&25.7; 24.0\\
\hline
NMT (en$\rightarrow$ja)&ASPEC&Compositional codes~\cite{shu2017compressing}&LSTM&BLEU (H)&2.97MB; 274MB&38.89; 37.93\\
\hline
NMT (de$\rightarrow$en)&IWSLT14&Compositional codes~\cite{shu2017compressing}&LSTM&BLEU (H)&2.11MB; 35MB&29.56; 29.45\\
\hline
NMT (en$\rightarrow$de)&WMT14&TT-embedding~\cite{khrulkov2019tensorized}&Transformer-Big&BLEU (H)&179M; 210M&28.53; 28.84\\
\hline
NMT (en$\rightarrow$de)&WMT16&BTD~\cite{ma2019tensorized}&Transformer&BLEU (H)&21.2M; 52M&34.91; 34.5\\
\hline
NP bracketing&Subset of PTB&Sparse coding~\cite{faruqui2015sparse}&Logistic regression&Acc (H)&120M; 120M&82.3; 77.9\\
\hline
Question classification&TREC Questions&Sparse coding~\cite{faruqui2015sparse}&Logistic regression&Acc (H)&120M; 120M&81.5; 76.2\\
\hline
Sentiment analysis&IMDB&Compositional codes~\cite{shu2017compressing}&LSTM&Acc (H)&1.23MB; 78MB&87.37; 87.18\\
\hline
Sentiment analysis&IMDB&TT-embedding~\cite{khrulkov2019tensorized}&LSTM&Acc (H)&0.81M; 7.19M&89.7; 88.6\\
\hline
Sentiment analysis&SST&Sparse coding~\cite{faruqui2015sparse}&Logistic regression&Acc (H)&120M; 120M&81.4; 77.7\\
\hline
Speech recognition&3M Google voice utterances&Joint-SVD~\cite{prabhavalkar2016compression}&5-layer RNN&WER (L)&3.1M; 9.7M&12.9; 12.4\\
\hline
Speech recognition&3M Google voice utterances&Projections~\cite{sak2014long}&LSTM&WER (L)&2M; 2.2M&14.8; 17.5\\
\hline
Speech recognition&Live traffic utterances&WEST~\cite{variani2019west}&3-layer LSTM&WER (L)&4.75MB; 15MB&13.6; 13.7\\
\hline
Speech recognition&Live traffic utterances&WEST+Quantization~\cite{variani2019west}&3-layer LSTM&WER (L)&1.35MB; 15MB&13.7; 13.7\\
\hline
Text classification&20 Newsgroup (Computers)&Sparse coding~\cite{faruqui2015sparse}&Logistic regression&Acc (H)&120M; 120M&87.0; 79.7\\
\hline
Text classification&20 Newsgroup (Religion)&Sparse coding~\cite{faruqui2015sparse}&Logistic regression&Acc (H)&120M; 120M&88.8; 86.7\\
\hline
Text classification&20 Newsgroup (Sports)&Sparse coding~\cite{faruqui2015sparse}&Logistic regression&Acc (H)&120M; 120M&96.3; 95.9\\
\hline
Word similarity&Simlex-999&Sparse coding~\cite{faruqui2015sparse}&Logistic regression&Correlation (H)&120M; 120M&38.9; 36.9\\
\hline
    \end{tabular}
\caption{Comparison of various tensor decomposition methods (sorted by Task and then Dataset). In the metric column, H means high is better while L means low is better. For compositional codes, 16x32 coding was used. For BTD, two block term tensors were used. In~\cite{faruqui2015sparse}, logistic regression uses GloVe embeddings.}
    \label{tab:tensorDecompSummary}
\end{table}

Table~\ref{tab:tensorDecompSummary} compares various tensor decomposition methods across different tasks and datasets. Accuracy of both the original and the compressed (comp.) model are shown. Also, we report model size (in terms of number of parameters or memory size) for both the student as well as the teacher models. For the same task, dataset and model combination, different papers report different accuracy of the original model because of slight changes in training hyper-parameters; hence we report accuracy of the original model for each row. 

For CTR prediction, the TT-embedding method~\cite{khrulkov2019tensorized} in Table~\ref{tab:tensorDecompSummary} uses 3 factors with TT-rank of 16. It actually leads to an embedding compression of 16. With 4 factors and TT-rank=2, test loss increases to 0.4530 with a massive embedding compression of 4193 and overall model size of 0.53M. Thus, substitution of large embedding layers with TT–embeddings leads to significant compression ratios (up to 2011 times) with a slight improvement in the test loss, and up to 4200 with a small drop in the test loss. 

For language modeling, BTD~\cite{ma2019tensorized} leads to an improved model with 20\% of the Transformer-XL large model. For character level language modeling using FLOP~\cite{wang2019glue}, an 8M sized FLOP model achieves 1.13 on Enwiki8 while gradual pruning achieves 1.14. Thus, FLOP based pruning which combines pruning with matrix factorization is better than both structured neuron pruning as well as gradual unstructured pruning. On PTB, BTD is clearly much better than TT-embedding. With half the model size compared to Transformer-XL-Base, BTD leads to a model with $\sim$10\% lower perplexity. On WikiText-103, while FLOP pruned model achieves 25.3 perplexity with model size of 50M, gradual unstructured pruning and neuron pruning achieve 25.7 and 26.7 respectively with the same model size for language modeling on Wiki-103 dataset. Thus, FLOP is better than other pruning methods. Again on Wiki-103, BTD is superior to TT-embedding.    

For NMT, the loss-free compression rate reaches 92\% on ASPEC dataset by pruning 90\% of the connections. However, with the same pruning ratio, a modest performance loss is observed in IWSLT14 dataset. For the models using compositional coding~\cite{shu2017compressing}, the loss-free compression rate is 94\% for the IWSLT14 dataset and 99\% for the ASPEC dataset. Thus, compositional codes are much better compared to pruning. TT-embedding with TT-rank=64 leads to embedding compression of 15.3 on WMT14 dataset with marginal loss in the BLEU score. Lastly, BTD has been used to reduce Transformer size by more than half with improved BLEU on WMT16 (en$\rightarrow$de) data.

Sparse coding for GloVe vectors~\cite{faruqui2015sparse} has led to improved accuracies for multiple tasks like Noun Phrase (NP) bracketing, question classification, text classification and word similarity, while retaining the same model size. For sentiment analysis on the IMDB dataset, compositional codes method achieves a compression rate of 98\% without performance loss. Further, TT-embedding method leads to a much smaller model compared to compositional codes with better accuracy. In the TT-embedding case, embedding compression rate is 441  with TT-rank=16.

For Speech recognition, Prabhavalkar et al.~\cite{prabhavalkar2016compression} experimented with a 3M Google voice utterances dataset and found that a joint SVD with explained variance retained after the SVD as 0.6 leads to a 3x smaller RNN model without much performance loss. They improved upon Sak et al.'s projections method which led to a much higher WER on the same dataset. On live traffic utterances dataset, WEST~\cite{variani2019west} leads to a 3x smaller model with a slightly reduced word error rate when using 3-layer LSTMs. Variani et al.~\cite{variani2019west} further compress this model using quantization to obtain a 11x compressed 3-layer LSTM model without any performance loss. Thus, using quantization along with matrix decomposition seems to work well.

Overall, matrix decomposition techniques are usually used in combination with parameter sharing and sometimes with quantization. They have been very effective in dealing with large input/output embedding matrices in RNNs and LSTMs. SVD, Tensor Train, CP, Tucker, BTD have been the most popular decomposition techniques found to be useful for model compression. 

\section{Transformers with Sub-Quadratic Complexity}
\label{sec:linearTransformers}
Memory usage throughout neural network training can be categorized into three main types: (1) Model memory is used to store model parameters; (2) Optimizer memory is the additional memory used by the specific learning algorithm during the process; (3) Activation memory consists of the outputs of each layer, which are cached for reuse in backpropagation to compute gradients. For a BERT-base model, model memory is around 0.2 GB, optimizer memory is around 1GB, while activation memory is around 8.5GB~\cite{qiu2020blockwise}. Time and activation memory in Transformers grows quadratically with the sequence length. This is because in every layer, every attention head attempts to come up with a transformed representation for every position by ``paying attention'' to tokens at every other position. Quadratic complexity implies that practically the maximum input size is rather limited. Thus, we cannot extract semantic representation for long documents by passing them as input to Transformers. 

\subsection{Transformers with Super-Linear Complexity}

A few efforts like BlockBERT~\cite{qiu2020blockwise} try to reduce this quadratic complexity by a constant factor by introducing sparse block structures into the attention matrix. If we split the length-$N$ input sequence into $n$ blocks, $N\times N$ attention matrix gets partitioned into $n\times n$ blocks, where each block matrix is of the size $\frac{N}{n}\times\frac{N}{n}$. Thus, BlockBERT reduces $O(N^2)$ memory consumption by a factor of $n$. 

Child et al.~\cite{child2019generating} propose sparse transformers where sparse factorizations of the attention matrix reduce the quadratic complexity  to $O(n\sqrt{n})$.  The key idea is to reduce the dense attention matrix to a sparse version by only computing attention on a sparse number of (query, key) pairs. They propose two kinds of sparse factorizations: strided and fixed. Strided attention implies having one head attend to the previous $l$ locations, and the other head attend to every $l^{th}$ location, where $l$ is the stride and chosen to be close to $\sqrt{n}$. More heads could be used with a different stride value. Fixed attention assumes that specific positions summarize previous locations and propagate that information to all future positions. 

The Reformer architecture~\cite{kitaev2020reformer} replaces the dot-product attention in a typical Transformer by one that uses locality-sensitive hashing (LSH), changing its complexity from $O(n^2)$ to $O(n\log n)$, where $n$ is the length of the sequence. In a standard Transformer, we compute scaled dot-product attention as follows.
\begin{eqnarray}
  \text{Attention}(Q,K,V)=\text{softmax}\left(\frac{QK^T}{\sqrt{d}}\right)V
\end{eqnarray}
where $Q$, $K$ and $V$ are the standard query, key and value components and $d$ is a scaling factor. Reformer uses a Shared QK Transformer, i.e., $Q=K$ enabled by sharing the matrix that projects words/hidden layer to $Q$ or $K$. Further, note that we are actually only interested in $\text{softmax}(QK^T)$. Since softmax is dominated by the largest elements, for each query $q_i$ we only need to focus on the keys in $K$ that are closest to $q_i$. How can we find the nearest neighbors among the keys? Reformer uses  LSH. LSH is used to cluster (hash-bucket) the positions into various groups, and then every position needs to focus only on others within the same bucket.

\subsection{Transformers with Linear Complexity}
Even better, there have been several efforts recently to reduce this quadratic complexity to linear. Most of these efforts choose a constant number of other positions to ``pay attention'' to so as to compute a transformed representation for any given position. They can model sequences tens of thousands of timesteps long using hundreds of layers. The methods differ in their approach towards selecting this constant number of other positions. We discuss a few of such recently proposed methods in this section.

In Star-Transformers~\cite{guo2019star}, to reduce model complexity from $O(n^2)$ to linear, we replace the fully-connected attention matrix structure with a star-shaped topology, in which every two non-adjacent nodes are connected through a shared relay node. While ring connections connect a satellite node with two other satellite nodes, a radical connection connects a satellite node with the relay node. The idea is to update the star-center relay node based on satellite nodes and then update satellite nodes using information from the star node, and adjacent satellite nodes. 

Linformer architecture~\cite{wang2020linformer} exploits low-rank factorization of the self-attention matrix to reduce overall self-attention complexity from $O(n^2)$ to $O(n)$ in both time and space. The main idea is to add two linear projection matrices $E_i, F_i\in R^{n\times k}$ when computing key and value. We first project the original $(n\times d)$-dimensional key and value layers into $(k\times d)$-dimensional projected key and value layers. We then compute an $(n \times k)$-dimensional context mapping matrix using scaled dot-product attention. If we can choose a very small projected dimension $k$, such that $k<<n$, then we can significantly reduce the memory and space consumption. Overall, it is $O(nk)$. Further, we can do three other forms of parameter sharing: (1) Headwise sharing: $E_i=E$ and $F_i=F$ across all heads $i$ in a  layer. (2) Key-value sharing: $E_i=F_i=E$ across all heads $i$ in a layer. (3) Layerwise sharing: Single projection matrix $E$ is used across all layers, all heads for both key and value.

Sparse Sinkhorn Attention based Transformer~\cite{tay2020sparse} is based on differentiable sorting of internal representations. First, they divide the input sequence into $B$ equal sized blocks each of size $n/B$. A meta sorting network learns to generate latent permutations over these block sequences. Given sorted sequences, we are then able to compute quasi-global attention with only local windows, improving the memory efficiency of the attention module. They also propose Causal Sinkhorn Balancing and SortCut algorithms for causal scenarios for tailoring Sinkhorn Attention for encoding and/or decoding purposes. Their method reduces the memory complexity from $O(n^2)$ to $O(B^2+(n/B)^2)$. The SortCut variant further reduces complexity to linear-time, i.e., $O(nk)$ where $k$ is a user defined budget hyper-parameter much smaller than $n$.

Shen et al.~\cite{shen2018efficient} propose a very simple mathematical trick to reduce quadratic complexity to linear. A typical dot-product attention can be written as $\text{softmax}(QK^T)V$ ignoring the scale factor. This is quadratic because $QK^T$ is $n^2$ in size. This can be rewritten as follows.
\begin{eqnarray}
   \text{Attention}(Q,K,V)=\text{softmax}_r(Q)(\text{softmax}_c(K^TV))
\end{eqnarray}
where $\text{softmax}_r$ and $\text{softmax}_c$ are softmax applied to rows and columns respectively. This revised formulation has terms which are only linear in $n$. Finally, Katharopoulos et al.~\cite{katharopoulos2020transformers} express the self-attention as a linear dot-product of kernel feature maps and make use of the associativity property of matrix products to reduce the complexity from $O(n^2)$ to $O(n)$, where $n$ is the sequence length.

Finally, Longformer~\cite{beltagy2020longformer} propose to reduce Transformer complexity to linear by sparsifying the full self-attention matrix using multiple kinds of attention patterns. The simplest attention pattern is the sliding window pattern which employs a fixed-size window attention surrounding each token. Given a fixed window size $w$, each token attends to $0.5w$ tokens on each side. The computation complexity of this pattern is $O(nw)$ which scales linearly with input sequence length $n$. To further increase the receptive field without increasing computation, the sliding window can be ``dilated''. This is analogous to dilated CNNs where the window has gaps of size dilation $d$. The windowed and dilated attention are not flexible enough to learn task-specific representations. Accordingly, Longformer also has ``global attention'' on few pre-selected input locations. Moreover, this attention operation is symmetric: that is, a token with a global attention attends to all tokens across the sequence, and all tokens in the sequence attend to it. Further, two different sets of projections are used: $Q_s$, $K_s$, $V_s$ to compute attention scores of sliding window attention, and $Q_g$, $K_g$, $V_g$ to compute attention scores for the global attention. The pretrained Longformer consistently outperforms RoBERTa on multiple downstream long document tasks.

\subsection{Summary}

\begin{table}
    \centering
    \scriptsize
    \begin{tabular}{|l|l|l|l|l|l|}
    \hline
Task&Dataset&Model&Base Model&Metric&Eval. (Opt.; Orig)\\
\hline
\hline
Char-level language modeling&1B Word Benchmark&Sinkhorn Mixture Big~\cite{tay2020sparse}&Transformer Big&BPC (L)&1.134; 1.825\\
\hline
Char-level language modeling&1B Word Benchmark&Sparse Transformer~\cite{child2019generating}&Transformer Big&BPC (L)&1.119; 1.825\\
\hline
Char-level language modeling&Enwik8&Longformer~\cite{beltagy2020longformer}&Transformer&BPC (L)&1.00; 1.11\\
\hline
Char-level language modeling&Enwik8&Reformer~\cite{kitaev2020reformer}&Transformer&BPC (L)&1.05; 1.11\\
\hline
Char-level language modeling&text8&Longformer~\cite{beltagy2020longformer}&Transformer&BPC (L)&1.10; 1.18\\
\hline
Coreference resolution&OntoNotes&Longformer-base~\cite{beltagy2020longformer}&RoBERTa-base&F1 (H)&78.6; 78.4\\
\hline
Language modeling&1B Word Benchmark&Sinkhorn Mixture Big~\cite{tay2020sparse}&Transformer Big&Perplexity (L)&27.34; 27.59\\
\hline
Language modeling&1B Word Benchmark&Sparse Transformer~\cite{child2019generating}&Transformer Big&Perplexity (L)&28.77; 27.59\\
\hline
Named Entity Recognition&CoNLL2003&Star Transformer~\cite{guo2019star}&Transformer&Acc (H)&90.93; 86.48\\
\hline
Named Entity Recognition&CoNLL2012&Star Transformer~\cite{guo2019star}&Transformer&Acc (H)&86.30; 83.57\\
\hline
NLI&MNLI&SortCut Sinkhorn~\cite{tay2020sparse}&Transformer&Acc (H)&55.80; 53.69\\
\hline
NLI&QNLI&Linformer~\cite{wang2020linformer}&BERT-base&Acc (H)&91.2; 91.8\\
\hline
NLI&SNLI&Star Transformer~\cite{guo2019star}&Transformer&Acc (H)&86.0; 82.2\\
\hline
NLI&SNLI&SortCut Sinkhorn~\cite{tay2020sparse}&Transformer&Acc (H)&80.30; 78.87\\
\hline
NMT (en$\rightarrow$de)&WMT14&Reformer big~\cite{kitaev2020reformer}&Transformer Big&BLEU (H)&29.1; 27.3\\
\hline
POS Tagging&PTB&Star Transformer~\cite{guo2019star}&Transformer&Acc (H)&97.14; 96.31\\
\hline
Question answering&HotpotQA&BlockBERT (n=2, N=1024)~\cite{qiu2020blockwise}&BERT&F1 (H)&78.94; 77.08\\
\hline
Question answering&HotpotQA&SparseBERT~\cite{child2019generating}&BERT&F1 (H)&76.02; 77.08\\
\hline
Question answering&NaturalQA&BlockBERT (n=2, N=1024)~\cite{qiu2020blockwise}&BERT&F1 (H)&79.39; 78.29\\
\hline
Question answering&NaturalQA&SparseBERT~\cite{child2019generating}&BERT&F1 (H)&77.31; 78.29\\
\hline
Question answering&NewsQA&BlockBERT (n=2, N=1024)~\cite{qiu2020blockwise}&BERT&F1 (H)&70.08; 66.25\\
\hline
Question answering&NewsQA&SparseBERT~\cite{child2019generating}&BERT&F1 (H)&67.16; 66.25\\
\hline
Question answering&SearchQA&BlockBERT (n=2, N=1024)~\cite{qiu2020blockwise}&BERT&F1 (H)&83.51; 80.37\\
\hline
Question answering&SearchQA&SparseBERT~\cite{child2019generating}&BERT&F1 (H)&80.54; 80.37\\
\hline
Question answering&SQuAD 1.1&BlockBERT (n=2, N=1024)~\cite{qiu2020blockwise}&BERT&F1 (H)&90.74; 88.45\\
\hline
Question answering&SQuAD 1.1&SparseBERT~\cite{child2019generating}&BERT&F1 (H)&88.37; 88.45\\
\hline
Question answering&SQuAD 2.0&BlockBERT (n=2, N=1024)~\cite{qiu2020blockwise}&BERT&F1 (H)&81.45; 77.16\\
\hline
Question answering&SQuAD 2.0&SparseBERT~\cite{child2019generating}&BERT&F1 (H)&77.57; 77.16\\
\hline
Question answering&TriviaQA&BlockBERT (n=2, N=1024)~\cite{qiu2020blockwise}&BERT&F1 (H)&79.41; 75.35\\
\hline
Question answering&TriviaQA&SparseBERT~\cite{child2019generating}&BERT&F1 (H)&75.34; 75.35\\
\hline
Question answering&TriviaQA&Longformer-base~\cite{beltagy2020longformer}&RoBERTa-base&F1 (H)&75.2; 74.3\\
\hline
Question answering&WikiHop&Longformer-base~\cite{beltagy2020longformer}&RoBERTa-base&Acc (H)&75.0; 72.4\\
\hline
Sentiment analysis&IMDB&Linformer~\cite{wang2020linformer}&BERT-base&Acc (H)&94.1; 93.5\\
\hline
Sentiment analysis&IMDB&Longformer-base~\cite{beltagy2020longformer}&RoBERTa-base&Acc (H)&95.7; 95.3\\
\hline
Sentiment analysis&SST&Star Transformer~\cite{guo2019star}&Transformer&Acc (H)&52.9; 50.4\\
\hline
Sentiment analysis&SST&Sinkhorn ~\cite{tay2020sparse}&Transformer&Acc (H)&77.52; 76.83\\
\hline
Sentiment analysis&SST-2&Linformer~\cite{wang2020linformer}&BERT-base&Acc (H)&93.1; 92.7\\
\hline
Speech recognition&WSJ&Linear Transformer~\cite{katharopoulos2020transformers}&Reformer&Phoneme Error Rate (L)&8.08; 9.33\\
\hline
Text classification&Hyperpartisan&Longformer-base~\cite{beltagy2020longformer}&RoBERTa-base&F1 (H)&94.8; 87.4\\
\hline
Text classification&MTL-16&Star Transformer~\cite{guo2019star}&Transformer&Acc (H)&86.98; 82.78\\
\hline
Textual similarity&QQP&Linformer~\cite{wang2020linformer}&BERT-base&Acc (H)&90.8; 89.6\\
\hline
    \end{tabular}
\caption{Comparison of various sub-quadratic complexity Transformer methods (sorted by Task and then Dataset). In the metric column, H means high is better while L means low is better. Note that in this case, model sizes do not reduce much; activation memory reduces as described in Section~\ref{sec:linearTransformers} (with comparable or better accuracy) compared to the standard Transformer.}
    \label{tab:linearTransSummary}
\end{table}

Table~\ref{tab:linearTransSummary} compares various sub-quadratic Transformer methods across different tasks and datasets. Accuracy of both the original Transformer and the optimized (opt.) model are shown. These models have been applied for various applications like language modeling (both word level as well as character level), coreference resolution, NER, NLI, NMT, POS, question answering, sentiment analysis, speech recognition, text classification and text similarity analysis. For the same task, dataset and model combination, different papers report different accuracy of the original model because of slight changes in training hyper-parameters; hence we report accuracy of the original model for each row. 

Star Transformers were shown to outperform the vanilla Transformer model across various tasks like Sentiment analysis, Text classification, NLI, POS and NER. On the SST, the Star Transformer achieves 2.5 points improvement against the standard Transformer. On MTL-16, the Star-Transformer outperform the standard Transformer in all 16 datasets, the improvement of the average accuracy is 4.2. Average test times are 10.94 and 49.31ms per batch with batch-size=128 for Star Transformer and standard Transformer respectively.

Longformer achieves a new state-of-the-art on both text8 and enwik8 using the small models with BPC of 1.10 and 1.00 on text8 and enwik8 respectively. Longformer-large model of the same size as Sparse Transformer achieves a BPC of 0.99 on Enwik8, which is the same as that obtained using Sparse Transformer. Also, for both character-level as well as word-level language modeling, on 1B word benchmark, we observe that Sinkhorn mixture (which is a combination of the Sinkhorn attention by mixing it with the vanilla standard dot product attention) performs better than Sparse Transformer. 

On question answering, results have been reported across multiple datasets like HotpotQA, NaturalQA, NewsQA, SearchQA, TriviaQA, WikiHop, SQuAD 1.1 and SQuAD 2.0. We observe that BlockBERT performs better than SparseBERT. Also, it is not surprising that BlockBERT with 2 blocks (n = 2) performs better than that with 3 blocks (n = 3), because it keeps more attention matrix entries. Also, not shown in the table, Longformer-large achieves scores of 81.9 and 73.2 for WikiHop and HotpotQA beating state-of-the-art results by 3.6 and 4 points respectively. 

Longformer consistently outperforms the RoBERTa baseline across many tasks like coreference resolution, question answering, sentiment analysis and text classification. Its performance gain is especially obvious for tasks that require long context such as WikiHop (question answering) and Hyperpartisan (text classification). For TriviaQA (question answering), the improvement is more modest as the local context is often sufficient to answer the question. In the case of HotpotQA (question answering), the supporting fact auxiliary supervision allows models to easily find relevant contexts and then focus on local context, leading to smaller gains. On the IMDB (sentiment analysis) and OntoNotes (coreference resolution) datasets the performance gains are smaller. For IMDB, the majority of the dataset consists of short documents and thus it is expected to see smaller improvements. For OntoNotes, the distance between any two mentions is typically quite small so that a baseline that processes smaller chunks separately is able to stitch together mentions into coreference chains without considering cross chunk interactions.

On speech recognition task, Linear transformers achieve similar performance to vanilla transformers and they are up to 4000x faster on autoregressive prediction of very long sequences. The linear Transformer model outperforms the LSTM and Reformer while being faster to train and evaluate. Reformer takes 2250 seconds per epoch, while Linear Transformers take just 824s/epoch.

To summarize, multiple methods have been proposed to reduce the quadratic complexity of the standard Transformer model. While Sparse Transformers reduce it to $O(n\sqrt{n})$, Reformers reduce it to $O(n\log n)$. Other methods like Star Transformer, Linformer, Sparse Sinkhorn Transformer, Efficient Attention, Linear Transformers and Longformer promise linear complexity. In particular Sparse Sinkhorn Transformer and Longformer have been shown to result into very good accuracy, latency and RAM tradeoff across many tasks.

\section{Summary and Future Directions}
\label{sec:summary}
We discussed various methods for compression of deep learning models for text. Broadly, we discussed pruning, quantization, knowledge distillation, parameter sharing, tensor decomposition, and sub-quadratic Transformer based methods. These methods not just help reduce the model size, but also lead to lower prediction latencies and low power consumption due to reduced computations. Pruning is the oldest method, but not commonly applied for Transformers. Quantization is effective however it is important to use mixed precision balanced quantization with GPU architectures that support efficient low-bit computations. Knowledge Distillation is the most popular method for compression of Transformer models. Parameter sharing is a very useful method but often needs to be combined with other techniques. Matrix decomposition is not very common but has lots of potential, especially the BTD method. Sub-quadratic Transformers are very important to enable processing long documents in applications like query-document similarity, long document summarization, etc.

\subsection{Comparison across model compression method types.}
In the previous six sections, we have compared across multiple methods within each of the six broad types of model compression. In this subsection, we attempt to compare across multiple model compression method types.

Table~\ref{tab:kdSummary2} provides a comparison of model size versus accuracy for various methods across various GLUE~\cite{wang2019glue} and SQuAD tasks. We observe that most of the methods tried on GLUE datasets have been based on knowledge distillation. However, some pruning methods (iterative magnitude pruning, RPP and LayerDrop), quantization (mixed precision QBERT), parameter sharing (ALBERT), tensor decomposition (FLOP) and linear Transformer (Linformer) have also been tried. Quantization methods do not reduce the number of parameters but reduce the number of bits per parameter. Particularly, mixed precision QBERT uses 8 bits for embeddings and 2/3/4 bits for encoder layer weights. Similarly, Linformer does not reduce number of weights but reduces activation memory as well as latency of the overall Transformer model. 
For the remaining models, for each model, we first computed an average GLUE score based on any of the 9 tasks for which scores have been reported. Next, we computed the ratio GLUE score/model size (in Millions). We find the following as the top three 
\begin{itemize}
    \item Distilled-BiLSTM~\cite{tang2019distilling} (ratio=79.3)
    \item Mixed-vocab. training~\cite{zhao2019extreme} (ratio=7.8)
    \item ALBERT-B~\cite{lan2019albert} (ratio=7.2)
\end{itemize}

This clearly tells us that Distilled BiLSTMs provide us the best accuracy versus size tradeoff on GLUE. However, each of these three models actually report results on only 4 out of 9 GLUE tasks. Hence, further, we considered only those methods for which results on at least 5 tasks have been reported and computed the GLUE score/model size ratio. We find the following as the top three 
\begin{itemize}
\item TinyBERT~\cite{jiao2019tinybert} (ratio=5.31)
\item BERT-EMD~\cite{li2020bert} (ratio=5.15)
\item MobileBERT+Quantization~\cite{sun2020mobilebert} (ratio=3.49)
\end{itemize}

Thus, on GLUE tasks, it is clear that distillation based methods (combined with quantization) are better than other types of methods. 

Tables~\ref{tab:pruningSummary},~\ref{tab:quantizationSummary},~\ref{tab:kdSummary1},~\ref{tab:paramSharingSummary},~\ref{tab:tensorDecompSummary} and~\ref{tab:linearTransSummary} compare various pruning, quantization, knowledge distillation, parameter sharing, tensor decomposition and sub-quadratic Transformer methods across different tasks and datasets. Overall, there are 123 (task, dataset) combinations across these six tables which implies that unlike GLUE, not many methods have been applied on the same set of (task, dataset) combinations\footnote{We make the entire statistics available as an excel file at \url{https://bit.ly/3vmaxZ9}.}. The most popular tasks are language modeling (on PTB and 1B word benchmark), sentiment analysis (on SST) and NMT (WMT14 en$\rightarrow$de). 

For language modeling on PTB, on LSTM models, bank balanced sparsity~\cite{cao2019efficient} based pruning method worked best. With Transformer models, Block Term Decomposition (BTD)~\cite{ma2019tensorized} method seems to work best. Among various methods like parameter sharing, tensor decomposition and sub-quadratic complexity Transformer which have been tried for language modeling on 1B Word Benchmark, again, BTD~\cite{ma2019tensorized} method seems to work best leading to a model with 0.16B parameters and a perplexity as low as 19.5. Multiple datasets have been used for Neural machine translation (NMT). Datasets from WMT and IWSLT are the most popular. Among these, the en$\rightarrow$de from WMT14 is the most popular dataset used for testing various NMT models. For en$\rightarrow$de NMT with WMT14, using 2-layer LSTMs, the best accuracy versus size tradeoff is using Pruned Seq-KD + Seq-Inter~\cite{kim2016sequence} which gives a 8M size model leading to 18.5 BLEU. Among Transformer based models, BTD~\cite{ma2019tensorized} leads to a Transformer model which provides 34.91 BLEU with 21.2M model size.

\begin{table*}
    \centering
    \scriptsize
    \begin{tabular}{|p{1in}|p{2in}|p{2.7in}|}
    \hline
Task&Popular Datasets&References\\
\hline
\hline
Language modeling&Penn TreeBank Corpus, One billion word benchmark, Europarl, WikiText-103, text8, source code of Linux kernel, 2013 ACL Workshop Morphological Language Datasets (ACLW), Arabic news commentary corpus, 2013 ACL workshop on MT, enwik8 (from Wikipedia), Lambada&Neuron Pruning~\cite{murray2015auto}, Iterative magnitude pruning~\cite{zhu2017prune}, Block sparsity~\cite{cao2019efficient},Loss -Aware Quantization~\cite{xu2018alternating,hou2018loss}, Uniform Quantization~\cite{he2016effective,kapur2017low}, Binary Quantization~\cite{hubara2017quantized}, HitNet~\cite{wang2018hitnet}, Sparse Word Representations~\cite{chen2016compressing}, LightRNN~\cite{li2016lightrnn}, Slim Embeddings~\cite{li2018slim}, C2W~\cite{ling2015finding}, LayerDrop~\cite{fan2019reducing}, Reformer~\cite{kitaev2020reformer}, Linformer~\cite{wang2020linformer}, Char-CNN~\cite{jozefowicz2016exploring}, CNN+Highway Network~\cite{kim2016character}, SparseBERT~\cite{child2019generating}, FLOP~\cite{wang2019structured}, Deep Equilibrium Models~\cite{bai2019deep}, WEST~\cite{variani2019west}, Sparse Sinkhorn Attention~\cite{tay2020sparse}, BTD~\cite{ma2019tensorized}, Universal Transformers~\cite{dehghani2018universal}, TT-embedding~\cite{khrulkov2019tensorized}, multiple methods~\cite{grachev2019compression}\\
\hline
Neural Machine translation (NMT)&IWSLT German-English, IWSLT Thai-English, ASPEC English-Japanese, WMT English-German, WMT German-English, WMT English-Russian, IWSLT English Vietnamese, WMT English-Romanian, WMT  English-Estonian, Ted Talk&Compositional codes~\cite{shu2017compressing}, LayerDrop~\cite{fan2019reducing}, Pruning attention heads~\cite{voita2019analyzing,michel2019sixteen}, Neuron Pruning~\cite{murray2015auto}, Magnitude Pruning~\cite{see2016compression}, Iterative magnitude pruning~\cite{zhu2017prune,cheong2019transformers}, Pruned Seq-KD + Seq-Inter~\cite{kim2016sequence}, Quantized Distillation~\cite{polino2018model}, Teacher ensembles KD~\cite{freitag2017ensemble}, Multiple teachers KD~\cite{tan2019multilingual}, BTD~\cite{ma2019tensorized}, Quaternion Attention~\cite{tay2019lightweight}, Universal Transformers~\cite{dehghani2018universal}, TT-embedding~\cite{khrulkov2019tensorized}\\
\hline
Sentiment Analysis&IMDB movie review, SST, SST-2, Elec (electronic product reviews)&Compositional codes~\cite{shu2017compressing}, Star Transformer~\cite{guo2019star}, TinyBERT~\cite{jiao2019tinybert}, MiniLM~\cite{wang2020minilm}, Linformer~\cite{wang2020linformer}, XtremeDistil~\cite{mukherjee2020xtremedistil}, Gaussian Quantization~\cite{alom2018effective}, Uniform Quantization~\cite{he2016effective}, Sparse coding~\cite{faruqui2015sparse}, Quaternion Attention~\cite{tay2019lightweight}, Sparse Sinkhorn Attention~\cite{tay2020sparse}, RPP~\cite{guo2019reweighted}, ALBERT~\cite{lan2019albert}, Patient KD~\cite{sun2019patient}, Mixed-vocabulary KD training~\cite{zhao2019extreme}, Distilled-BiLSTM~\cite{tang2019distilling}, MTDNN~\cite{liu2019multi}, TT-embedding~\cite{khrulkov2019tensorized}\\
\hline
Question Answering&SQuAD1.1, SQuAD2.0, ELI5, SemEval, BABI&LayerDrop~\cite{fan2019reducing}, MiniLM~\cite{wang2020minilm}, RPP~\cite{guo2019reweighted}, BS-Fixed Quantization~\cite{lam2018word2bits}, ALBERT~\cite{lan2019albert}, KD~\cite{damani20_pakdd}, Universal Transformers~\cite{dehghani2018universal}\\
\hline
Natural Language Inference&SNLI, MNLI-m, MNLI-mm, QNLI, RTE, WNLI, XNLI&Star Transformer~\cite{guo2019star}, LayerDrop~\cite{fan2019reducing}, TinyBERT~\cite{jiao2019tinybert}, MiniLM~\cite{wang2020minilm}, Linformer~\cite{wang2020linformer}, Sparse Sinkhorn Attention~\cite{tay2020sparse}, RPP~\cite{guo2019reweighted}, ALBERT~\cite{lan2019albert}, Patient KD~\cite{sun2019patient}, Mixed-vocabulary KD training~\cite{zhao2019extreme}, Distilled-BiLSTM~\cite{tang2019distilling}, MTDNN~\cite{liu2019multi}\\
\hline
Paraphrasing&QQP, STS-B&TinyBERT~\cite{jiao2019tinybert}, MiniLM~\cite{wang2020minilm}, Linformer~\cite{wang2020linformer}, RPP~\cite{guo2019reweighted}, ALBERT~\cite{lan2019albert}, Patient KD~\cite{sun2019patient}, Distilled-BiLSTM~\cite{tang2019distilling}, MTDNN~\cite{liu2019multi}\\
\hline
Image captioning&MSCOCO&Grow and Prune~\cite{dai2018grow}, Magnitude Pruning~\cite{han2015deep}, Iterative Magnitude Pruning and Densification~\cite{han2016dsd}\\
\hline
Handwritten character recognition&ICDAR&SVD and Pruning\cite{yang2018accelerating}\\
\hline
Part-of-speech (POS) tagging&Wall Street Journal of the Penn Treebank dataset, WikiAnn NER corpus&C2W~\cite{ling2015finding},  XtremeDistil~\cite{mukherjee2020xtremedistil}\\
\hline
Summarization&CNN-DailyMail, XSum&LayerDrop~\cite{fan2019reducing}, MiniLM~\cite{wang2020minilm}\\
\hline
Machine Reading Comprehension& Microsoft Research Paraphrase Corpus (MRPC), ReAding Comprehension from Examinations (RACE)&LayerDrop~\cite{fan2019reducing}, TinyBERT~\cite{jiao2019tinybert}, MiniLM~\cite{wang2020minilm}, RPP~\cite{guo2019reweighted}, ALBERT~\cite{lan2019albert}, Patient KD~\cite{sun2019patient}, Mixed-vocabulary KD training~\cite{zhao2019extreme}, MTDNN~\cite{liu2019multi}\\
\hline
Linguistic Acceptability&CoLA&TinyBERT~\cite{jiao2019tinybert}, MiniLM~\cite{wang2020minilm}, RPP~\cite{guo2019reweighted}, ALBERT~\cite{lan2019albert}, MTDNN~\cite{liu2019multi}\\
\hline
Topic Classification&DbPedia, Ag News,20 Newsgroup&XtremeDistil~\cite{mukherjee2020xtremedistil}, Sparse coding~\cite{faruqui2015sparse}\\
\hline
Question Type Classification&TREC&Sparse coding~\cite{faruqui2015sparse}\\
\hline
Noun Phrase Bracketing&Lazaridou~\cite{lazaridou2013fish}&Sparse coding~\cite{faruqui2015sparse}\\
\hline
Word Similarity&SimLex-999, MEN, MTurk, RARE, SCWS, WSR, WSS&Sparse coding~\cite{faruqui2015sparse}, Shared reference vectors~\cite{suzuki2016learning}\\
\hline
Mathematical Language Understanding&Wangperawong's MLU~\cite{wangperawong2018attending}&Quaternion Attention~\cite{tay2019lightweight}\\
\hline
Subject Verb Agreement&Linzen~\cite{linzen2016assessing}&Quaternion Attention~\cite{tay2019lightweight}, Universal Transformers~\cite{dehghani2018universal}\\
\hline
Word Analogy&GSEM, GSYN, MSYN&Shared reference vectors~\cite{suzuki2016learning}\\
\hline
Sentence Completion&MSC&Shared reference vectors~\cite{suzuki2016learning}\\
\hline
Learning to execute&Zaremba and Sutskever~\cite{zaremba2014learning}&Universal Transformers~\cite{dehghani2018universal}\\
\hline
Ad Click Through Rate Prediction&Criteo Kaggle&TT-embedding~\cite{khrulkov2019tensorized}\\
\hline
Speech Recognition&2100 hours English Speech, AN4, Switchboard, TIMIT, WSJ 92, WSJ 93, TIDIGITS, 3M Google voice utterances, Live traffic utterances&Iterative Magnitude Pruning~\cite{narang2017exploring}, Grow and Prune~\cite{dai2018grow}, Neuron Pruning~\cite{he2014reshaping}, Block Sparsity~\cite{cao2019efficient}, BBS~\cite{cao2019efficient}, DSD~\cite{han2016dsd}, Pow2 Ternarization~\cite{ott2016recurrent}, Loss Aware Quantization~\cite{hwang2014fixed}, Toeplitz-like~\cite{lu2016learning}, Joint-SVD~\cite{prabhavalkar2016compression}, Projections~\cite{sak2014long}, WEST~\cite{variani2019west}\\
\hline
Named entity recognition (NER)&CoNLL2003, Wikiann-41, CoNLL2012&QBERT~\cite{shen2019q}, XtremeDistill~\cite{mukherjee2020xtremedistil}, Star Transformer~\cite{guo2019star}\\
\hline
Intent Detection&SNIPS&Mixed-vocabulary KD training~\cite{zhao2019extreme}\\
\hline
Question Generation&SQuAD 1.1&MiniLM~\cite{wang2020minilm}\\
\hline
Slot Filling&SNIPS&Mixed-vocabulary KD training~\cite{zhao2019extreme}\\
\hline
Text classification&20 Newsgroup, Hyperpartisan, MTL-16&Sparse coding~\cite{faruqui2015sparse}, Longformer~\cite{beltagy2020longformer}, Star Transformer~\cite{guo2019star}\\
\hline
Coreference Resolution&OntoNotes&Longformer~\cite{beltagy2020longformer}\\
\hline
    \end{tabular}
    \caption{Applications of Model Compression Methods for Text}
    \label{tab:applications}
\end{table*}

Combinations of multiple model compression method types has also been experimented with and found to be effective. Some examples of such combinations include the following:
\begin{itemize}
    \item Pruning + Tensor Decomposition ~\cite{he2014reshaping,wang2019structured}
    \item Pruning + Quantization~\cite{cao2019efficient}
    \item Knowledge distillation + Quantization~\cite{bengio2013estimating,mishra2017apprentice,polino2018model,sun2020mobilebert}
    \item Knowledge distillation + Pruning~\cite{kim2016sequence}
    \item Tensor decomposition + Parameter sharing~\cite{lu2016learning,lan2019albert}
    \item Tensor decomposition + Quantization~\cite{shu2017compressing}
\end{itemize}

Recently, Kim et al.~\cite{kim2020fastformers} combined  knowledge distillation, structured pruning and quantization leading to drastic improvements on inference efficiency. First, they  investigate the efficacy of various Knowledge Distillation techniques to significantly reduce the size of the models with respect to the depth and hidden state sizes while preserving the
accuracy. Second, they explore Structured Pruning that further reduces the size of the models by reducing the number of self-attention heads and the number of intermediate hidden states in the feedforward layers to achieve more efficiency while
trying to preserve the accuracy as well. Finally, they 
explore model quantization which enables faster model executions by optimally utilizing hardware acceleration capabilities. Such a combined method leads to heavily reduced model size, 12.4x GPU speed-up and 6.9x-125.8x reduction in energy consumption.

\subsection{Summary of Applications}

The model compression methods mentioned in this survey have been used across a wide variety of text processing tasks. In Table~\ref{tab:applications}, we list down the tasks, popular datasets and references where the readers can find more discussion around model size versus accuracy tradeoff.

\subsection{Future Trends}

Although there has been so much of work already in this field, there is a lot more work to be done. 
\begin{itemize}
    \item With linear Transformer models, one can afford to have input with tens of thousands of tokens. Hence, many tasks need to be redesigned where large context can now be included as input to improve accuracy. 
    \item Combinations of several methods have not been tested well. Recently, Fastformers method~\cite{kim2020fastformers} showed that combining multiple methods like knowledge distillation, 16-bit quantization, structured pruning and numerical optimizations can lead to drastic improvements. However, lot of experiments are needed to further check how models respond to combination of model compression methods.
    \item Latency results vary based on GPU architectures. With new GPU architectures (Nvidia RTX 3080, Nvidia T4), some methods like quantization may become more impactful.
    \item Real world settings are often complex: multi-modal, multi-task, multi-label, small-data, noisy labels, multi-teachers, mismatching teacher-student architectures. Efficient ways of recommending the most promising method is necessary.
    \item Different components/structures of a model may respond to different kinds of compression methods with specific hyper-parameters. A generic method to choose the right method for various structures is needed.
    \item How does compression of models impact their interpretability? Can we design model compression mechanisms aimed at looking at a tradeoff between model accuracy, size, latency and interpretability.
    \item None of the model compression methods performs any application specific compression. Can we obtain further compression by exploiting some task-specific patterns?
    \item Recently, deep reinforcement learning based methods have been proposed in the computer vision community~\cite{he2018amc,yuan2019enhanced}. It will be nice to check the effectiveness of such methods for NLP tasks.
\end{itemize}

We hope that this survey acts as a good guide to folks across academia and industry. Also, we hope that a significantly large chunk of research gets done in the area of model compression to enable good accuracy across many NLP tasks while keeping model sizes and latencies in check. 


\bibliographystyle{ACM-Reference-Format}
\bibliography{referencesShort}

\end{document}